\documentclass[10pt,journal,compsoc]{IEEEtran}

\ifCLASSOPTIONcompsoc
  \usepackage[nocompress]{cite}
\else
  \usepackage{cite}
\fi

\usepackage{wrapfig}
\usepackage{graphicx}
\usepackage[table]{xcolor}
\definecolor{localcolor}{rgb}{1.0, 0.89, 0.88}
\definecolor{fullcolor}{rgb}{0.6, 1.0, 0.6}
\definecolor{dispcolor}{rgb}{1,1,0.7}
\definecolor{imagecolor}{rgb}{0.88,1,1}
\definecolor{lrcolor}{rgb}{1, 0.74, 0.53}
\definecolor{distcolor}{rgb}{0.88,0.69,1}
\definecolor{sgmcolor}{rgb}{0.8, 0.8, 0.8}
\definecolor{forestvol}{rgb}{0.4, 0.6, 0.8}
\definecolor{forestdisp}{rgb}{0.6, 0.8, 1}
\definecolor{deepvol}{rgb}{1, 0.3, 0.3}
\definecolor{deepdisp}{rgb}{1, 0.6, 0.6}
\usepackage{verbatim}
\usepackage{amsmath}
\usepackage{hyperref}
\usepackage[percent]{overpic}
\usepackage{pifont}
\usepackage{algorithm}
\usepackage{algpseudocode}
\usepackage{tabularx,booktabs}
\usepackage{subfigure}
\usepackage{mdframed}
\usepackage[export]{adjustbox}

\def\eg{\emph{e.g.}}

\def\ie{\emph{i.e.}}

\def\kitti{KITTI}
\def\eth{ETH3D}
\def\midd{Middlebury 2014}

\newcommand{\cmark}{\ding{51}}%
\newcolumntype{;}{!{\vrule width 2pt}}
\newcolumntype{.}{!{\vrule width 3pt}}

\begin{document}

\title{On the confidence of stereo matching in a deep-learning era: a quantitative evaluation}

\author{Matteo~Poggi,~\IEEEmembership{Member,~IEEE,}
        Seungryong~Kim,~\IEEEmembership{Member,~IEEE,}
        Fabio~Tosi,~\IEEEmembership{Student Member,~IEEE,}
        Sunok~Kim,~\IEEEmembership{Member,~IEEE,}
        Filippo~Aleotti,        
        Dongbo~Min,~\IEEEmembership{Senior Member,~IEEE,}
        Kwanghoon~Sohn,~\IEEEmembership{Senior Member,~IEEE,}
        Stefano~Mattoccia,~\IEEEmembership{Member,~IEEE}
\IEEEcompsocitemizethanks{\IEEEcompsocthanksitem M. Poggi, F. Tosi, F. Aleotti and S. Mattoccia are with University of Bologna, Italy, 40136.

\IEEEcompsocthanksitem S. Kim is with Korea University, Seoul, Korea

\IEEEcompsocthanksitem S. Kim is with Korea Aerospace University, Goyang, Korea

\IEEEcompsocthanksitem D. Min is with Ewha Womans University, Seoul, Korea

\IEEEcompsocthanksitem K. Sohn is with Yonsei University, Seoul, Korea}

}


\IEEEtitleabstractindextext{%
\begin{abstract}
Stereo matching is one of the most popular techniques to estimate dense depth maps by finding the disparity between matching pixels on two, synchronized and rectified images.
Alongside with the development of more accurate algorithms, the research community focused on finding good strategies to estimate the reliability, i.e. the confidence, of estimated disparity maps. This information proves to be a powerful cue to naively find wrong matches as well as to improve the overall effectiveness of a variety of stereo algorithms according to different strategies. 
In this paper, we review more than ten years of developments in the field of confidence estimation for stereo matching. We extensively discuss and evaluate existing confidence measures and their variants, from hand-crafted ones to the most recent, state-of-the-art learning based methods.
We study the different behaviors of each measure when applied to a pool of different stereo algorithms and, for the first time in literature, when paired with a state-of-the-art deep stereo network. Our experiments, carried out on five different standard datasets, provide a comprehensive overview of the field, highlighting in particular both strengths and limitations of learning-based strategies.

\end{abstract}

}

\maketitle

\IEEEdisplaynontitleabstractindextext

\IEEEpeerreviewmaketitle

\IEEEraisesectionheading{\section{Introduction}\label{sec:introduction}}

\IEEEPARstart{D}{}epth estimation is often the starting point for solving higher level computer vision tasks such as tracking, localization, navigation and more.
Although a variety of active sensors are available for this purpose, image-based techniques are often preferred thanks to the increasing availability of standard cameras on most consumer devices. Among them, binocular stereo \cite{scharstein2002taxonomy} is one of the most popular and studied in the literature. Given two synchronized images acquired by a calibrated stereo rig, depth can be estimated by means of triangulation after finding the displacement between matching pixels on the two images, \ie{} the \textit{disparity}. This search is limited to a 1D search range in case of rectified images. Specifically, by selecting one of the two images as \textit{reference}, for each pixel we look for the corresponding one on the other view, namely \textit{target}, among a number of candidates on the same, horizontal scanline.

Over the past few decades, a great variety of algorithms have been proposed, broadly classified into local or global methods according to the deployed steps formalized in \cite{scharstein2002taxonomy}, that are i) matching cost computation, ii) cost aggregation, iii) disparity optimization and selection, and iv) refinement. 
Common to all algorithms is the definition of a \textit{cost volume}, collecting for each pixel in the reference image matching costs for corresponding candidates on the target image. Among all, solutions based on the Semi-Global Matching pipeline (SGM \cite{hirschmuller2005accurate}) resulted in the years the most popular thanks to the good trade-off between accuracy and computational complexity.

Similar to other computer vision tasks, deep learning has hit stereo matching as well \cite{poggi2020synergies}, at first replacing single steps in the pipeline such as matching cost computation with convolutional neural networks (CNNs) \cite{zbontar2016stereo}, rapidly converging towards end-to-end deep networks \cite{Mayer_2016_CVPR} embodying the entire pipeline. Nowadays, the state-of-the-art is represented by these latter approaches \cite{Zhang2019GANet}, although several limitations still preclude their seamless deployment in real world applications \cite{Tonioni_2019_CVPR,Tonioni_2019_learn2adapt,tonioni2019unsupervised}. 

In parallel with this rapid evolution, estimating the \textit{confidence} of estimated disparity maps, as shown in Figure \ref{fig:overview}, has grown in popularity. At first used for selecting most reliable estimates or filtering out outliers, more techniques leveraging confidence measures have been studied and developed. Specifically, most methods aim at improving pre-existing stereo algorithms \cite{spyropoulos2014learning,spyropoulos2016correctness}, with particular focus on SGM variants \cite{park2015leveraging,park2018learning,seki2016patch,poggi2016learning,poggi2020learning}. Other notable applications consist into fusion with Time-Of-Flight sensors \cite{marin2016reliable,poggi2020confidence}, as well as domain adapation of deep stereo networks \cite{Tonioni_2017_ICCV,tonioni2019unsupervised}.
Starting from the first review in the field \cite{hu2012quantitative}, several strategies to estimate a confidence measure have been proposed in the literature, either hand-made or learned from data by means of machine learning \cite{poggi2017quantitative}. More recent works belonging to this latter category \cite{tosi2018beyond,gul2019pixel,mehltretter2019cnn,Kim_2019_CVPR} rapidly established as state-of-the-art. 

In this paper, we provide a comprehensive review and evaluation of confidence measures, covering more than 10 years of studies in this field. 
This extensive survey extends our previous work \cite{poggi2017quantitative}, representing the most recent evaluation available in literature, with the following novelties:

\begin{figure*}[t]
    \centering
    \renewcommand{\tabcolsep}{1pt}
    \begin{tabular}{ccc}
        \includegraphics[width=0.28\textwidth]{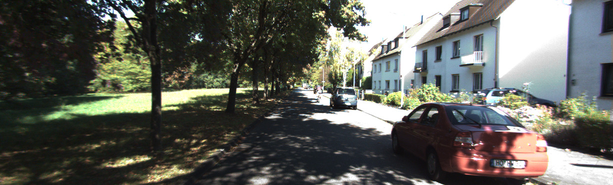} &
        \includegraphics[width=0.28\textwidth]{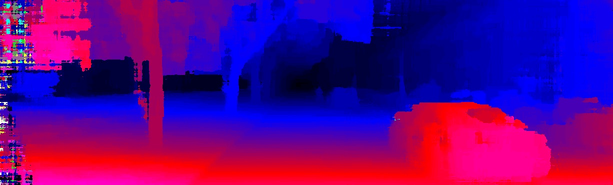} &
        \includegraphics[width=0.28\textwidth]{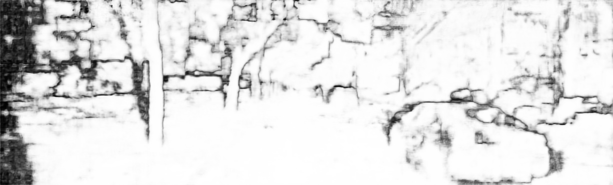} \\
    \end{tabular}
    \caption{\textbf{Confidence estimation example.} From left to right, reference image, disparity map and estimated confidence map (pixels from black to white encode confidence from lower to higher).}
    \label{fig:overview}
\end{figure*}

\begin{itemize}
    \item We include the latest advances in the field of confidence estimation, either hand-made \cite{veld2018novel} or based on deep learning \cite{tosi2018beyond,gul2019pixel,mehltretter2019cnn,Kim_2019_CVPR}
    
    \item We evaluate each confidence measure on a total of 4 realistic datasets, respectively KITTI 2012 \cite{Geiger_CVPR_2012} and 2015 \cite{Menze2015CVPR}, Middlebury 2014 \cite{scharstein2014high} and ETH3D \cite{schops2017multi}, together with the SceneFlow Driving synthetic dataset \cite{Mayer_2016_CVPR}. 
    Indeed, for the first time, we assess the ability of learning-based measures to tackle domain shift issues, training in one domain (\eg, on a synthetic dataset) and testing in different ones (\eg, real).
    
    \item For the first time in literature, we evaluate all the considered measures when applied to state-of-the-art 3D end-to-end stereo network, \ie{} GANet \cite{Zhang2019GANet}. 
    
\end{itemize}

The rest of the manuscript is organized as follows: Section \ref{sec:related} briefly resumes the progressive development in the field of confidence estimation and its applications, Section \ref{sec:hand} introduces the taxonomy of hand-crafted confidence measures, while Section \ref{sec:learning} lists and classifies learning based approaches. Then, Section \ref{sec:results} collects the outcome of our extensive experiments, summarized in Section \ref{sec:discussion} before drawing conclusions in Section \ref{sec:conclusions}.

\section{Related Work}
\label{sec:related}
In last decades, there have been extensive works in stereo confidence measures, mainly based on handcrafted confidence measures~\cite{Egnal2002,Egnal2004,Mordohai2009}. Hu and Mordohai~\cite{hu2012quantitative} performed a taxonomy and evaluation of stereo confidence measures, considering 17 confidence measures and two local algorithms on the two datasets available at that time. 	
Since then novel confidence measures were proposed~\cite{breiman2001random,haeusler2013ensemble,spyropoulos2014learning,spyropoulos2016correctness,park2015leveraging,park2018learning} and more importantly this field was affected by methodologies inspired by the machine learning. To account for the fact that there is no single confidence feature yielding stably optimal performance for all datasets and stereo matching algorithms, methods aiming to benefit from the combination of multiple confidence measures have been proposed~\cite{haeusler2013ensemble,spyropoulos2014learning} with a random forest. Following this strategy, the reliability of confidence measures was further improved by considering more effective features~\cite{park2015leveraging,park2018learning}, an efficient O(1) computation~\cite{poggi2016learning}, and hierarchical aggregation at a superpixel-level~\cite{kim2017feature}.   

Recent approaches have tried to measure the confidence through deep CNNs~\cite{poggi2016bmvc,seki2016patch,Fu_2017_RFMI,fu2018learning,tosi2018beyond,Shaked_2017_CVPR,kim2019unified,Kim_2019_CVPR,Kim2020adversarial,gul2019pixel,mehltretter2019cnn}. Formally, CNN-based methods first extract the confidence features directly from input cues, i.e. reference image, cost volume, and disparity maps, and then predict the confidence with a classifier. Various methods have been proposed that use the single- or bi-modal input, i.e. left disparity~\cite{poggi2016bmvc}, both left and right disparity~\cite{seki2016patch}, a matching cost~\cite{Shaked_2017_CVPR}, matching cost and disparity~\cite{kim2019unified}, and disparity and color image~\cite{Fu_2017_RFMI,tosi2018beyond}. More recently, Kim et al.~\cite{Kim_2019_CVPR} present a deep network that estimates the confidence by making full use of tri-modal input, including matching cost, disparity, and reference image with a novel fusion technique. 
Concerning unsupervised training of confidence measures, Mostegel et al.~\cite{MOSTEGEL_CVPR_2016} proposed to determine training labels made of a set of correct disparity assignment and a set of wrong ones, exploiting, respectively, consistencies and  contradictions between multiple depth maps. Differently, Tosi et al.~\cite{Tosi_2017_BMVC} leveraged on a pool of confidence measures for the same purpose.

This field has also seen the deployment of confidence measures plugged into stereo vision pipelines to improve the overall accuracy as proposed in~\cite{poggi2017quantitative,marin2016reliable,Poggi_ICIAP_2017}. Most previous approaches aimed at improving the accuracy of SGM~\cite{Hirschmuller2008} algorithm exploiting as a cue an estimated match reliability. In addition, confidence measures have been effectively deployed for sensor fusion combining depth maps from multiple sensors~\cite{marin2016reliable} and for embedded stereo systems~\cite{Poggi_ICIAP_2017}.

\begin{figure}[t]
    \centering
    \begin{tabular}{c}
        \includegraphics[trim=0.1cm 13cm 21.5cm 0.2cm,clip,width=0.48\textwidth]{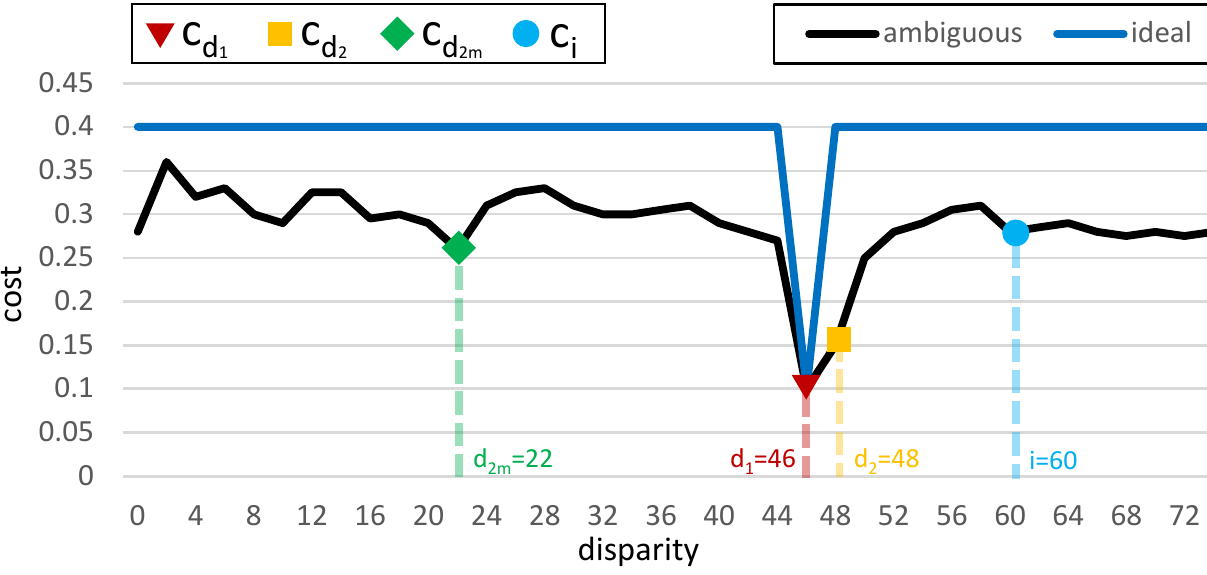} \\ 
    \end{tabular}
    \caption{ \textbf{Example of cost curves} for a pixel $p$: on x axis, disparity hypotheses $i$, on y axis, matching cost $c_i$. We show an ambiguous curve in black, for which ${d_1}$ and ${d_2}$ (respectively equal to 46 and 48) compete for the role of minimum and other local minima exist (at disparities 22 and 60, the former corresponding to ${d_{2m}}$). We also show an ideal cost curve in blue with a clear winner. Best viewed in color.}
    \label{fig:costs}
\end{figure}

\section{Hand-crafted confidence measures}
\label{sec:hand}

Common to the variety of confidence measures proposed in the literature is using the cost volume as the source of information. However, most measures only process a portion of the cues available from it, ranging from properties of the per-pixel full cost curve down to simply leveraging the output disparity map only.
In compliance with previous works \cite{hu2012quantitative,poggi2017quantitative,park2018learning}, we define a taxonomy with the aim of grouping confidence measures into categories according to the input cues extracted from the cost volume. 

We first define the naming convention used in the rest of this section. Given two rectified images $l$ and $r$ and assuming the former as reference, for each pixel $p$ at coordinates $(x,y)$ a cost curve $c(p)$ is computed. We define the following terms

\begin{itemize}
    \item $l(p),r(p)$: pixel intensity in image $l,r$
    \item $c_i(p)$: matching cost for disparity hypothesis $i \in D$
    \item $d_1(p)$: winning disparity hypothesis
    \item $d_2$(p): disparity hypothesis of the second minimum
    \item $d_{2m}(p)$: disparity hypothesis of the second \textit{smallest} local minimum. Any $c_i$ is a local minimum if $c_i<c_{i\pm1}$, yet $d_{2m}(p)$ may not be defined
    \item $c_{d_1}(p)$: minimum cost in the curve
    \item $c_{d_2}(p)$: second minimum in the cost curve
    \item $c_{d_{2m}}(p)$: second local minimum in the cost curve
    \item $p^r$: pixel on $r$ at coordinates $x-d_1(p)$ 
    \item $\mu$: mean over a local window, \eg{} $\mu(l(p))$ represents the mean intensity over a window in image $l$. 
\end{itemize}
When omitted, costs and disparities always refer to $l$ as the reference image. Otherwise, we label them as $c^r$ and $d^r$ when assuming $r$ as reference. 
Fig. \ref{fig:costs} depicts an example of a cost curve, highlighting the positions of specific costs defined earlier. In particular, we show an \textit{ambiguous} curve (black) where several scores compete for the minimum, increasing the likelihood of selecting a wrong disparity with respect to the case of having an \textit{ideal} curve (blue).

We are now going to define, in Sec. from \ref{sec:local} to \ref{sec:sgm}, different families of measures according to the input cues they process. To each category, we assign a color that will be recalled when discussing the results of our evaluation.

\subsection{\colorbox{localcolor}{Minimum cost and local properties}}\label{sec:local}

We group here methods considering only local information in the cost curve, mostly encoded by $c_{d_1}$, $c_{d_2}$, and $c_{d_{2m}}$, from pixel $p$ and eventually its neighbors. Most measures use $c_{d_{2m}}$, that may not be defined as in the case, for instance, of an ideal cost curve. Indeed, variants of these measures, called \textit{naive}, use $c_{d_2}$ that is always defined.

\textbf{Matching Score Measure} (MSM) \cite{Egnal2004}, expressed by the negative minimum cost itself (the lower, the higher confidence)
\begin{equation}\label{msm}
    \text{MSM}(p) = -c_{d_1}(p)
\end{equation}

\textbf{Maximum Margin} (MM) \cite{poggi2017quantitative}, has the difference between the second smallest local minimum and the minimum cost
\begin{equation}\label{mm}
    \text{MM}(p) = c_{d_{2m}}(p) - c_{d_1}(p)
\end{equation}

\textbf{Maximum Margin Naive} (MMN) \cite{hu2012quantitative}, has the difference between the second minimum and the minimum cost 
\begin{equation}\label{mmn}
    \text{MMN}(p) = c_{d_2}(p) - c_{d_1}(p)
\end{equation}

\textbf{Non-Linear Margin} (NLM) \cite{Haeusler2012evaluation}, has exponential of the MM
\begin{equation}\label{nlm}
    \text{NLM}(p) = e^{\frac{c_{d_{2m}}(p) - c_{d_1}(p)}{2\sigma^2}}
\end{equation}

\textbf{Non-Linear Margin Naive} (NLMN) \cite{poggi2017quantitative}, has exponential of the MMN
\begin{equation}\label{nlmn}
    \text{NLMN}(p) = e^{\frac{_{d_2}(p) - c_{d_1}(p)}{2\sigma^2}}
\end{equation}

\textbf{Curvature} (CUR) \cite{Egnal2004}, as the local shape of the cost curve in correspondence of the minimum cost
\begin{equation}\label{cur}
    \text{CUR}(p) = -2c_{d_1}(p) + c_{d_1-1}(p) + c_{d_1+1}(p)
\end{equation}

\textbf{Local Curve} (LC) \cite{Wedel2009detection}, as the slope of the cost curve between the minimum cost and the higher of its neighbors
\begin{equation}\label{lc}
    \text{LC}(p) = \frac{\max{[c_{d_1-1}(p),c_{d_1+1}(p)]} -c_{d_1}(p)}{\gamma}
\end{equation}

\textbf{Peak Ratio} (PKR) \cite{Egnal2004}, as the ratio between the second local minima and the minimum cost
\begin{equation}\label{pkr}
    \text{PKR}(p) = \frac{c_{d_{2m}}(p)}{c_{d_1}(p)}
\end{equation}

\textbf{Peak Ratio Naive} (PKRN) \cite{hu2012quantitative}, as the ratio between the second minima and the minimum cost
\begin{equation}\label{pkrn}
    \text{PKRN}(p) = \frac{c_{d_2}(p)}{c_{d_1}(p)}
\end{equation}

\textbf{Disparity Ambiguity Measure} (DAM) \cite{haeusler2013ensemble}, as the distance between two disparity hypotheses expressed by the minimum cost and the second minima
\begin{equation}\label{dam}
    \text{DAM}(p) = |d_1(p) - d_2(p)|
\end{equation}

\textbf{Average Peak Ratio} (APKR) \cite{kim2014stereo}, as the average of ratios between costs for pixels $q$ in window, respectively in the same position of the second smallest local minimum and the minimum cost in $p$

\begin{equation}\label{apkr}
    \text{APKR}(p) = \sum_{q \in N(p)} \frac{c_{d_{2m}(p)}(q)}{c_{d_1(p)}(q)}
\end{equation}

\textbf{Average Peak Ratio Naive} (APKRN) \cite{poggi2017quantitative}, naive variant of the previous measure replacing second smallest local minimum with second minimum
\begin{equation}\label{apkrn}
    \text{APKRN}(p) = \sum_{q \in N(p)} \frac{c_{d_2(p)}(q)}{c_{d_1(p)}(q)}
\end{equation}

\textbf{Weighted Peak Ratio} (WPKR) \cite{kim2016weighted}, as the average of ratios as defined for APKR. Each ratio is multiplied by a binary weight $\alpha(p,q)=1$ if the difference between pixel intensities $l(p)$ and $r(q)$ is lower than a threshold $w$
\begin{equation}\label{wpkr}
    \text{WPKR}(p) = \sum_{q \in N(p)} \alpha(p,q) \cdot \frac{c_{d_{2m}(p)}(q)}{c_{d_1(p)}(q)}
\end{equation}

\textbf{Weighted Peak Ratio Naive} (WPKRN) \cite{poggi2017quantitative}, naive variant of the previous measure replacing second smallest local minimum with second minimum
\begin{equation}\label{wpkrn}
    \text{WPKRN}(p) = \sum_{q \in N(p)} \alpha(p,q) \cdot \frac{c_{d_2(p)}(q)}{c_{d_1(p)}(q)}
\end{equation}

\textbf{Semi-Global Energy} (SGE) \cite{haeusler2013ensemble}, inspired by the the Semi-Global Matching algorithm and computing confidence by summing penalties P1, P2 to the minimum cost as

\begin{equation}\label{sge}
\begin{split}
    \text{SGE} = \sum_{s \in \mathcal{S}} \sum_{q \in s(p)} c_{d_1}(q) &+ P1\cdot t[d_1(q)-d_1(q')= 1] \\ 
    &+ P2\cdot t[d_1(q)-d_1(q') > 1]
\end{split}
\end{equation}
with $s$ being a scanline ray from a set of rays $\mathcal{S}$ emerging from $p$, $s(p)$ a set of pixels along the ray in a local window $N(p)$ and $q'$ the successor of $q$ along the ray. The binary operator $t[\cdot]$ is 1 when the expression holds and 0 otherwise. Thus, P1 and P2 penalize respectively small and larger disparity margins between neighboring pixels along $s$.

\subsection{\colorbox{fullcolor}{The entire cost curve}}

We group here methods considering the entire cost curve, from pixel $p$ and eventually its neighbors.

\textbf{Perturbation measure} (PER) \cite{haeusler2013ensemble}, capturing the deviation of the cost curve to an ideal one as shown in Fig. \ref{fig:costs}
\begin{equation}\label{per}
    \text{PER}(p) = \sum_{i \neq d_1} e^{-\frac{[c_{d_1}(p) - c_i(p)]^2}{s^2}}
\end{equation}

\textbf{Maximum Likelihood Measure} (MLM) \cite{Matthies1192stereo}, as a probability density function for the predicted disparity given the matching costs, by assuming that the cost function follows a normal distribution and that the disparity prior is uniform \cite{hu2012quantitative}
\begin{equation}\label{mlm}
    \text{MLM}(p) = \frac{e^{-\frac{c_{d_1}(p)}{2\sigma}}}{\sum_{i \in D} e^{-\frac{c_{i}(p)}{2\sigma}}}
\end{equation}

\textbf{Attainable Likelihood Measure} (ALM) \cite{Matthies1192stereo}, modeling the cost function in $p$ using a Gaussian distribution centered at $c_{d_1}(p)$, thus making the numerator equal to 1
\begin{equation}\label{alm}
    \text{ALM}(p) = \frac{1}{\sum_{i \in D} e^{-\frac{c_{i}(p)}{2\sigma}}}
\end{equation}

\textbf{Number of Inflections} (NOI) \cite{hu2012quantitative}, as the number of local minima in the cost curve
\begin{equation}\label{noi}
    \text{NOI}(p) = \# \bigcup_{i \in D} c_{i}(p) < c_{i\pm1}(p)
\end{equation}
with $\#$ denoting the cardinality of the set.

\textbf{Local Minima in Neighborhood} (LMN) \cite{kim2014stereo}, as the number of pixels $q$ in a local window $N(p)$ for which costs $c_{d_1(p)}(q)$ are local minima
\begin{equation}\label{lmn}
    \text{LMN}(p) = \# \bigcup_{q \in N(p)} c_{d_1(p)}(q) < c_{d_1(p)\pm1}(q)
\end{equation}

\textbf{Winner Margin} (WMN) \cite{Scharstein1996stereo}, as the difference between the second local minima and the minimum cost, normalized over the entire cost curve
\begin{equation}\label{wmn}
    \text{WMN}(p) = \frac{c_{d_{2m}}(p) - c_{d_1}(p)}{\sum_{i \in D} c_i}
\end{equation}

\textbf{Winner Margin Naive} (WMNN) \cite{hu2012quantitative}, as the difference between the second minima and the minimum cost, normalized over the entire cost curve
\begin{equation}\label{wmnn}
    \text{WMNN}(p) = \frac{c_{d_2}(p) - c_{d_1}(p)}{\sum_{i \in D} c_i}
\end{equation}

\textbf{Negative Entropy Measure} (NEM) \cite{Scharstein1996stereo}, relates the degree of uncertainty to the negative entropy of the minimum matching cost
\begin{equation}\label{nem}
    \text{NEM}(p) = \sum \frac{e^{-c_{d_1}}}{\sum_{i \in D} e^{-c_i}} \cdot \log{\frac{e^{-c_{d_1}}}{\sum_{i \in D} e^{-c_i}}}
\end{equation}

\textbf{Pixel-Wise Cost Function Analysis} (PWCFA) \cite{veld2018novel}, aimed at detecting multiple local minima close to c$_{d_1}$
\begin{equation}\label{pwcfa}
    \text{PWCFA}(p) = \frac{1}{\sum_{i \in D} \frac{\max( \min( |i-d_1|-1, \frac{d_{max}-d_{min}}{3}), 0)^2}{\max(c_i-c_{d_1}- \frac{\sum_{i \in D} c_i}{3(d_{max}-d_{min})} ,1)}}
\end{equation}
In our experiments, being costs normalized in $[0,1]$, we replace $\frac{d_{max}-d_{min}}{3}$ with $\frac{1}{3}$.

\subsection{\colorbox{lrcolor}{Left-right consistency}}

This category evaluates the consistency between corresponding pixels across left and right views according to symmetric (on both left and right views) or asymmetric cues (based on left view only).

\textbf{Left-Right Consistency} (LRC) \cite{Egnal2002}, measures the difference between disparity $d_1$ estimated for $p$ on the left map and disparity $d^r_1$ on the right map for $p^r$. The lower the difference, the higher the confidence
\begin{equation}\label{lrc}
    \text{LRC}(p) = - |d_1(p) - d^r_1(p^r)|
\end{equation}

\textbf{Left-Right Difference} (LRD) \cite{hu2012quantitative}, encodes the margin between the first and second minima on the left disparity map, divided by the difference between minimum costs of corresponding pixels $p,p^r$, respectively on left and right disparity maps
\begin{equation}\label{lrd}
    \text{LRD}(p) = \frac{c_{d_2}(p)-c_{d_1}(p)}{|c_{d_1}(p) - c_{d^r_1}(p^r)|}
\end{equation}

\textbf{Zero-Mean Sum of Absolute Differences} (ZSAD) \cite{haeusler2013ensemble}, as the zero-mean difference in intensities between local windows centered in $p$ and $p^r$
\begin{equation}\label{zsad}
    \text{ZSAD}(p) = \sum_{q \in N(p)} | l(q) - \mu l(p) - r(q^r) + \mu r(p^r) |
\end{equation}

\textbf{Asymmetric Consistency Check} (ACC) \cite{Min2010asymmetric}, checking if any neighbor of $p$, on the same horizontal scanline, collides with it, \ie{} if it matches with the same pixel on the right image. In such a case, low confidence is assigned if $d_1(p)$ is not the maximum hypothesis among colliding hypothesis $d_1(q)$ or if $c_{d_1}(p)$ is not the minimum among costs $c_{d_1}(q)$
\begin{equation}\label{acc}
    \text{ACC}(p) = 
    \begin{cases}
0 & \text{if } p^r \in \bigcup_{q \in Q} q^r \text{ and} \\
  & [d_1(p) \neq \max{q \in Q} d_1(q) \text{ or }\\
  & c_1(p) \neq \min{q \in Q} c_1(q)] \\
1 & \text{otherwise } \\
\end{cases}
\end{equation}
with $Q$ being the set of pixels $q$ having $x$ coordinate varying between $-d_1(p)$ and $(d_{max}-d_1(p))$ around $p$.

\textbf{Uniqueness Constraint} (UC) \cite{UC}, a binary confidence assigning 0 to all colliding pixels, except the one with minimum cost

\begin{equation}\label{uc}
    \text{UC}(p) = 
    \begin{cases}
    0 & \text{if } p^r \in \bigcup_{q \in Q} q^r \text{ and} \\
    & c_1(p) \neq \min{q \in Q} c_1(q)] \\
    1 & \text{otherwise } \\
\end{cases}
\end{equation}

\textbf{Uniqueness Constraint Cost} (UCC) \cite{poggi2017quantitative}, assigning 0 to all colliding pixels, except the one with minimum cost for which the cost itself is assumed as confidence (the lower, the more confident)

\begin{equation}\label{ucc}
    \text{UCC}(p) = 
    \begin{cases}
    0 & \text{if } p^r \in \bigcup_{q \in Q} q^r \text{ and} \\
    & c_1(p) \neq \min{q \in Q} c_1(q)] \\
    -c_{d_1} & \text{otherwise } \\
\end{cases}
\end{equation}

\textbf{Uniqueness Constraint Occurrence} (UCO) \cite{poggi2017quantitative}, as the number of pixels $q$ colliding with $p$ (the lower, the more confident)

\begin{equation}\label{uco}
    \text{UCO}(p) = - \# \bigcup_{q \in Q} p^r = q^r
\end{equation}

\subsection{\colorbox{dispcolor}{Disparity map analysis}}

Confidence measures belonging to this group are obtained by extracting features
from the reference disparity map, with no additional cues from the cost volume.

\textbf{Distance to Discontinuities} (DTD) \cite{spyropoulos2014learning}, as the minimum distance to a depth discontinuity, which often represents a challenge for correct matching

\begin{equation}\label{dtd}
    \text{DTD}(p) = \min_{q \in \hat{d}} |p-q|
\end{equation}
with $\hat{d}$ being obtained by applying an edge detector to the disparity map

\textbf{Disparity Map Variance} (DMV) \cite{haeusler2013ensemble}, defined as the norm of the gradient computed over the disparity map 

\begin{equation}\label{dmv}
    \text{DMV}(p) = ||\nabla d_1(p)||
\end{equation}

\textbf{Variance of disparity} (VAR) \cite{park2015leveraging}, as the statistical variance on a neighborhood $N(p)$. The higher is the variance, the noisier the disparity map is

\begin{equation}\label{var}
    \text{VAR}(p) = -\frac{1}{\#N(p)} \sum_{q \in N(p)} [d_1(q) - \mu(d_1(p)) ]^2
\end{equation}

\textbf{Disparity skewness} (SKEW) \cite{park2018learning}, as the asymmetry on the statistical distribution on a neighborhood $N(p)$. High skewness can identify noisy regions in the disparity map

\begin{equation}\label{skew}
    \text{SKEW}(p) = -\frac{1}{\#N(p)} \sum_{q \in N(p)} [d_1(q) - \mu(d_1(p)) ]^3
\end{equation}

\textbf{Median Disparity Deviation} (MDD) \cite{spyropoulos2014learning}, as the distance from the median disparity (MED) computed over $N(p)$ (the lower, the more confident)

\begin{equation}\label{mdd}
    \text{MDD}(p) = -|d_1(p) - \text{MED}(d_1(p))|
\end{equation}

\textbf{Mean Disparity Deviation} (MND) \cite{park2018learning}, as the distance from the mean disparity computed over $N(p)$ (the lower, the more confident)

\begin{equation}\label{mnd}
    \text{MND}(p) = -|d_1(p) - \mu(d_1(p))|
\end{equation}

\textbf{Disparity Agreement} (DA) \cite{poggi2016learning}, as the number of pixels sharing the same disparity estimate in a local neighborhood (the higher, the more confident)

\begin{equation}\label{da}
    \text{DA}(p) = H[d_1(p)](p)
\end{equation}
with H being the histogram of disparity distribution defined over N(p)

\begin{equation}
    H[i](p) = \# \bigcup_{q \in N(p)} d_1(q) = i
\end{equation}

\textbf{Disparity Scattering} (DS) \cite{poggi2016learning}, encoding the amount of different disparity hypothesis in a local neighborhood (the lower, the more confident)

\begin{equation}\label{ds}
    \text{DS}(p) = - \log \frac{\sum_{i \in D} t[H[i](p) > 0] }{\#N(p)}
\end{equation}
with $t[..]$ being 1 when the expression holds and 0 otherwise.

\subsection{\colorbox{imagecolor}{Reference image analysis}}


Confidence measures belonging to this category use as input domain only the
reference image or some priors

\textbf{Distance from Border} (DB) \cite{spyropoulos2014learning}, encoding the distance from the closest image border, where information is lower

\begin{equation}\label{db}
    \text{DB}(p) = \min (x,y,W-x,H-y)
\end{equation}
with $W,H$ respectively image width and height.

\textbf{Distance from Left Border} (DLB) \cite{park2015leveraging}, as the distance from the left border with $d_{max}$ as upper bound, encoding a portion of the reference image $l$ with no matches on $r$

\begin{equation}\label{dlb}
    \text{DLB}(p) = \min (x,d_{max})
\end{equation}

\textbf{Horizontal Gradient Magnitude} (HGM) \cite{haeusler2013ensemble}, as the horizontal gradient over image intensity. Higher gradients should encode regions rich of texture and easier to be matched

\begin{equation}\label{hgm}
    \text{HGM}(p) = |\nabla_x l(p)|
\end{equation}

\textbf{Distance to image edge} (DTE) \cite{park2015leveraging}, as the minimum distance to an image edge, which often represents a challenge for correct matching

\begin{equation}\label{dte}
    \text{DTE}(p) = \min_{q \in \hat{l}} |p-q|
\end{equation}
with $\hat{l}$ being obtained by applying an edge detector to reference image $l$.

\textbf{Intensity Variance} (IVAR) \cite{park2018learning}, as the statistical variance of pixel intensity on a neighborhood $N(p)$. The higher variance should encode regions rich of texture and easier to be matched

\begin{equation}\label{ivar}
    \text{IVAR}(p) = \frac{1}{\#N(p)} \sum_{q \in N(p)} [l(q) - \mu l(p) ]^2
\end{equation}

\subsection{\colorbox{distcolor}{Self-matching}}

The idea behind these confidence measures is to exploit the notion of distinctiveness of the examined point within its neighborhoods along the horizontal scanline of the same image. To study such a cue, the self-matching between two instances of the same image is performed, \eg{} a cost curve $c^{ll}(p)$ is obtained by running the stereo algorithm on two $l$ images, assuming $D^{ll}=[-d_{max},d_{max}]$ centered on $p$ and symmetric as in \cite{hu2012quantitative}.

\textbf{Distinctiveness} (DTS) \cite{Manduchi1999Distinctiveness}, as the minimum among all costs over $D^{ll}$ range. It encodes the presence of pixels that are very similar to $p$ on the same horizontal scanline

\begin{equation}\label{dts}
    \text{DTS}^l(p) = \min_{i \in D^{ll}} c^{ll}_{i}(p)
\end{equation}

\textbf{Distinctive Similarity Measure} (DSM) \cite{Yoon2008Distinctive}, combining distinctiveness over $l$ and $r$ and considering the similarity between two potentially corresponding pixels

\begin{equation}\label{dsm}
    \text{DSM}(p) = \frac{\text{DTS}^l(p) \cdot \text{DTS}^r(p^r)}{c_{d_1}(p)^2}
\end{equation}

\textbf{Self-Aware Matching Measure} (SAMM) \cite{Mordohai2009}, as the
correlation coefficient between cost curves $c(p)$ and $c^{ll}(p)$

\begin{equation}\label{samm}
    \text{SAMM}(p) = \frac{\sum_{i \in D} [c_{i-d_1}(p) - \mu( c(p))]\cdot [c^{ll}_{i}(p) - \mu( c^{ll}(p))]} {\sigma(p)\cdot\sigma^{ll}(p)}
\end{equation}
with $\sigma$ and $\sigma^{ll}$ being respectively the variance of costs $c(p)$ and $c^{ll}(p)$.

\subsection{\colorbox{sgmcolor}{Semi-Global Matching measures}}\label{sec:sgm}

This family of measures is tailored to the SGM algorithm, considering specific cues available through this pipeline.

\textbf{Sum of Consistent Scanlines} (SCS) \cite{esgm}, as the number of scanline optimizations out of $s$ sharing the same disparity outcome $d^s_1(p)$ of the full SGM algorithm 

\begin{equation}\label{scs}
    \text{SCS}(p) = \# \cup_s d^s_1(p) = d_1(p)
\end{equation}

\textbf{Local-global relationship} (PS) \cite{marin2016reliable}, it studies the relationship between matching costs before and after the semi-global cost aggregation

\begin{equation}\label{ps}
\begin{split}
    \text{PS}(p) = \frac{c_{d_2}^*(p) - c_{d_1}^*(p)}{c_{d_1}^*(p)}\cdot&(1-\frac{\min|d_2^*(p)-d_1^*(p)|,\gamma}{\gamma})\cdot \\
    &(1-\frac{\min|d_1^*(p)-d_1(p)|,\gamma}{\gamma})
\end{split}
\end{equation}

\section{Learned confidence measures}
\label{sec:learning}

The most recent trend in stereo confidence estimation concerns the possibility of \textit{learning} this task directly from data, as in the case of most computer vision problems. We can broadly classify these approaches into two main families: machine learning frameworks and deep learning frameworks. 
In both, we can distinguish between approaches processing or not the cost volume. 

\subsection{Machine learning approaches}

Methods belonging to this category use classifiers, more specifically random forests \cite{breiman2001random}, fed with a subset of the confidence measures reviewed so far to infer a new confidence value. Among these frameworks, we distinguish three main subcategories, respectively processing the cost volume, the disparity map or being specifically designed for SGM algorithm.
In the remainder, we report the composition of the per-pixel features vectors adopted by each proposal, omitting $p$ in the notation for the sake of space.

\subsubsection{\colorbox{forestvol}{Cost-volume forests}}

\hspace{0.3cm} \textbf{Ensemble Learning (23 features)} (ENS23) \cite{haeusler2013ensemble}, the first attempt to infer a confidence estimate by means of machine learning. It combines several hand-crafted measures and features extracted by running the stereo algorithm at multiple resolutions. The main configuration consists into the following features
    $\mathcal{F}(\text{ENS}_{23})$ = 
    $(\text{PKR}^{f,h,q},\text{NEM}^{f,h,q},\text{PER}^{f,h,q},\text{LRC}^f,\text{HGM}^{f,h,q}, \\
    \text{DMV}^{f,h,q},\text{DAM}^{f,h,q},\text{ZSAD}^{f,h,q},\text{SGE}^f)$,
with $^{f,h,q}$ apexes referring to results obtained by running stereo algorithms on $l,r$ at full, half and quarter resolution respectively.

\textbf{Ground Control Points} (GCP) \cite{spyropoulos2014learning,spyropoulos2016correctness}, it proposes a compact feature vector computed at single scale \\
    $\mathcal{F}(\text{GCP})$ = (\text{MSM}, \text{DB}, \text{MMN}, \text{ALM}, \text{LRC}, \text{LRD}, \text{DTD}, \text{MDD})
with MDD being obtained over a $5\times5$ window $N(p)$.

\textbf{Leveraging stereo confidence} (LEV) \cite{park2015leveraging,park2018learning}, it introduces features computed on multiple windows $N(p)$ of increasing size. Two versions with respectively 22 and 50 features have been proposed: 
    $\mathcal{F}(\text{LEV}_{22})$ = 
    (\text{PKR}, \text{PKRN}, \text{MSM}, \text{MM}, \text{WMN}, \text{MLM}, \text{PER}, \text{NEM}, \text{LRD}, \text{LC}, \text{VAR}$^{1,\dots,4}$, \text{DTD}, \text{MDD}$^{1,\dots,4}$, 
    \text{LRC}, \text{HGM}, \text{DLB}) and 
    $\mathcal{F}(\text{LEV}_{50})$ = 
    (\text{MSM}, \text{PKR}, \text{PKRN}, \text{MM}, \text{MMN}, \text{WMN}, 
    \text{WMNN}, \text{MLM}, \text{PER}, \text{NEM}, \text{LRD}, \text{LC}, \text{ALM},
    \text{DTD}, \text{DTE}, \text{LRC}, \text{HGM}, \text{DLB}, \text{DB}, \text{NOI}, 
    \text{VAR}$^{1,3,4,6,9,14}$, \text{MDD}$^{1,3,4,6,9,14}$, \text{MND}$^{1,3,4,6,9,14}$, \text{SKEW}$^{1,3,4,6,9,14}$, \text{IVAR}$^{1,3,4,6,9,14})$.
Features with apex $^i$ are computed on $(3 + 2i)\times(3 + 2i)$ windows, \eg{} MDD$^1$ is computed over a $5\times5$. We replace image gradients with HGM, achieving slightly better results.
The authors also propose a method to select the most important features and reduce the vector dimensionality. In our evaluation, we consider the complete vectors, being them the best performing.

\textbf{Feature Augmentation} (FA)~\cite{kim2017feature}. Unlike previous methods that predict the confidence based on per-pixel features, FA~\cite{kim2017feature} imposes a spatial consistency on the confidence estimation by introducing a robust set of features extracted from super-pixels, $\mathcal{F}(\text{FA1})$ = (\text{LRC}, \text{DB}, \text{LRD}, \text{MDD}$^{1,2,3}$, \text{MLM}, \text{MSM}) and $\mathcal{F}(\text{FA2})$ = (\text{LRD}, \text{PKRN}, \text{MDD}$^{1,2,3,4}$, \text{MLM}, \text{NEM}), which are concatenated with per-pixel features and enhanced through adaptive filtering.

\subsubsection{\colorbox{forestdisp}{Disparity forests}}
\hspace{0.3cm} \textbf{Ensemble Learning (7 features)} (ENS7) \cite{haeusler2013ensemble}, a variant of ENS23 extracting seven features from the disparity map and the reference image, resulting in  
$\mathcal{F}(\text{ENS}_7)$ = (\text{LRC}$^f$,\text{HGM}$^{f,h,q}$,\text{DMV}$^{f,h,q}$).

\textbf{O(1) Features} (O1) \cite{poggi2016learning,poggi2020learning}, 
these methods aim at learning to infer a confidence score only from features that can be computed in constant time from the reference disparity map domain, thus not requiring the cost volume. Two version with respectively 20 \cite{poggi2016learning} and 47 \cite{poggi2020learning} features have been proposed:
    $\mathcal{F}(\text{O1})$ = (\text{DA}$^{1,\dots,4}$, \text{DS}$^{1,\dots,4}$, \text{MED}$^{1,\dots,4}$,
    \text{MDD}$^{1,\dots,4}$, \text{VAR}$^{1,\dots,4}$)
and
    $\mathcal{F}(\text{O2})$ = (\text{DA}$^{1,\dots,9}$, \text{DS}$^{1,\dots,9}$, \text{MED}$^{1,\dots,9}$, \text{MDD}$^{1,\dots,9}$, \text{VAR}$^{1,\dots,9}$, \text{DLB}, \text{UC})

\subsubsection{\colorbox{sgmcolor}{SGM-specific forest}}
\hspace{0.3cm} \textbf{SGMForest} (SGMF) \cite{schonberger2018SGMForest}, suited for the SGM algorithm, it consider the disparities $d^s_1$ selected by each single scanline $s \in \mathcal{S}$ and their cost $c^z_{d^s_1}$ for each scanline $z \in \mathcal{S}$

\begin{equation}
    \text{SGMF} = (\bigcup_{s \in \mathcal{S}}d^{s}_1, \bigcup_{(s,z) \in \mathcal{S}\times\mathcal{S} } c^z_{d^s_1})
\end{equation}
Originally proposed to improve SGM by selecting the most reliable scanline for each pixel, we recast it to infer a confidence value for the SGM algorithm.

\subsection{Deep learning approaches}

This latter family groups methods leveraging on CNNs to infer confidence maps. Conversely from previous machine learning approaches, these techniques directly process the input cues, \ie{} reference image, cost volume and disparity maps, without explicit features extraction. In this case we define two subcategories, respectively processing the disparity map as main cue or the cost volume as well.

\subsubsection{\colorbox{deepdisp}{Disparity CNNs}}
\hspace{0.3cm} \textbf{Confidence CNN} (CCNN) \cite{poggi2016bmvc}, a patch-based CNN processing $9\times9$ patches from the reference disparity map only. A full confidence map can be processed in a single forward pass by means of a fully-convolutional design.

\textbf{Patch Based Confidence Prediction} (PBCP) \cite{seki2016patch}, a patch-based network jointly processing $15\times15$ patches from both reference and target disparity map. This latter is warped according to the former. Two versions exist, trading-off accuracy for speed: one for which pixels in the patches are normalized according to the central pixel disparity (\textit{disposable}), for which each patch needs to be process independently, and one for which normalization is turned off (\textit{reusable}), allowing for a single inference on the full-resolution disparity map. 

\textbf{Early Fusion Network} (EFN) \cite{Fu_2017_RFMI}, extending CCNN by processing the input reference image together with the disparity map. In this variant, $9\times9$ image and disparity patches are concatenated and fed to a single features extractor.

\textbf{Late Fusion Network} (LFN) \cite{Fu_2017_RFMI}, combining image and disparity  as EFN does, but processing the two $9\times9$ patches by means of two distinct features extractor, then concatenating the resulting features before confidence estimation. 

\textbf{Multi Modal CNN} (MMC) \cite{fu2018learning}, extending the late fusion model proposed in \cite{Fu_2017_RFMI}. In particular, $15\times15$ patches from the two modalities are processed by two different encoders for disparity and RGB, the latter using dilated convolutions to enlarge the receptive field. 

\textbf{Global Confidence Network} (ConfNet) \cite{tosi2018beyond}, deploying an U-Net like architecture with larger receptive field in order to include larger content from both the image and disparity map. This network decimates the input resolution by means of max-pool operations, then restoring it by means of transposed convolutions in the decoding part.

\textbf{Local-Global Confidence Network} (LGC) \cite{tosi2018beyond}, combines patch-based methodologies \cite{poggi2016bmvc} with ConfNet allowing to reason for both local and global cues at once, combining the fine-grained features extracted by the former with the large image context of the latter.

\subsubsection{\colorbox{deepvol}{Cost-volume CNNs}}
\hspace{0.3cm} \textbf{Reflective Confidence Network} (RCN)~\cite{Shaked_2017_CVPR} proposes to jointly estimate a confidence measure together with cost optimization at the end of the stereo matching pipeline. By deploying a two-layer fully connected  network processing the matching costs, a confidence map is predicted together with the final disparity map.
	
\textbf{Matching Probability Network} (MPN)~\cite{Kim_2017_ICIP} processes the matching cost volume together with the disparity map, through a novel network consisting of cost feature extraction, disparity feature extraction, and fusion modules. To deal with a varying size of cost volume according to stereo pairs, a top-$K$ matching probability volume layer is also proposed in the cost feature extraction module.
	
\textbf{Unified Confidence Network} (UCN)~\cite{kim2019unified}. Similar to RCN~\cite{Shaked_2017_CVPR}, it is also based on the observation that jointly learning cost optimization and confidence estimation is effective at improving the accuracy of the final disparity map of a stereo matching pipeline. UCN~\cite{kim2019unified} proposes a unified network architecture for cost optimization and confidence estimation. An encoder-decoder module refines the matching costs with a larger receptive field in order to obtain a more accurate disparity map. Then a subnetwork processes it together with top-$K$ refined costs to output a confidence map.
	
\textbf{Locally Adaptive Fusion Network} (LAF)~\cite{Kim_2019_CVPR} estimates a confidence map of an initial disparity by making full use of tri-modal input, including cost, disparity, and color image. A key element is to learn locally-varying attention and scale maps to fuse the tri-modal confidence features. In addition, the confidence map is recursively refined to enforce a spatial context and local consistency. 
	
\textbf{Adversarial Confidence Network} (ACN)~\cite{Kim2020adversarial}. Similar to RC~\cite{Shaked_2017_CVPR} and UN~\cite{kim2019unified}, it jointly estimates disparity and confidence from stereo image pairs. Especially, ACN~\cite{Kim2020adversarial} accomplishes this via a minmax optimization to learn the generative cost aggregation networks and discriminative confidence estimation networks in an adversarial manner. 
To fully exploit complementary information of cost, disparity, and color image, a dynamic fusion module is also proposed.
	
\textbf{Pixel-Wise Confidence RNN} (CRNN)~\cite{gul2019pixel} is the first attempt to use a recurrent neural network architecture to compute confidences. To maintain a low complexity, the confidence for a given pixel is purely computed from its associated costs without considering any additional neighbouring pixels.
	
\textbf{Cost Volume Analysis Network} (CVA)~\cite{mehltretter2019cnn}. In order to combine the advantages of deep learning and cost volume features, it directly learns features for estimating confidence  from the volumetric data. Specifically, CVA~\cite{mehltretter2019cnn} first fuses a cost volume into a single cost curve using 3D convolutions, and the curve is then processed along the disparity axis by other 3D convolutions with varying depth.

\subsection{Others}

For completeness, we report techniques aimed at improving the effectiveness of pre-computed confidence maps, although not directly evaluating them in this paper. 

\textbf{Learning Local Consistency} (++) \cite{poggi2017learning}. This framework learns a more reliable measure exploiting local consistency within neighboring points by processing a pre-computed confidence map by means of a patch-based CNN. 

\textbf{Even More Confident} (EMC) \cite{poggi2017even}. In this framework, random forest based measures are improved by replacing the ensemble classifier with a patch-based CNN.

\begin{table*}[t]
    \centering
    \scalebox{0.6}{
    \renewcommand{\tabcolsep}{2pt}
    
    \begin{tabularx}{2\textwidth}{ccc}
    \begin{tabular}{.l;l;l.}
    \toprule
    Measure & Acronym & Definition\\
    \midrule
    \rowcolor{localcolor}
    Average Peak Ratio \cite{kim2014stereo}& APKR & Eq. \ref{apkr}\\
    Average Peak Ratio Naive \cite{poggi2017quantitative} & APKRN & Eq. \ref{apkrn}\\
    Curvature \cite{Egnal2004} & CUR & Eq. \ref{cur} \\
    Disparity Ambiguity Measure \cite{haeusler2013ensemble} & DAM & Eq. \ref{dam}  \\
    Local Curve \cite{Wedel2009detection} & LC & Eq. \ref{lc}\\
    Maximum Margin \cite{poggi2017quantitative} & MM & Eq. \ref{mm}\\
    Maximum Margin Naive \cite{hu2012quantitative} & MMN & Eq. \ref{mmn}\\
    Matching Score Measure \cite{Egnal2004} & MSM & Eq. \ref{msm}\\
    Non-Linear Margin \cite{Haeusler2012evaluation} & NLM & Eq. \ref{nlm}\\
    Non-Linear Margin Naive \cite{poggi2017quantitative} & NLMN & Eq. \ref{nlmn}\\
    Peak Ratio \cite{Egnal2004} & PKR & Eq. \ref{pkr}\\
    Peak Ratio Naive \cite{hu2012quantitative} & PKRN & Eq. \ref{pkrn}\\
    Semi-Global Enery \cite{haeusler2013ensemble} & SGE & Eq. \ref{sge}\\
    Weighted Peak Ratio \cite{kim2016weighted} & WPKR & Eq. \ref{wpkr}\\
    Weighted Peak Ratio Naive \cite{poggi2017quantitative} & WPKRN & Eq. \ref{wpkrn}\\
    \rowcolor{fullcolor}
    Attainable Likelihood Measure \cite{Matthies1192stereo} & ALM & Eq. \ref{alm}\\
    Local Minima in Neightrhood \cite{kim2014stereo} & LMN & Eq. \ref{lmn}\\
    Maximum Likelihood Measure \cite{Matthies1192stereo} & MLM & Eq. \ref{mlm}\\
    Negative Entropy Measure \cite{Scharstein1996stereo} & NEM & Eq. \ref{nem}\\
    Number of Inflections \cite{hu2012quantitative} & NOI & Eq. \ref{noi}\\
    Perturbation measure \cite{haeusler2013ensemble}& PER  & Eq. \ref{per}\\
    Pixel-Wise Cost Function Analysis \cite{veld2018novel} & PWCFA & Eq. \ref{pwcfa}\\
    Winner Margin \cite{Scharstein1996stereo} & WMN & Eq. \ref{wmn}\\
    Winner Margin Naive \cite{hu2012quantitative} & WMNN & Eq. \ref{wmnn}\\        
    \rowcolor{sgmcolor}
    Local-global relationship \cite{marin2016reliable} & PS & Eq. \ref{ps}\\    
    \midrule
    \end{tabular}
    &
    \begin{tabular}{.l;l;l.}
    \toprule
     Measure & Acronym & Definition\\
    \midrule
    \rowcolor{dispcolor}
    Disparity Agreement \cite{poggi2016learning} & DA  & Eq. \ref{da}\\
    Disparity Scattering \cite{poggi2016learning} & DS  & Eq. \ref{ds}\\
    Disparity Map Variance \cite{haeusler2013ensemble} & DMV  & Eq. \ref{dmv}\\
    Distance To Discontinuities \cite{spyropoulos2014learning} & DTD  & Eq. \ref{dtd}\\
    Median Disparity Deviation \cite{spyropoulos2014learning} & MDD  & Eq. \ref{mdd}\\
    Mean Disparity Deviation \cite{park2018learning} & MND  & Eq. \ref{mnd}\\
    Disparity skewness \cite{park2018learning} & SKEW  & Eq. \ref{skew}\\
    Disparity Variance \cite{park2015leveraging} & VAR  & Eq. \ref{var}\\
    \rowcolor{lrcolor}
    Asymmetric Consistency Check \cite{Min2010asymmetric} & ACC  & Eq. \ref{acc}\\
    Left-Right Consistency \cite{Egnal2002} & LRC  & Eq. \ref{lrc}\\
    Left-Right Difference \cite{hu2012quantitative} & LRD  & Eq. \ref{lrd}\\
    Uniqueness Constraint \cite{UC} & UC  & Eq. \ref{uc}\\
    Uniqueness Constraint (Cost) \cite{poggi2017quantitative} & UCC  & Eq. \ref{ucc}\\
    Uniqueness Constraint (Occurrence) \cite{poggi2017quantitative} & UCO  & Eq. \ref{uco}\\
    Zero-Mean Sum of Absolute Differences \cite{haeusler2013ensemble} & ZSAD  & Eq. \ref{zsad}\\
    \rowcolor{distcolor}
    Distinctiveness \cite{Manduchi1999Distinctiveness} & DTS & Eq. \ref{dts}\\
    Distinctive  Similarity  Measure \cite{Yoon2008Distinctive} & DSM & Eq. \ref{dsm} \\
    Self-Aware Matching Measure \cite{Mordohai2009} & SAMM  & Eq. \ref{samm}\\
    \rowcolor{imagecolor}
    Distance from Border \cite{spyropoulos2014learning} & DB & Eq. \ref{db} \\
    Distance from Left Border \cite{park2015leveraging} & DLB  & Eq. \ref{dlb}\\
    Distance to image Edge \cite{park2015leveraging} & DTE  & Eq. \ref{dte}\\
    Horizontal Gradient Magnitude \cite{haeusler2013ensemble} & HGM  & Eq. \ref{hgm}\\
    Intensity Variance \cite{park2018learning} & IVAR & Eq. \ref{ivar} \\
    & & \\
    \rowcolor{sgmcolor}
    Sum of Consistent Scanlines \cite{esgm} & SCS & Eq. \ref{scs}\\        
    \midrule
    \end{tabular}   
    &
    \begin{tabular}{.l;l;l;l;l;l;l.}
    \toprule
    Measure & Acronym & Forest & CNN & Image & Volume & Disparity\\
    \midrule
    \rowcolor{forestvol}
    Ensemble Learning (23 features) \cite{haeusler2013ensemble} & ENS$_{23}$ & \cmark & & \cmark & \cmark & \cmark\\
    Ground Control Points \cite{spyropoulos2014learning} & GCP & \cmark & &  \cmark & \cmark & \cmark\\
    Leveraging stereo confidence (22 features) \cite{park2015leveraging} & LEV$_{22}$ & \cmark & & \cmark & \cmark & \cmark\\
    Leveraging stereo confidence (50 features) \cite{park2018learning} & LEV$_{50}$ & \cmark & & \cmark & \cmark & \cmark\\
    Feature augmentation \cite{kim2017feature} & FA & \cmark & &  \cmark & \cmark & \cmark \\
    \rowcolor{forestdisp}
    Ensemble Learning (7 features) \cite{haeusler2013ensemble} & ENS$_{7}$ & \cmark & & \cmark & & \cmark\\
    O(1) (20 features) \cite{poggi2016learning} & O1 & \cmark & & & & \cmark\\
    O(1) (47 features) \cite{poggi2020learning} & O2 & \cmark & & \cmark & & \cmark\\
    \rowcolor{deepdisp}
    Confidence CNN \cite{poggi2016bmvc} & CCNN & & \cmark & & &  \cmark\\
    Patch-based confidence prediction (reusable) \cite{seki2016patch} & PBCP$_{r}$ & & \cmark & & & \cmark \\
    Patch-based confidence prediction (disposable) \cite{seki2016patch} & PBCP$_{d}$ & & \cmark & & & \cmark \\
    Early Fusion Network \cite{Fu_2017_RFMI} & EFN & & \cmark & \cmark & & \cmark \\
    Late Fusion Network \cite{Fu_2017_RFMI} & LFN & & \cmark & \cmark & & \cmark \\
    Multi Modal CNN \cite{fu2018learning} & MMC & & \cmark & \cmark & & \cmark \\
    Global Confidence Network \cite{tosi2018beyond} & ConfNet & & \cmark & \cmark & & \cmark \\
    Local-Global Network \cite{tosi2018beyond} & LGC & & \cmark & \cmark & & \cmark \\
    \rowcolor{deepvol}
    Reflective Confidence Network \cite{Shaked_2017_CVPR} & RCN & & \cmark & & \cmark & \\
    Matching Probability Network \cite{Kim_2017_ICIP} & MPN & & \cmark & & \cmark & \cmark \\
    Unified Confidence Network \cite{kim2019unified} & UCN & & \cmark & & \cmark & \cmark\\
    Locally Adaptive Fusion Network \cite{Kim_2019_CVPR} & LAF & & \cmark & \cmark & \cmark & \cmark\\
    Adversarial Confidence Network \cite{Kim2020adversarial} & ACN & & \cmark & \cmark & \cmark & \cmark\\
    Pixel-Wise Confidence RNN \cite{gul2019pixel} & CRNN & & \cmark & & \cmark & \\
    Cost Volume Analysis Network \cite{mehltretter2019cnn} & CVA & & \cmark & & \cmark & \\
    & & \multicolumn{5}{c.}{}\\
    \rowcolor{sgmcolor}
    SGMForest \cite{schonberger2018SGMForest} & SGMF & \cmark & & & \cmark & \\
    \midrule
    \end{tabular}  \\
    \end{tabularx}
    }
    \caption{ \textbf{Taxonomy of confidence measures.} Different colors encode different categories. For each measure, we report its full name, reference paper and acronym. For hand-crafted measures, We point to their definition. For learned measures, we highlight the type of classifier and its input cues.}
    \label{tab:acronyms}
\end{table*}

\section{Experimental results}
\label{sec:results}

In this section, we introduce the reader to our experimental evaluation by describing each dataset and stereo algorithm involved, as well as the evaluation metrics.

\subsection{Evaluated measures}

We collect the names, acronyms and definition of each of the measures classified in our taxonomy and involved in our evaluation in Table \ref{tab:acronyms}. Measures belonging to the same category are grouped in blocks colored according to the category and listed in alphabetical order. For hand-crafted measures, we point to equations detailing their definition. For learned measures (right-most in the table), we highlight the classifier they use and the input cues they process. The same table structure will be used when evaluating the measures on the different datasets and stereo algorithms.

\subsection{Datasets}

We describe in detail the datasets on which our evaluation is carried out. Since ground truth disparity is required to assess the performance of confidence estimation, we refer to the training sets made available by each dataset.

\textbf{SceneFlow Driving.} The Freiburg SceneFlow dataset \cite{Mayer_2016_CVPR} is a large collection of synthetic images, made of about 39K stereo pairs with ground truth disparity maps. We run experiments on this dataset, aiming in particular at studying the impact of domain shifts on the confidence estimation task for the first time in literature. Purposely, we sample a training set made of 22 stereo pairs from the \textit{backwards} sequences Driving split, since learned-based approaches require very few images for training \cite{poggi2016bmvc,seki2016patch}. We also collect a testing set made of 22 images from \textit{forward} sequences, thus non-overlapping with those from which the training images are sampled.

\textbf{\kitti\ 2012.} An outdoor dataset, acquired from static scenes in a driving environment. It is composed of 194 grayscale stereo pairs, recently made available in color format as well. Sparse ground truth disparity was obtained from LIDAR measurements, post-processed by registering a set of consecutive frames (5 before and 5 after) with ICP, then re-projecting accumulated point clouds onto the image and finally manually filtering all ambiguous depth values.
We manually split them into 20 training images and keep the remaining 174 for testing following \cite{poggi2017quantitative}, in order to allow for training learned measures on real data as well.

\textbf{\kitti\ 2015.} Improved with respect to \kitti\ 2012 and thought for scene flow evaluation, this dataset frames dynamic scenes in driving environments. It is composed of 200 color stereo pairs for which sparse ground truth disparity was obtained with a similar procedure, except for moving objects that were replaced by 3D CAD models (\eg, in the case of cars) fitted into accumulated point clouds and re-projected onto the image and manually filtered.

\textbf{\midd.} An indoor dataset, made of 15 stereo pairs reaching up to 6 megapixels resolution. Dense ground truth maps are obtained by means of an active stereo pipeline \cite{scharstein2014high}. It represents an open challenge for most stereo algorithms, either hand-crafted or based on deep learning. In our experiments, we process quarter resolution images as in previous works \cite{poggi2017quantitative}.

\textbf{\eth.} One of the most recent among real-world datasets, made of 27 low-resolution grayscale stereo pairs.
To obtain ground truth disparities, the authors recorded the scene geometry with a Faro Focus X 330 laser scanner, taking one or more 360$^\circ$ scans with up to 28 million points each. We evaluate confidence measures on this dataset for the first time in literature.

\subsection{Evaluation metrics}
We measure the effectiveness of each confidence measure at detecting correct matches, as introduced in \cite{hu2012quantitative}. To this aim, we sort pixels in a disparity map following decreasing order of confidence and gradually compute the error rate (D1) on sparse maps obtained by iterative sampling (\eg, 5\% of pixels each time) from the dense map. D1 is computed as the percentage of pixels having absolute error larger than $\tau$. Plotting the error rates results in a ROC curve, whose AUC quantitatively assesses the confidence effectiveness (the lower, the better). Optimal AUC is obtained if the confidence measure is capable of sampling all correct matches first and is equal to:

\begin{equation}
    \text{AUC}_\text{opt} = \int_{1-\varepsilon}^1 \frac{x - (1-\varepsilon)}{x} dx = \varepsilon + (1-\varepsilon)\ln{(1-\varepsilon)}
\end{equation}
with $\varepsilon$ being the D1 computed over the disparity map. 
To have a view over an entire dataset, we compute macro-average AUC scores over the total number of images.
To ease readability, we report each AUC score, together with optimal AUCs, multiplied by a factor $\times10^2$. In all the experiments, we set $\tau$ to 3 for Driving, \kitti\ 2012 and \kitti\ 2015 datasets, to 1 for \midd\ and \eth. 
According to \cite{hu2012quantitative}, we may also define the AUC for the random chance (i.e., assuming no knowledge about pixels confidence). This is equal to the D1 itself, since no correct matches can be selected in absence of confidence information. Every time the AUC achieved by a confidence measure is lower than the D1, it means it is somehow useful for selection with respect to the random choice.

For each stereo algorithm, we will report AUC for the five considered datasets. We will also report the ranking (R.) for each confidence measure according to its average performance over them. Concerning measures computed over a local window, we report in the table the top performing configuration, while we show the behavior of each of them by varying the window size in form of plots.

Concerning learned measures, we report results in two main configuration: 1) when trained on the Driving train split and 2) when trained on the 20 KITTI 2012 stereo pairs, on left and right columns in a single table. In the former case, we rank measures both according to their performance on synthetic data (R.) and their cross-domain ranking (CR.) on real data averaging over the four real datasets. In the latter case, we rank measures according to performance on the real domain (R.).

\subsection{Stereo Algorithms}

We measure the effectiveness of each confidence measure when dealing with the output of four different stereo algorithms, ranging from noisier to more robust, as well as on a deep stereo network. The four hand-crafted pipelines are obtained selecting among two matching costs and two aggregation strategies, described in detail in the remainder. 

\subsubsection{Matching cost functions}

The very first step in a stereo pipeline consists into computing per-pixel matching costs. To this aim, we selected two popular choices, AD-CENSUS and MCCNN-fst.

\textbf{AD-CENSUS.} A robust matching function based on the census transform \cite{Zabih_1994_ECCV}. For both left and right images, pixels intensities are replaced by 81 bits strings, computed by cropping a $9\times9$ image patch centered around a given pixel and comparing the intensity values of each neighbor in the patch to the intensity value of the pixel in the center. 
Then, the absolute distance between pixels is computed in form of the Hamming distance between bits strings.

\textbf{MCCNN-fst.} In this case, matching costs are inferred by a deep neural network \cite{zbontar2016stereo} trained to compare image patches and estimate a similarity score between the two. We use the MCCNN-fst variant, because it is much faster, although almost equivalent to the accurate one MCCNN-acrt.
We use weights made available by the authors and respectively trained on \kitti\ 2012, \kitti\ 2015 and \midd\ for the corresponding datasets. We used weights trained on \midd\ to run experiments on \eth\ as well, while we trained from scratch a model on the Driving train split for experiments on the same dataset test split. 

\subsubsection{Aggregation strategies}

Given an initial cost volume, the aggregation step aims at reducing noise and ambiguity in the cost curves. According to the strategy deployed, stereo algorithms are usually classified into local and global \cite{scharstein2002taxonomy}. We select two main approaches representative of the two worlds, Cross-based Cost Aggregation (CBCA) and SGM. For both, the source code and parameters as defined in \cite{zbontar2016stereo} are used in our experiments.

\textbf{CBCA.} An adaptive, local aggregation strategy. Given a pixel, it builds a support window over a cross \cite{zhang2009cross} including neighbors for which both spatial distance and intensity difference are lower than two respective thresholds. Supports regions $U_l,U_r$ are computed over  $I_l, I_r$ and combined as 

\begin{equation}
    U_d(p) = \{q | q \in U_l(p), (q - d) \in U_r(p - d)\}
\end{equation}
Then, initial costs $C_0(p,d)$ sharing the same disparity hypothesis $d$ are summed over the support region $U_d(p)$ to obtain aggregated costs $C_\text{CBCA}(p,d)$. 

\textbf{SGM.} A semi-global aggregation strategy \cite{hirschmuller2005accurate} combining multiple scanline optimizations. For each, smoothness is enforced by means of two penalties P1 and P2, starting from locally aggregated costs by means of CBCA, as follows:

\begin{equation}\label{eq:sgm}
\begin{split}
C_s(p,d) = &C_\text{CBCA}(p,d) + \min_{o>1}[C_\text{CBCA}(q,d), \\
&C_\text{CBCA}(q,d\pm1)+P1, C_\text{CBCA}(q,d\pm o)+P2] -\\ &\min_{k<d_{max}}(C(q,k)) \\
\end{split}
\end{equation}
The outcome $C_s$ over each scanline $s$ is then summed to obtain the final cost volume $C_\text{SGM}$. Four paths are considered, along horizontal and vertical directions.

\subsubsection{End-to-end stereo}

Confidence measures have always been studied in synergy with hand-crafted stereo algorithms, but nowadays end-to-end deep networks represent the preferred choice to infer dense disparity maps. Thus, for the first time in literature, we deeply investigate about confidence estimation in the case of deep stereo networks. 

\textbf{GANet} \cite{Zhang2019GANet}. A state-of-the-art 3D architecture whose output is a feature volume $\mathcal{C}$ of size $D\times H\times W$ similar to the cost volume processed by hand-crafted stereo algorithms, from which disparity is selected by means of soft-argmax

\begin{equation}
    d = \sum_{i \in D} i \cdot \mathcal{C}_i(p)
\end{equation}
Accordingly, $\mathcal{C}$ encodes matching probabilities.
In our experiments, we convert $\mathcal{C}$ into matching costs by multiplying for $-1$ and compute disparity by replacing the soft-argmax operation with a traditional WTA selection during disparity inference at testing time. 
This way, all confidence measures can be applied seamlessly as done with hand-crafted algorithms, being disparity selected from the minimum cost. Table \ref{tab:softmax_vs_wta} shows how the WTA selection impacts on disparity accuracy compared to soft-argmax. In general, WTA selection seems better when assuming higher threshold $\tau$, such as on Driving, KITTI 2012 and KITTI 2015. On the other hand, the subpixel accuracy enabled by the soft-argmax strategy allows to improve the error rate when considering $\tau=1$, as in Middlebury and ETH3D dataset.

\begin{table}[]
    \centering
    \renewcommand{\tabcolsep}{2pt}
    \begin{tabular}{c|ccccc}
    \toprule
    & Driving & KITTI 2012 & KITTI 2015 & Middlebury & ETH \\
    & (bad3) & (bad3) & (bad3) & (bad1) & (bad1) \\
    \midrule
    soft-argmax & 17.65\% & 9.51\% & 10.77\% & \textbf{26.89\%} & \textbf{8.73\%} \\
    WTA & \textbf{16.66\%} & \textbf{8.47\%} & \textbf{10.02\%} & 28.61\% & 10.80\% \\
    \toprule
    \end{tabular}
    \caption{\textbf{GANet disparity map accuracy}, with different selection strategies.}
    \label{tab:softmax_vs_wta}
\end{table}

For our experiments, we use the weights made available by the authors trained on SceneFlow to avoid over-fitting to any real dataset and simulate deployment in-the-wild.

\subsection{Hyper-parameters, training setup, implementation.}

In this section, we resume implementations details and parameters tuning for both hand-crafted and learned confidence methods, referring to existing works the sake of space.

Concerning hand-crafted measures, all hyper-parameters have been set following our previous work \cite{poggi2017quantitative}\footnote{\url{http://vision.deis.unibo.it/~mpoggi/code/ICCV2017.zip}}. To study the impact of the local windows over confidence measures exploiting local content, we considered the following window sizes, already used in previous works \cite{park2015leveraging,park2018learning,poggi2016learning,poggi2020learning}: $5\times5$, $7\times7$, $9\times9$, $11\times11$, $13\times13$, $15\times15$, $17\times17$, $19\times19$, $21\times21$ and $31\times31$.

Concerning learning-based measures, we follow the authors training settings, using the original source code when available\footnote{\url{https://github.com/fabiotosi92/LGC-Tensorflow}}\footnote{\url{https://github.com/seungryong/LAF}}. For methods for which the source code has not been released, we ran experiments using our own code, implementing each approach following the authors' advice at the best of our knowledge. 


\subsection{CBCA Algorithms}

We start by evaluating the performance on local stereo algorithms leveraging CBCA cost optimization. Although they rarely are the final source of disparity maps, several works \cite{park2015leveraging,park2018learning,poggi2016learning,poggi2020learning,seki2016patch} proposed improved SGM variants exploiting the confidence estimated over intermediate results, often coming from CBCA methods.
This makes the evaluation of confidence in this setting valuable as well.

In the remainder, all results will be collected in tables, where each entry is colored differently to recall the aforementioned classes of measures.

\subsubsection{CENSUS-CBCA}

In this section, we discuss the outcome of our experiments carried out with Census-CBCA algorithm.

\textbf{Hand-crafted measures.} Table \ref{tab:census_cbca_hand} shows the performance achieved by the hand-crafted measures, \ie not involving machine learning at all.
Among them, the top-3 measures are DA$_{31}$, VAR$_{9}$ and APKR$_{7}$, that are computed over a local window. This suggests that the local context, either from the disparity domain or the cost volume, can be a powerful cue to estimate the per-pixel confidence. The first measures using single pixel information are LRD and PKR. The top-9 measures, except LRD, belong to the \colorbox{localcolor}{local properties} or \colorbox{dispcolor}{disparity domain} families, with WMN and WMNN ranking 11 and 12 and being the first measures using the \colorbox{fullcolor}{entire cost curve}. Confidence estimated from \colorbox{lrcolor}{left-right consistency}, after finding LRD at rank 5, only appears at rank 25 with UCC, performing better than LRC on the noisy outputs of Census-CBCA. \colorbox{distcolor}{Self-matching} measures show up at 26 and 28 positions with SAMM and DSM, while \colorbox{imagecolor}{image properties} produce, not surprisingly, poor results.

\begin{table}[t]
    \centering
    \scalebox{0.68}{
    \renewcommand{\tabcolsep}{2pt}
    \begin{tabularx}{\textwidth}{cc}
    
    \begin{tabular}{.l;c|c|c|c|c;c.}
    \toprule
    & Driv. & 2012 & 2015 & Midd. & ETH & R.\\
    \midrule
    \rowcolor{localcolor}
    APKR$_{7}$ & 20.64 & 9.32 & 7.78 & 12.54 & 8.33 & 3\\
    APKRN$_{5}$ & 25.95 & 11.15 & 9.79 & 11.91 & 8.74 & 14\\
    CUR & 33.33 & 19.94 & 14.71 & 14.51 & 10.72 & 33\\
    DAM & 30.56 & 17.81 & 15.97 & 22.07 & 16.13 & 36\\
    LC & 31.45 & 18.63 & 14.29 & 14.24 & 10.66 & 32\\
    MM & 21.50 & 10.46 & 8.83 & 12.14 & 8.57 & 7\\
    MMN & 28.51 & 13.15 & 11.68 & 12.31 & 9.47 & 20\\
    MSM & 22.28 & 17.12 & 15.06 & 17.61 & 15.30 & 30\\
    NLM & 21.50 & 10.46 & 8.83 & 12.15 & 8.57 & 8\\
    NLMN & 28.51 & 13.16 & 11.68 & 12.31 & 9.47 & 21\\
    PKR & 20.85 & 10.55 & 8.90 & 12.40 & 8.64 & 6\\
    PKRN & 25.66 & 11.83 & 10.32 & 11.42 & 8.84 & 15\\
    SGE & 22.06 & 16.98 & 14.97 & 17.81 & 15.41 & 29\\
    WPKR$_{5}$ & 21.96 & 10.16 & 8.63 & 12.49 & 8.51 & 9\\
    WPKRN$_{5}$ & 26.95 & 12.93 & 11.63 & 12.58 & 9.18 & 18\\
    \rowcolor{fullcolor}
    ALM & 20.76 & 14.77 & 13.03 & 16.20 & 12.79 & 24\\
    LMN & 39.19 & 26.79 & 19.57 & 22.43 & 15.85 & 39\\
    MLM & 20.23 & 13.31 & 11.58 & 14.67 & 11.70 & 17\\
    NEM & 30.99 & 24.75 & 21.30 & 29.93 & 20.55 & 44\\
    NOI & 33.15 & 25.56 & 21.36 & 28.15 & 18.74 & 42\\
    PER & 20.70 & 14.59 & 12.84 & 15.98 & 12.65 & 23\\
    PWCFA & 20.72 & 12.91 & 11.63 & 14.01 & 11.21 & 16\\
    WMN & 20.94 & 11.25 & 9.45 & 12.63 & 9.07 & 11\\
    WMNN & 24.64 & 11.73 & 10.15 & 11.33 & 9.11 & 12\\
    \midrule
    \rowcolor{white}
    Opt. & 12.00 & 4.72 & 3.40 & 5.31 & 4.07 & -\\
    D1(\%) & 43.58 & 27.19 & 22.28 & 28.70 & 21.27 & -\\
    \midrule
    \end{tabular}
    &
    \begin{tabular}{.l;c|c|c|c|c;c.}
    \toprule
    & Driv. & 2012 & 2015 & Midd. & ETH & R. \\
    \midrule
    \rowcolor{dispcolor}
    DA$_{31}$ & 23.12 & \underline{7.95} & 6.45 & 12.92 & \underline{6.07} & 1\\
    DMV & 25.91 & 11.46 & 9.26 & 18.45 & 14.40 & 27 \\
    DS$_{17}$ & 21.85 & 9.27 & 7.62 & 12.27 & 8.05 & 4\\
    DTD & 22.18 & 11.87 & 11.93 & 17.75 & 11.50 & 22\\
    MDD$_{21}$ & 22.79 & 7.88 & \underline{6.01} & 17.95 & 12.54 & 13\\
    MND$_{19}$ & 20.93 & 9.93 & 7.98 & 14.05 & 9.57 & 10\\
    SKEW$_{7}$ & 22.25 & 11.27 & 9.47 & 17.57 & 13.22 & 19\\
    VAR$_{9}$ & \underline{20.16} & 9.88 & 8.03 & \underline{11.99} & 8.36 & 2\\
    \rowcolor{lrcolor}
    ACC & 32.75 & 17.35 & 13.70 & 19.16 & 14.53 & 35\\
    LRC & 31.13 & 14.20 & 11.35 & 18.91 & 13.45 & 31\\
    LRD & 23.54 & 9.65 & 8.05 & 10.70 & 8.16 & 5\\
    UC & 31.49 & 16.52 & 13.11 & 19.29 & 14.81 & 34\\
    UCC & 21.58 & 14.01 & 12.49 & 16.27 & 13.40 & 25\\
    UCO & 36.19 & 18.54 & 14.85 & 22.60 & 15.18 & 37\\
    ZSAD & 29.63 & 23.52 & 18.70 & 21.07 & 16.86 & 38\\
    \rowcolor{distcolor}
    DTS & 50.75 & 30.49 & 22.71 & 23.65 & 21.32 & 46\\
    DSM & 25.03 & 15.06 & 12.33 & 16.00 & 13.44 & 28\\
    SAMM & 21.52 & 11.95 & 9.65 & 22.09 & 13.72 & 26\\
    \rowcolor{imagecolor}
    DB & 37.90 & 22.96 & 19.69 & 26.13 & 17.38 & 40\\
    DLB & 39.70 & 22.26 & 20.01 & 25.26 & 18.57 & 41\\
    DTE & 42.64 & 20.24 & 17.64 & 27.94 & 18.82 & 43\\
    HGM & 42.11 & 23.40 & 19.21 & 27.72 & 19.38 & 45\\
    IVAR$_{5}$ & 43.57 & 34.29 & 41.03 & 30.11 & 22.76 & 47\\
    & & & & & &  \\
    \midrule
    \rowcolor{white}
    Opt. & 12.00 & 4.72 & 3.40 & 5.31 & 4.07 & -\\
    D1(\%) & 43.58 & 27.19 & 22.28 & 28.70 & 21.27 & -\\
    \midrule
    \end{tabular}    
    \end{tabularx}
    }
    \caption{\textbf{Results with Census-CBCA algorithm}, hand-crafted measures.}
    \label{tab:census_cbca_hand}
\end{table}

\begin{figure}[t]
    \centering
    \begin{tabular}{c}
         \includegraphics[width=0.38\textwidth]{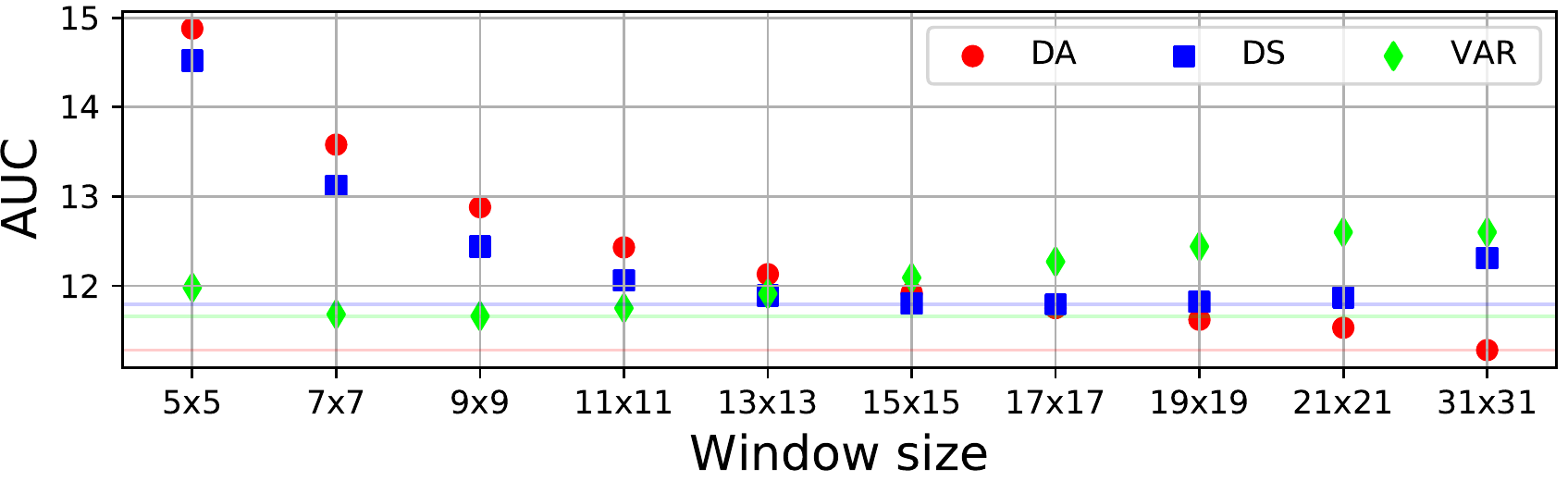} \\
         \includegraphics[width=0.38\textwidth]{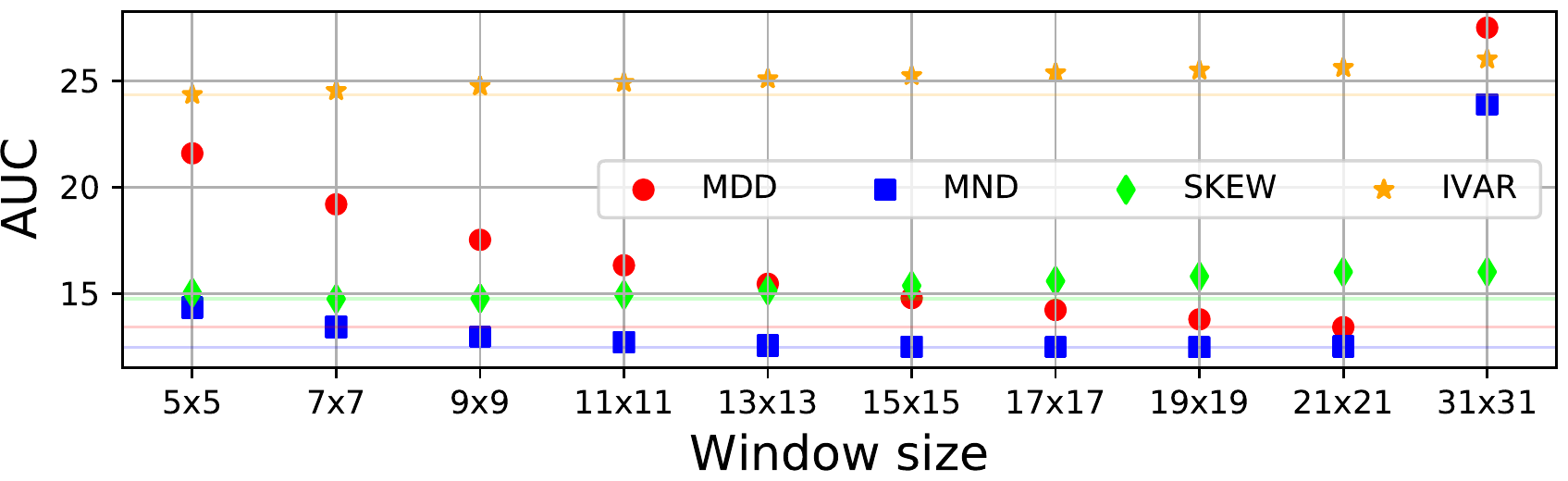} \\
         \includegraphics[width=0.38\textwidth]{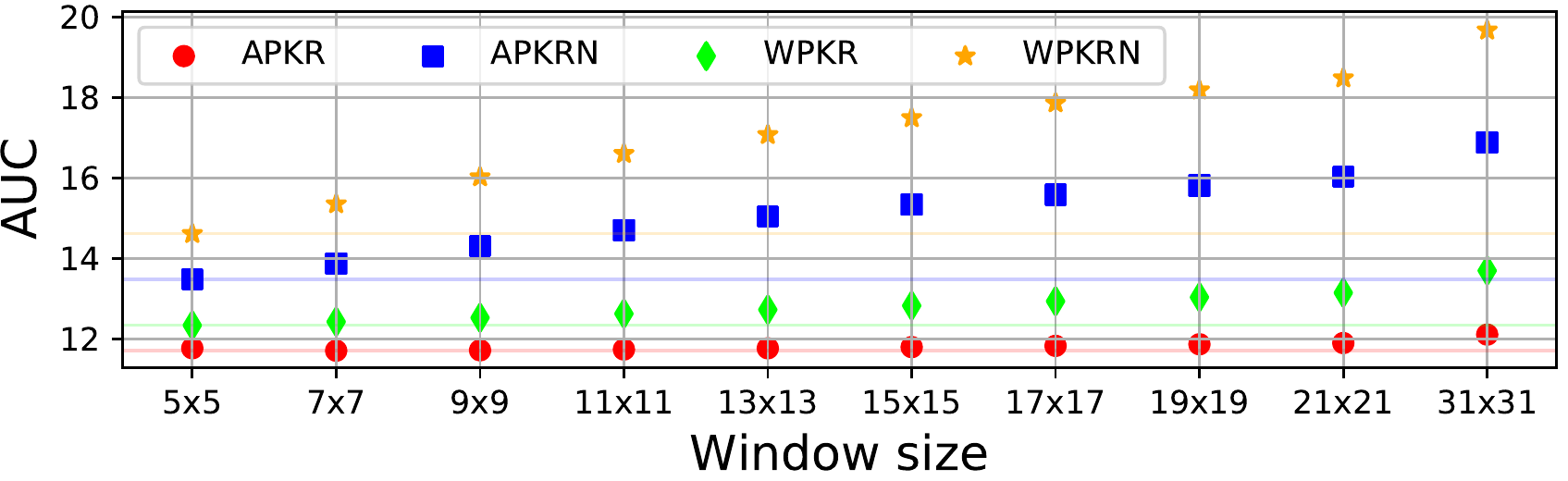} \\
    \end{tabular}
    \caption{\textbf{Impact of N(p) size}, Census-CBCA algorithm.}
    \label{fig:census_cbca_radius}
\end{figure}

\begin{table}[t]
    \centering
    \scalebox{0.68}{
    \renewcommand{\tabcolsep}{2pt}
    \begin{tabularx}{\textwidth}{cc}
    
    \begin{tabular}{.l;c;c|c;c|c;c;c.}
    \multicolumn{8}{c}{Train set: Driving} \\
    \toprule
    & Driv. & 2012 & 2015 & Midd. & ETH & R. & CR.\\
    \midrule
    \rowcolor{forestvol}
    ENS$_{23}$ & 17.08 & 8.49 & 7.44 & 13.29 & 9.36 & 19 & 16\\
    GCP & 16.10 & 7.16 & 5.93 & 11.84 & 8.33 & 16 & 5\\
    LEV$_{22}$ & 15.32 & 6.56 & 5.43 & 12.01 & 9.39 & 11 & 7\\
    LEV$_{50}$ & 14.59 & 6.25 & 5.26 & 12.37 & 7.80 & 2 & 3\\
    FA & 15.07 & 7.45 & 6.05 & 13.99 & 11.40 & 8 & 18\\
    \rowcolor{forestdisp}
    ENS$_{7}$ & 19.11 & 8.86 & 7.85 & 15.72 & 11.72 & 21 & 21\\
    O1 & 15.41 & 7.41 & 6.03 & 14.68 & 9.49 & 12 & 14\\
    O2 & 14.77 & 6.73 & 5.60 & 13.17 & 7.83 & 6 & 6\\    
    \rowcolor{deepdisp}
    CCNN & 15.30 & 6.64 & 5.46 & 16.37 & 8.98 & 10 & 13\\
    PBCP$_{r}$ & 16.07 & 6.74 & 5.44 & 14.61 & 8.56 & 15 & 11\\
    PBCP$_{d}$ & 15.90 & 6.22 & 5.22 & 14.24 & 11.94 & 13 & 15\\
    EFN & 16.73 & 7.65 & 6.80 & 18.02 & 11.54 & 18 & 20\\
    LFN & 15.15 & 7.77 & 6.32 & 15.65 & 10.30 & 9 & 19\\
    MMC & 14.65 & 7.22 & 5.78 & 14.79 & 9.37 & 3 & 12\\
    ConfNet & 15.97 & 6.56 & 5.60 & 13.30 & 8.00 & 14 & 9\\
    LGC & 14.74 & \underline{6.06} & \underline{4.94} & 14.01 & 9.18 & 4 & 10\\
    \rowcolor{deepvol}
    RCN & 23.46 & 16.97 & 14.10 & 21.73 & 15.99 & 23 & 22\\
    MPN & 16.22 & 6.53 & 5.22 & \underline{11.19} & \underline{7.08} & 17 & 1\\
    UCN & 14.97 & 6.33 & 5.05 & 12.48 & 7.28 & 7 & 2\\
    LAF & \underline{13.76} & 6.87 & 5.35 & 12.06 & 9.12 & 1 & 8\\
    ACN & 14.76 & 6.97 & 5.49 & 12.00 & 7.57 & 5 & 4\\
    CRNN & 22.27 & 16.87 & 13.91 & 21.99 & 16.86 & 22 & 23\\
    CVA & 17.38 & 9.47 & 7.52 & 12.82 & 8.87 & 20 & 17\\  
    \midrule
    \rowcolor{white}
    Opt. & 12.00 & 4.72 & 3.40 & 5.31 & 4.07 & - & -\\
    D1(\%) & 43.58 & 27.19 & 22.28 & 28.70 & 21.27 & - & -\\
    \midrule
    \end{tabular}    
    &
    \begin{tabular}{.l;c|c;c|c;c.}
    \multicolumn{6}{c}{Train set: KITTI 2012} \\
    \toprule
    & 2012 & 2015 & Midd. & ETH & R.\\
    \midrule
    \rowcolor{forestvol}
    ENS$_{23}$ & 6.62 & 5.60 & 11.15 & 8.20 & 15\\
    GCP & 6.37 & 5.29 & 11.18 & 8.54 & 14\\
    LEV$_{22}$ & 5.75 & 4.56 & 11.47 & 8.33 & 6\\
    LEV$_{50}$ & 5.67 & 4.49 & 11.88 & 8.70 & 11\\
    FA & 6.01 & 4.71 & 13.34 & 12.68 & 19\\    
    \rowcolor{forestdisp}
    ENS$_{7}$ & 7.53 & 6.28 & 14.59 & 10.92 & 20\\
    O1 & 6.15 & 4.77 & 11.45 & 9.73 & 16\\
    O2 & 5.81 & 4.53 & 11.06 & 9.28 & 10\\
    \rowcolor{deepdisp}
    CCNN & 5.76 & 4.40 & 11.24 & 9.09 & 9\\
    PBCP$_{r}$ & 6.01 & 4.89 & 10.20 & 9.19 & 7\\
    PBCP$_{d}$ & 5.54 & 4.44 & 15.01 & 14.43 & 21\\
    EFN & 6.16 & 4.74 & 13.64 & 9.84 & 17\\
    LFN & 5.80 & 4.43 & 11.81 & 9.02 & 13\\
    MMC & 5.71 & 4.36 & 11.26 & 8.98 & 8\\
    ConfNet & 6.11 & 4.85 & 11.85 & 8.24 & 12\\
    LGC & 5.59 & 4.25 & 9.93 & 7.58 & 4\\
    \rowcolor{deepvol}
    RCN & 14.79 & 13.29 & 17.59 & 13.21 & 23\\
	MPN & 5.58 & 4.31 & 9.00 & \underline{6.23} & 1\\
	UCN & 5.52 & 4.28 & 9.17 & 6.55 & 3\\ 
	LAF & \underline{5.33} & \underline{4.20} & 10.30 & 9.50 & 5\\
	ACN & 5.69 & 4.35 & \underline{8.86} & 6.49 & 2\\
	CRNN & 11.81 & 10.48 & 15.52 & 11.62 & 22\\
	CVA & 8.12 & 6.39 & 11.04 & 9.63 & 18\\
    \midrule
    \rowcolor{white}
    Opt. & 4.72 & 3.40 & 5.31 & 4.07 & -\\
    D1(\%) & 27.19 & 22.28 & 28.70 & 21.27 & -\\
    \midrule
    \end{tabular}    
    \end{tabularx}
    }
    \caption{\textbf{Results with Census-CBCA algorithm}, learned measures.} 
    \label{tab:census_cbca_learning}
\end{table}

\begin{figure*}[t]
    \scriptsize
    
    \centering
    \renewcommand{\tabcolsep}{1pt}
        \rotatebox{90}{KITTI} \rotatebox{90}{2015}
        \subfigure{
        \includegraphics[width=0.11\textwidth, frame]{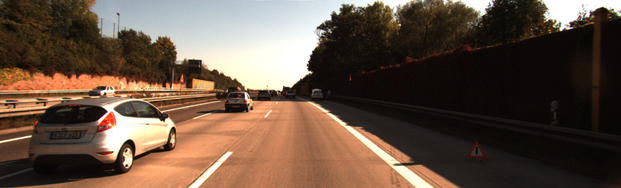}}
        \subfigure{
        \includegraphics[width=0.11\textwidth, frame]{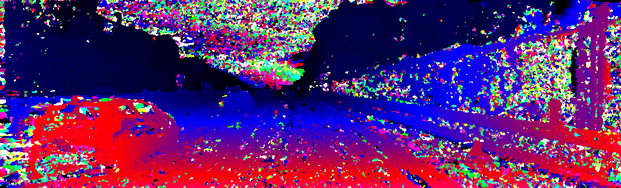}}
        \subfigure{
         \includegraphics[width=0.11\textwidth, frame]{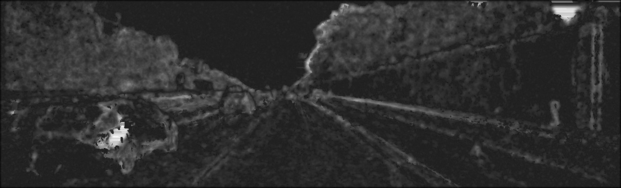}}
         \subfigure{
        \includegraphics[width=0.11\textwidth, frame]{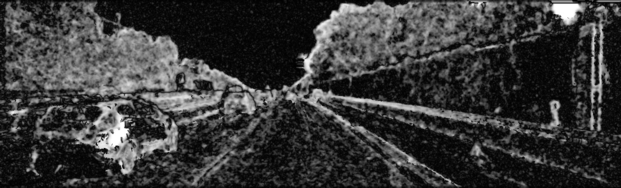}}
        \subfigure{
        \includegraphics[width=0.11\textwidth, frame]{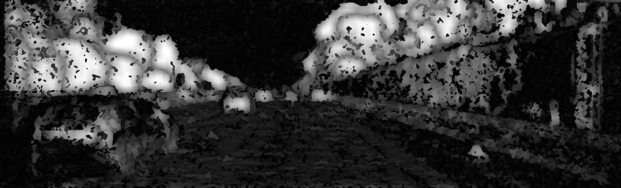}}
        \subfigure{
        \includegraphics[width=0.11\textwidth, frame]{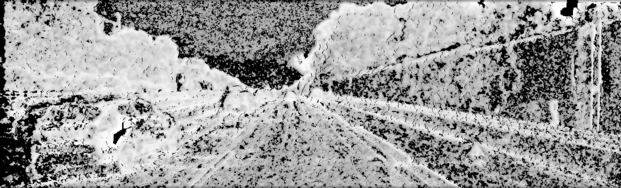}}
        \subfigure{
        \includegraphics[width=0.11\textwidth, frame]{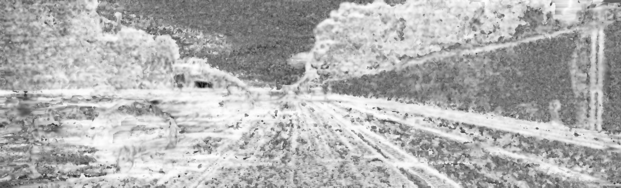}}
        \subfigure{
        \includegraphics[width=0.11\textwidth, frame]{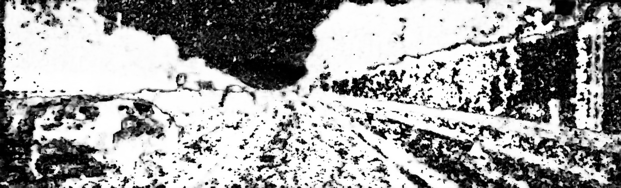}  }
        \\
        \rotatebox{90}{Middlebury}
        \subfigure{
        \includegraphics[width=0.11\textwidth, frame]{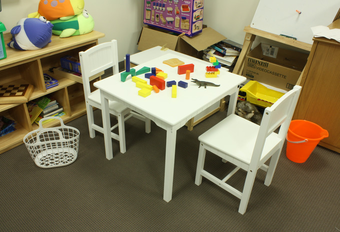}}
        \subfigure{
        \includegraphics[width=0.11\textwidth, frame]{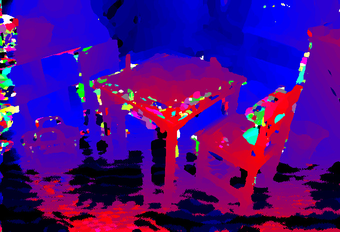}}
        \subfigure{
        \includegraphics[width=0.11\textwidth, frame]{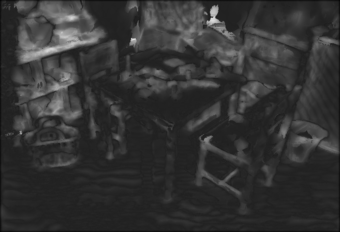}}
        \subfigure{
        \includegraphics[width=0.11\textwidth, frame]{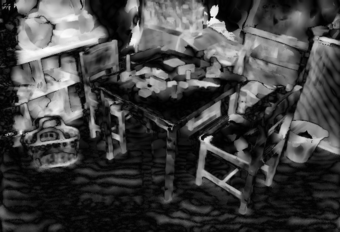}} 
        \subfigure{
        \includegraphics[width=0.11\textwidth, frame]{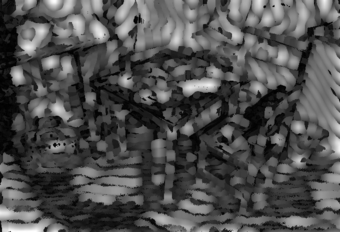}}
        \subfigure{
        \includegraphics[width=0.11\textwidth, frame]{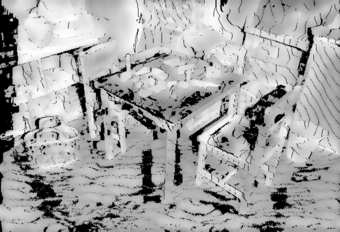}} 
        \subfigure{
        \includegraphics[width=0.11\textwidth, frame]{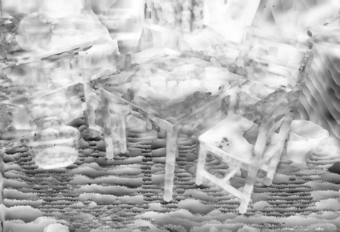}}
        \subfigure{
        \includegraphics[width=0.11\textwidth, frame]{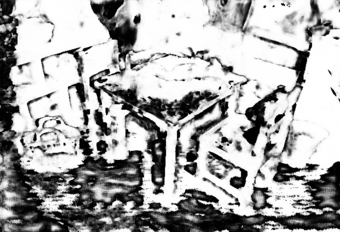}  }
        \\
    \caption{\textbf{Qualitative results concerning Census-CBCA algorithm.} Results on KITTI 2015 and Middlebury showing a variety of confidence measures. From top left to bottom right: reference image, disparity map and confidence maps by APKR$_7$, WMN, DA$_{31}$, UCC, SAMM and LAF.} 
    \label{fig:qualitative_census_cbca}
\end{figure*}

\textbf{Impact of the windows size.} Figure \ref{fig:census_cbca_radius} plots the AUC achieved by varying the radius of $N(p)$ for measures computed over a local neighborhood. Interestingly, we can notice how different measures behave differently, highlighting that not always the larger local context leads to the better performance. Indeed, this is true only for DA, achieving its best performance with $31\times31$ windows. In general, measures computed from the \colorbox{dispcolor}{disparity domain} such as DS, MDD and MND get the best results with medium/large windows size, respectively 17, 19 and 21 size. Finally, measures APKR, APKRN, WPKR, WPKRN based on \colorbox{localcolor}{local properties} tend to perform better with smaller kernels of size 5 or 7. Similarly, IVAR performs better with a small window, \ie $5\times5$.

\textbf{Learned measures, synthetic data training.} Table \ref{tab:census_cbca_learning}, on the left, collects results for learned measures when trained on synthetic images from the Driving train split. 

Concerning results on the synthetic test split, LAF performs the best, followed by LEV$_{50}$. This highlights that, either using \colorbox{forestvol}{cost-volume forests} or \colorbox{deepvol}{cost-volume CNNs}, the information in the cost volume can be useful if properly leveraged, in particular with adequate receptive fields. Nevertheless, MMC and LGCNet show that \colorbox{deepdisp}{disparity CNNs} can be very close to the top-2 methods using the cost volume.
Finally, comparing tables \ref{tab:census_cbca_hand} and \ref{tab:census_cbca_learning} we can notice that, excluding RF and CRNN, learned measures always outperform hand-crafted ones on the synthetic test split.

Concerning generalization to real data, comparing the two tables again, we point out that most learned measures outperform the top-performing hand-crafted measure (DA$_{31}$) on KITTI 2012 and 2015, except ENS$_{23}$. This evidence suggests that learning confidence estimation suffers from the domain shift from synthetic to real much less than other tasks such as, for instance, learning stereo matching \cite{Tonioni_2017_ICCV,Tonioni_2019_CVPR,Tonioni_2019_learn2adapt,tonioni2019unsupervised}.
A possible reason we ascribe it to is the much more structured appearance observed in the disparity domain, the primary cue processed for this task, where smooth surfaces are very likely to be met, and sharp edges occur near depth discontinuities either in real or simulated environments, conversely to raw image appearance that differs a lot from synthetic to real scenes, for instance, because of lightning conditions and noise. Thus, detecting outliers in such a well-defined domain is a simpler task than facing stereo matching from raw images.

Despite this fact, many learned measures (\eg FA) poorly perform on Middlebury and ETH, being often outperformed by the best hand-crafted ones, such as DA and VAR. This behavior is probably due to the different geometry of indoor vs outdoor scenes, confirming our previous findings in \cite{poggi2017quantitative}.
Other methods performing very well on synthetic images and affected by domain shift issues are LAF and LGC.

One might argue that methods trained by assuming $\tau=3$ are penalized when the dataset threshold is lower, i.e. $\tau=1$. However, by setting $\tau=3$ for Middlebury and ETH3D, the relative order is, in most cases, unaltered.

In contrast, some methods keep their good ranking unaltered (\eg LEV$_{50}$, O2) or even significantly improve it, such as GCP and ConfNet. Moreover, MPN surprisingly jumps to rank 1, exposing excellent generalization properties.

\textbf{Learned measures, real data training.} Table \ref{tab:census_cbca_learning}, on the right, collects results for measures trained on KITTI 2012 20 training images. By comparing the numbers on the left and right side of the table, on KITTI 2012 and 2015, we can notice how, not surprisingly, training on real images allows for better accuracy on these datasets. However, the tiny improvements (we recall that reported AUC are multiplied by $10^2$) confirm a marginal impact of domain-shift on the confidence estimation task. A similar behavior, except for PBCP$_{d}$, can be noticed on Middlebury while on ETH3D, many learned measures trained on KITTI 2012 (\eg LEV$_{50}$, O2, PBCPs, LAF and CVA) achieve worse performance. We ascribe this behavior to the same reason outlined previously. Overall, the top-performing learned measures training on KITTI 2012 turn out  \colorbox{deepvol}{cost-volume CNNs} and LGC belonging to \colorbox{deepdisp}{disparity CNNs} category.
Finally, we observe that top-performing hand-crafted measures are competitive and, sometimes, better than learned ones training on KITTI 2012.

\textbf{Qualitative results.} We report in Fig. \ref{fig:qualitative_census_cbca} disparity maps from KITTI 2015 and Middlebury. The figure also shows, on different columns, confidence maps for five hand-crafted measures to highlight their behaviors and the outcome of a learned confidence measure trained on KITTI 2012 splits to highlight the effects of the domain shift. We can notice how most traditional measures tend to assign low confidence, probably because of the noisy cost volumes and disparities produced by Census-CBCA. More advanced measures like SAMM or learned as LAF, when assigning low confidence, can better focus on the real outliers.

\textbf{Summary.} When dealing with noisy algorithms such as Census-CBCA, hand-crafted measures computed on a local window result very effective, even without processing the cost volume. This is also confirmed for learned measures, where a large receptive field exploited by networks and forests always improves the results. The disparity map alone allows for the best results in the case of hand-crafted measures and for competitive effectiveness in the case of learned methods.

\begin{table}[t]
    \centering
    \scalebox{0.68}{
    \renewcommand{\tabcolsep}{2pt}
    \begin{tabularx}{\textwidth}{cc}
    
    \begin{tabular}{.l;c|c|c|c|c;c.}
    \toprule
    & Driv. & 2012 & 2015 & Midd. & ETH & R.\\
    \midrule
    \rowcolor{localcolor}
    APKR$_{19}$ & 16.18 & 4.91 & 4.88 & \underline{8.63} & 13.94 & 1\\
    APKRN$_{5}$ & 24.07 & 8.47 & 8.42 & 10.97 & 17.11 & 27\\
    CUR & 30.86 & 13.17 & 12.10 & 14.14 & 23.14 & 34\\
    DAM & 29.83 & 12.95 & 12.66 & 19.69 & 24.98 & 37\\
    LC & 29.93 & 12.80 & 11.94 & 13.60 & 23.01 & 33\\
    MM & 17.34 & 6.23 & 6.17 & 10.21 & 17.73 & 13\\
    MMN & 28.68 & 10.54 & 10.42 & 13.05 & 19.68 & 29\\
    MSM & 18.38 & 9.70 & 8.84 & 10.40 & 21.37 & 25\\
    NLM & 17.34 & 6.63 & 6.17 & 10.21 & 17.73 & 14\\
    NLMN & 28.68 & 10.54 & 10.42 & 13.05 & 19.68 & 30\\
    PKR & 16.48 & 6.55 & 6.25 & 9.53 & 18.09 & 12\\
    PKRN & 22.75 & 8.24 & 8.16 & 11.06 & 17.98 & 23\\
    SGE & 17.80 & 9.36 & 8.48 & 10.18 & 20.93 & 22\\
    WPKR$_{11}$ & 16.47 & 5.44 & 5.47 & 8.71 & 15.34 & 4\\
    WPKRN$_{5}$ & 24.34 & 9.42 & 9.56 & 11.17 & 17.53 & 28\\
    \rowcolor{fullcolor}
    ALM & 15.82 & 6.26 & 6.47 & 9.37 & 17.71 & 11\\
    LMN & 27.37 & 9.59 & 8.41 & 15.68 & 21.99 & 31\\
    MLM & 15.55 & 5.56 & 5.77 & 9.23 & 16.71 & 6\\
    NEM & 27.22 & 19.36 & 17.06 & 26.06 & 33.36 & 42\\
    NOI & 31.06 & 22.20 & 18.94 & 27.40 & 32.63 & 46\\
    PER & 15.78 & 6.13 & 6.34 & 9.35 & 17.54 & 8\\
    PWCFA & 16.63 & 5.60 & 6.09 & 9.78 & 17.42 & 10\\
    WMN & 16.84 & 7.19 & 6.90 & 9.31 & 18.67 & 15\\
    WMNN & 20.23 & 7.63 & 7.52 & 10.40 & 17.67 & 18\\    
    \midrule
    \rowcolor{white}
    Opt. & 9.63 & 2.29 & 2.16 & 5.63 & 10.31 & -\\
    D1(\%) & 39.00 & 18.71 & 16.93 & 29.80 & 34.28 & -\\
    \midrule
    \end{tabular}
    &
    \begin{tabular}{.l;c|c|c|c|c;c.}
    \toprule
    & Driv. & 2012 & 2015 & Midd. & ETH & R.\\
    \midrule
    \rowcolor{dispcolor}
    DA$_{31}$ & 17.45 & 5.38 & 5.16 & 8.81 & \underline{12.70} & 2\\
    DMV & 18.44 & 6.16 & 6.01 & 11.29 & 17.68 & 16\\
    DS$_{9}$ & 16.15 & 5.07 & 4.71 & 8.95 & 15.24 & 3\\
    DTD & 17.69 & 6.78 & 6.82 & 13.43 & 20.99 & 20\\
    MDD$_{21}$ & 16.85 & \underline{4.58} & \underline{4.42} & 11.62 & 15.38 & 7\\
    MND$_{9}$ & 16.43 & 5.32 & 5.12 & 10.85 & 17.53 & 9\\
    SKEW$_{5}$ & 17.33 & 6.24 & 6.23 & 12.21 & 18.81 & 17\\
    VAR$_{5}$ & \underline{15.64} & 5.33 & 5.02 & 9.57 & 16.52 & 5\\
    \rowcolor{lrcolor}
    ACC & 28.95 & 12.50 & 11.29 & 19.32 & 26.47 & 36\\
    LRC & 26.66 & 10.27 & 9.24 & 18.76 & 23.90 & 32\\
    LRD & 20.86 & 8.02 & 7.66 & 10.68 & 18.45 & 19\\
    UC & 28.33 & 12.26 & 11.06 & 19.21 & 26.36 & 35\\
    UCC & 18.84 & 8.74 & 8.08 & 10.73 & 19.70 & 21\\
    UCO & 32.70 & 13.59 & 11.59 & 22.73 & 27.16 & 38\\
    ZSAD & 25.55 & 19.49 & 16.77 & 22.51 & 28.79 & 39\\
    \rowcolor{distcolor}
    DTS & 34.11 & 22.87 & 20.95 & 38.30 & 37.16 & 47\\
    DSM & 18.39 & 9.73 & 8.85 & 10.54 & 21.49 & 26\\
    SAMM & 15.91 & 6.07 & 5.47 & 19.33 & 21.70 & 24\\
    \rowcolor{imagecolor}
    DB & 34.69 & 16.71 & 15.52 & 27.59 & 29.84 & 43\\
    DLB & 34.39 & 14.78 & 14.52 & 26.39 & 31.21 & 41\\
    DTE & 40.06 & 14.49 & 13.75 & 28.73 & 27.64 & 44\\
    HGM & 38.23 & 16.31 & 14.78 & 28.17 & 30.05 & 45\\
    IVAR$_{5}$ & 40.21 & 12.71 & 12.75 & 27.11 & 25.54 & 40\\
    & & & & & &  \\
    \midrule
    \rowcolor{white}
    Opt. & 9.63 & 2.24 & 2.16 & 5.63 & 10.31 & -\\
    D1(\%) & 39.00 & 18.88 & 16.93 & 29.80 & 34.28 & -\\
    \midrule
    \end{tabular}    
    \end{tabularx}
    }
    \caption{\textbf{Results with MCCNN-CBCA algorithm}, hand-crafted measures.}
    \label{tab:mccnn_cbca_hand}
\end{table}

\begin{figure}[t]
    \centering
    \begin{tabular}{c}
         \includegraphics[width=0.38\textwidth]{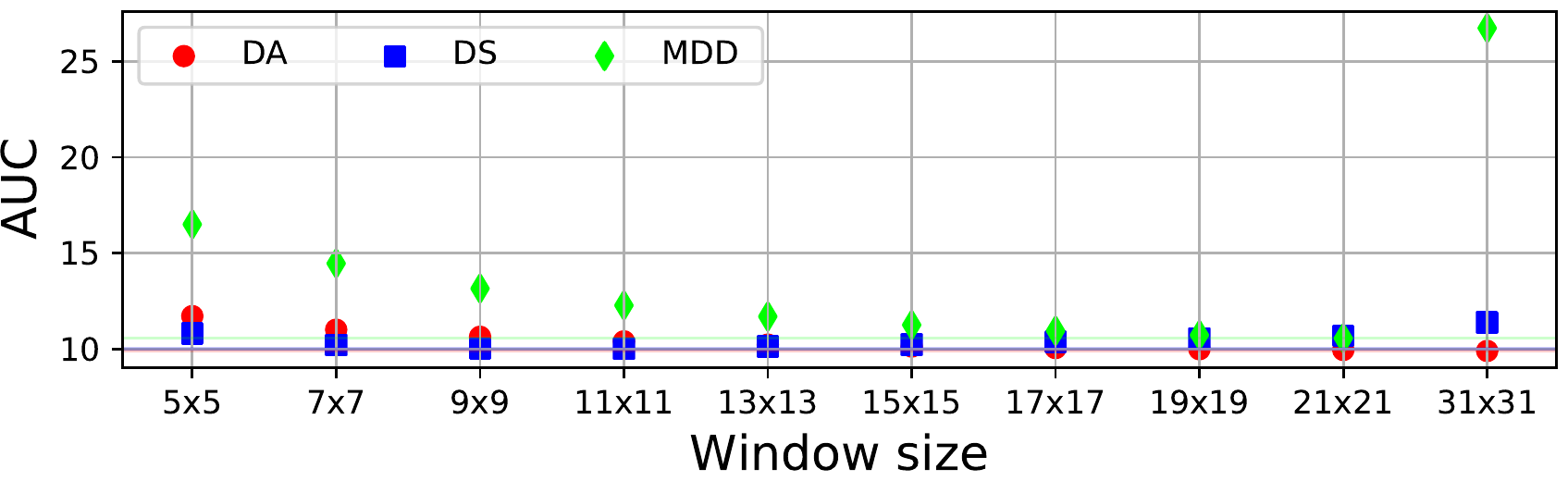} \\
         \includegraphics[width=0.38\textwidth]{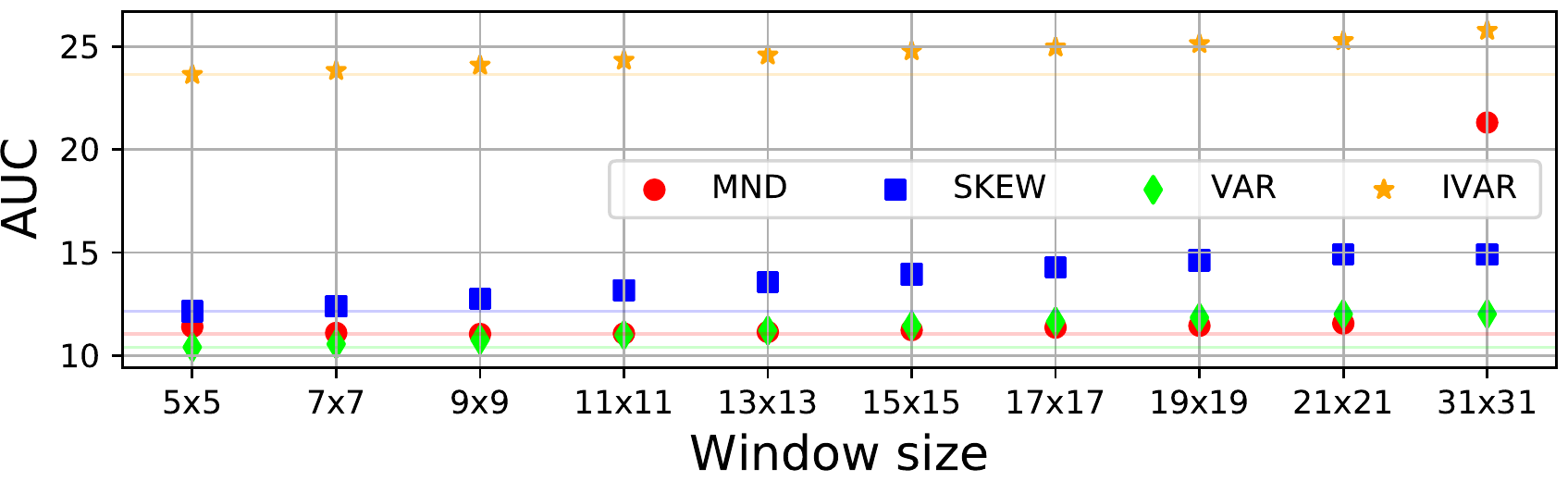} \\
         \includegraphics[width=0.38\textwidth]{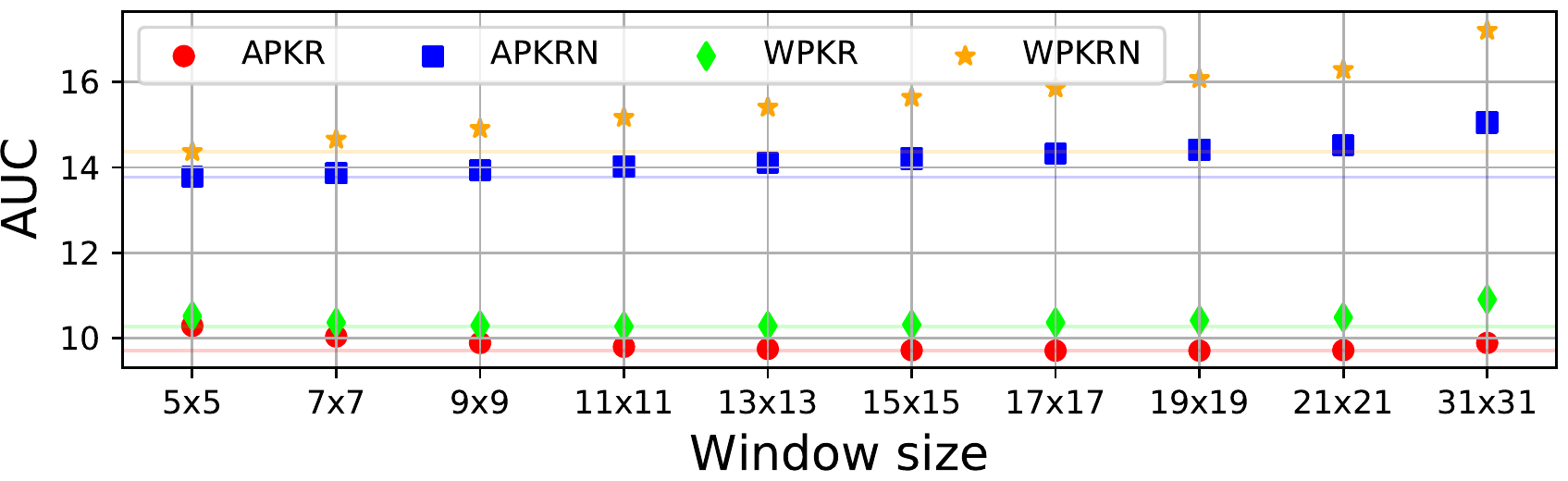} \\
    \end{tabular}
    \caption{\textbf{Impact of N(p) size}, MCCNN-CBCA algorithm.}
    \label{fig:mccnn_cbca_radius}
\end{figure}

\begin{table}[t]
    \centering
    \scalebox{0.68}{
    \renewcommand{\tabcolsep}{2pt}
    \begin{tabularx}{\textwidth}{cc}
    
    \begin{tabular}{.l;c;c|c;c|c;c;c.}
    \multicolumn{8}{c}{Train set: Driving} \\
    \toprule
    & Driv. & 2012 & 2015 & Midd. & ETH & R. & CR. \\
    \midrule
    \rowcolor{forestvol}
    ENS$_{23}$ & 13.67 & 4.30 & 4.75 & 10.36 & 16.21 & 20 & 20\\
    GCP & 12.65 & 3.64 & 3.78 & 10.27 & 17.34 & 18 & 18\\
    LEV$_{22}$ & 12.16 & 3.11 & 3.43 & 9.26 & 14.72 & 13 & 7\\
    LEV$_{50}$ & 11.70 & \underline{3.04} & 3.37 & 10.13 & 14.10 & 5 & 9\\
    FA & 11.38 & 3.66 & 3.66 & 9.32 & 13.79 & 2 & 5\\
    \rowcolor{forestdisp}
    ENS$_{7}$ & 14.88 & 5.05 & 5.46 & 11.03 & 16.67 & 21 & 21\\
    O1 & 12.19 & 3.88 & 4.16 & 9.73 & 13.39 & 14 & 12\\
    O2 & 11.82 & 3.77 & 4.02 & 9.79 & 13.00 & 7 & 8\\
    \rowcolor{deepdisp}
    CCNN & 12.15 & 3.45 & 3.78 & 10.10 & 13.13 & 12 & 6\\
    PBCP$_{r}$ & 12.51 & 3.21 & \underline{3.35} & 9.28 & 13.68 & 17 & 3\\
    PBCP$_{d}$ & 11.95 & 3.96 & 4.49 & 9.73 & 14.57 & 10 & 15\\
    EFN & 12.48 & 4.31 & 4.84 & 10.58 & 14.65 & 15 & 17\\
    LFN & 11.79 & 4.36 & 4.53 & 9.76 & 14.18 & 6 & 16\\
    MMC & 11.40 & 4.14 & 4.21 & 9.41 & 13.74 & 3 & 14\\
    ConfNet & 12.48 & 3.56 & 3.77 & 9.17 & 13.59 & 16 & 4\\
    LGC & 11.66 & 3.12 & 3.52 & 8.84 & 13.31 & 4 & 2\\
    \rowcolor{deepvol}
    RCN & 18.94 & 6.89 & 6.77 & 24.35 & 28.77 & 23 & 23\\
    MPN & 11.99 & 3.80 & 4.13 & 9.96 & 13.51 & 11 & 13\\
    UCN & 11.86 & 3.34 & 3.69 & 10.10 & 13.99 & 8 & 12\\
    LAF  & \underline{11.00} & 3.79 & 4.00 & 9.24 & 14.08 & 1 & 11\\
    ACN  & 11.88 & 3.45 & 3.60 & \underline{8.59} & \underline{12.81} & 9 & 1\\
    CRNN & 16.01 & 5.89 & 5.67 & 21.69 & 26.06 & 22 & 22\\
    CVA & 12.67 & 4.24 & 4.52 & 11.48 & 15.09 & 19 & 19\\
    \midrule
    \rowcolor{white}
    Opt. & 9.63 & 2.35 & 2.16 & 5.63 & 10.31 & - & -\\
    D1(\%) & 39.00 & 18.88 & 16.93 & 29.80 & 34.28 & - & -\\
    \midrule
    \end{tabular}    
    &
    \begin{tabular}{.l;c|c;c|c;c.}
    \multicolumn{6}{c}{Train set: KITTI 2012} \\
    \toprule
    & 2012 & 2015 & Midd. & ETH & R.\\
    \midrule
    \rowcolor{forestvol}
    ENS$_{23}$ & 3.53 & 3.76 & 9.58 & 14.48 & 17\\
    GCP & 3.44 & 3.44 & 9.86 & 15.69 & 19\\
    LEV$_{22}$ & 3.05 & 3.05 & 8.48 & 13.74 & 10\\
    LEV$_{50}$ & 2.89 & 2.97 & 8.45 & 13.45 & 5\\
    FA & 3.00 & 3.02 & 8.33 & 14.00 & 11\\
    \rowcolor{forestdisp}
    ENS$_{7}$ & 3.94 & 4.33 & 10.95 & 16.37 & 21\\
    O1 & 2.96 & 2.93 & 8.25 & 15.18 & 14\\
    O2 & 2.79 & 2.87 & 8.09 & 14.92 & 13\\
    \rowcolor{deepdisp}
    CCNN & 2.84 & 2.91 & 8.23 & 14.23 & 9\\
    PBCP$_{r}$ & 3.27 & 3.48 & 7.86 & 14.98 & 15\\
    PBCP$_{d}$ & 2.95 & 3.06 & 11.26 & 17.10 & 20\\
    EFN & 3.22 & 3.15 & 9.44 & 13.98 & 16\\
    LFN & 3.04 & 3.00 & 8.45 & 14.15 & 12\\
    MMC & 2.95 & 2.94 & 8.15 & 13.78 & 6\\
    ConfNet & 3.22 & 3.32 & 8.31 & 13.34 & 8\\
    LGC & 2.96 & \underline{2.77} & 8.09 & 14.37 & 7\\
    \rowcolor{deepvol}
    RCN & 5.53 & 5.35 & 17.75 & 24.99 & 23\\
    MPN & 3.03 & 3.09 & 8.26 & 13.02 & 4\\
    UCN & 2.85 & 2.90 & 8.07 & \underline{13.01} & 1\\
    LAF & \underline{2.81} & 2.90 & \underline{8.03} & 13.17 & 2\\
    ACN & 2.94 & 3.03 & 8.24 & 13.09 & 3\\
    CRNN & 5.08 & 4.87 & 17.75 & 25.02 & 22\\
    CVA & 3.31 & 3.38 & 9.34 & 15.78 & 18\\
    \midrule
    \rowcolor{white}
    Opt. & 2.35 & 2.16 & 5.63 & 10.31 & -\\
    D1(\%) & 18.88 & 16.93 & 29.80 & 34.28 & -\\
    \midrule
    \end{tabular}    
    \end{tabularx}
    }
    \caption{\textbf{Results with MCCNN-CBCA algorithm}, learned measures.}
    \label{tab:mccnn_cbca_learning}
\end{table}

\subsubsection{MCCNN-CBCA}

\textbf{Hand-crafted measures.} Table \ref{tab:mccnn_cbca_hand} reports the performance achieved by hand-crafted measures highlighting how, with MCCNN-CBCA, the top-performing measure is APKR$_{17}$. Similar to Census-CBCA experiments measures processing the \colorbox{dispcolor}{disparity map} perform very well, with DA$_{31}$ and DS$_9$ in the top-3 with others three out of six positionings in the top-10. Confidence measures processing \colorbox{localcolor}{local properties}, in particular WPKR$_{11}$ in addition to APKR$_{17}$, or the \colorbox{fullcolor}{entire cost curve} as MLM, PER and PWCFA perform very well. In contrast, the best methods exploiting \colorbox{lrcolor}{left-right consistency} features rank 19 and 21 with LRD and UCC, respectively. Among \colorbox{distcolor}{self-matching} measures, SAMM achieves better results and ranks 24, while estimating confidence only from \colorbox{imagecolor}{image properties} confirms ineffective as always.

\textbf{Impact of the windows size.} Figure \ref{fig:mccnn_cbca_radius} plots the AUC achieved by varying the radius of $N(p)$ for measures computed over a local neighborhood. DA confirms to perform better on large $31\times31$ windows, 
while the other \colorbox{dispcolor}{disparity map} features show mixed behaviors, with VAR and SKEW preferring a small window of size 5, MND and DS of size 9 and MDD of 21. Methods based on \colorbox{localcolor}{Local properties} perform differently, with APKR achieving the top-1 position with a kernel of size 19 and WPKR ranking 4th with a window of size 11. \textit{Naive} variants, as well as IVAR, worsen with kernels larger than 5.

\begin{figure*}[t]
    \scriptsize
    
    \centering
    \renewcommand{\tabcolsep}{1pt}
        \rotatebox{90}{KITTI} \rotatebox{90}{2015}
        \subfigure{
        \includegraphics[width=0.11\textwidth, frame]{images/qualitative/census-CBCA/2015/000197_10.png}}
        \subfigure{
        \includegraphics[width=0.11\textwidth, frame]{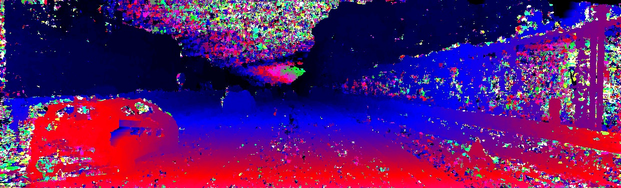}}
        \subfigure{
         \includegraphics[width=0.11\textwidth, frame]{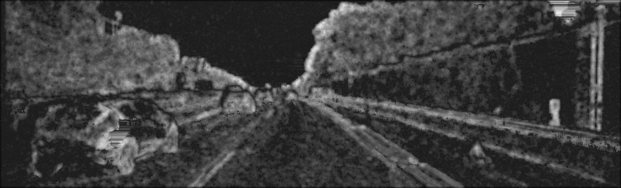}}
         \subfigure{
        \includegraphics[width=0.11\textwidth, frame]{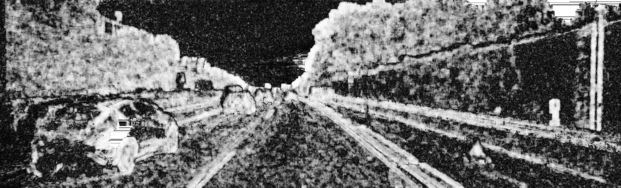}}
        \subfigure{
        \includegraphics[width=0.11\textwidth, frame]{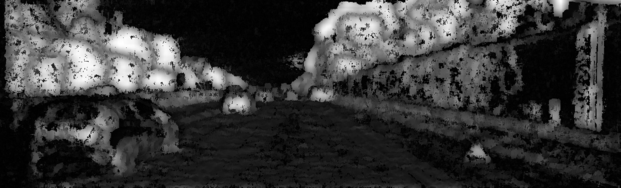}}
        \subfigure{
        \includegraphics[width=0.11\textwidth, frame]{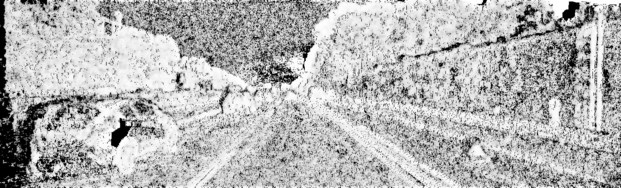}}
        \subfigure{
        \includegraphics[width=0.11\textwidth, frame]{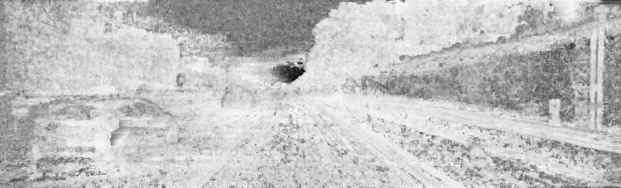}}
        \subfigure{
        \includegraphics[width=0.11\textwidth, frame]{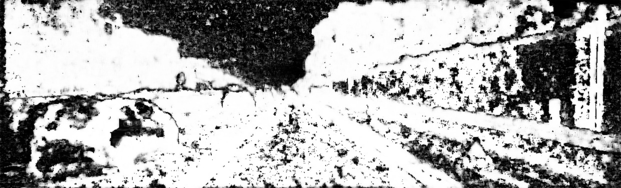}  }
        \\
        \rotatebox{90}{Middlebury}
        \subfigure{
        \includegraphics[width=0.11\textwidth, frame]{images/qualitative/census-CBCA/MiddEval3/Playtable.png}}
        \subfigure{
        \includegraphics[width=0.11\textwidth, frame]{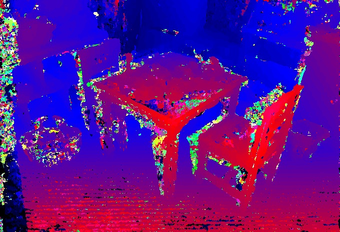}}
        \subfigure{
        \includegraphics[width=0.11\textwidth, frame]{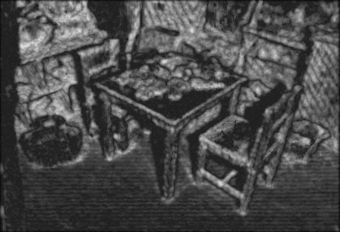}}
        \subfigure{
        \includegraphics[width=0.11\textwidth, frame]{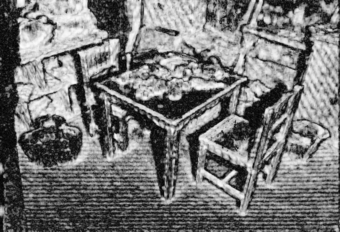}} 
        \subfigure{
        \includegraphics[width=0.11\textwidth, frame]{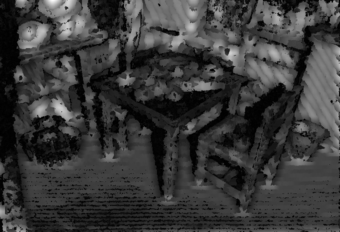}}
        \subfigure{
        \includegraphics[width=0.11\textwidth, frame]{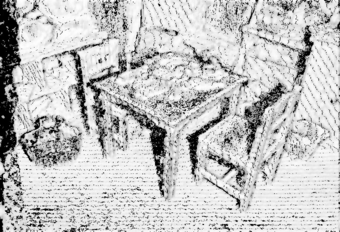}} 
        \subfigure{
        \includegraphics[width=0.11\textwidth, frame]{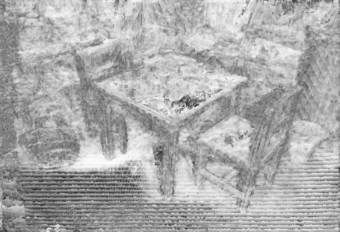}}
        \subfigure{
        \includegraphics[width=0.11\textwidth, frame]{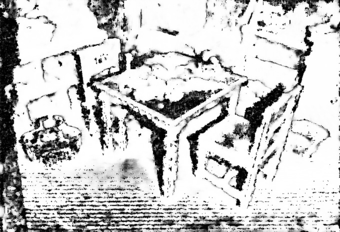}  }
        \\
    \caption{\textbf{Qualitative results concerning MCCNN-CBCA algorithm.} Results on KITTI 2015 and Middlebury showing a variety of confidence measures. From top left to bottom right: reference image, disparity map and confidence maps by APKR$_7$, WMN, DA$_{31}$, UCC, SAMM and LAF.} 
    \label{fig:qualitative_mccnn_cbca}
\end{figure*}

\textbf{Learned measures, synthetic data training.} Table \ref{tab:mccnn_cbca_learning}, on left, collects results for learned measures when trained on synthetic images from the Driving train split. 

When testing on synthetic data, LAF outperforms all the competitors as observed on Census-CBCA experiments, this time followed by FA. MMC and LGC follow, confirming that for noisy CBCA algorithms, \colorbox{deepdisp}{disparity CNNs} are competitive with both \colorbox{forestvol}{cost-volume forests} and \colorbox{deepvol}{cost-volume CNNs}. This fact is confirmed by O1 and O2, being both outperformed with minor margins, respectively by LEV and LEV$_{50}$. Again, larger receptive fields seem beneficial when the cost volume is not processed.

Concerning generalization to real data, as for Census-CBCA, we observe that most learned measures outperform the top-performing hand-crafted one APKR$_{17}$ on KITTI 2012 and 2015, confirming that the domain shift from synthetic to real is much less evident when dealing with CBCA algorithms. Nonetheless, the performance on Middlebury and ETH3D are still comparable with hand-crafted methods.
Overall, ACN surprisingly jumps to rank 1, while LAF drops to rank 11, while LGC and O2 and show a more stable behavior and keep their position almost unaltered, with the former achieving rank 2.

Looking at patch-based methods, the comparison between PBCP$_{r}$, CCNN and PBCP$_{d}$ confirms the previous findings, with the latter performing better on images similar to the training set but dropping when evaluating across domains. PBCP$_{r}$ variant is, on the contrary, much more robust to domain shifts and reaches rank 3 in this case.
Since RGB data can be significantly affected by the domain shift, most measures processing the reference image witness a massive drop in accuracy when crossing domains. Nonetheless, a notable exception is ConfNet, a component of LGC, which seems particularly good at generalization, as observed in the previous experiments.

\textbf{Learned measures, real data training.} Table \ref{tab:mccnn_cbca_learning}, on the right, gathers results for learned measures when trained on KITTI 2012 20 training images. 
\colorbox{deepvol}{Cost-volume CNNs} cover the top-4 positions in the leaderboard, followed by LEV$_{50}$. Then MMC, LGC, ConfNet and CCNN that are \colorbox{deepdisp}{disparity CNNs}. Finally, \colorbox{forestvol}{cost-volume forests} such as FA and LEV outperform, on average, O1 and O2, while being less effective on specific datasets such as in the case of KITTI 2015 and Middlebury.

Concerning patch-based methods, this time CCNN outperforms both PBCP variants, with PBCP$_{r}$ better at generalizing compared to PBCP$_{d}$ as observed so far. On the other hand, MMC outperforms CCNN thanks to the much larger receptive field. Moreover, MMC also surpasses networks, such as ConfNet and LGC, with comparable receptive fields.

\textbf{Qualitative results.} To conclude this section, Fig. \ref{fig:qualitative_mccnn_cbca} shows some disparity maps from KITTI 2015 and Middlebury together with confidence maps. In particular, it highlights how hand-crafted measures belonging to different categories behave very differently and, in the rightmost column, the effects introduced by the domain shift on a learned measure.
Compared to what seen with Census-CBCA, the traditional measures assign high confidence more often, probably because of the more robust cost volumes and disparities produced by replacing the census transform with MCCNN-fst. However, from confidence maps by LAF it is still easier to discriminate good matches from outliers.

\textbf{Summary.} By replacing the matching costs computed by the census transform with those computed by MCCNN-fst, the behavior of most confidence measures remains unaltered, both for hand-crafted measures, among which several measures processing the disparity map only still result the most effective among the hand-crafted ones, or are competitive in the case of learned measures.

\subsection{SGM Algorithms}

We now assess confidence estimation performance with stereo methods leveraging SGM for cost volume optimization, one of the most versatile and popular solutions for its good trade-off between accuracy and computational complexity. Differently from the local algorithms tested so far, SGM disparity maps are much more accurate due to the \textit{global} nature of the optimization carried out, making the outlier detection task significantly more challenging.

\subsubsection{CENSUS-SGM}

In this section, we discuss the outcome of our experiments carried out with the Census-SGM algorithm.

\textbf{Hand-crafted measures.} Table \ref{tab:census_sgm_hand} shows the performance achieved by hand-crafted measures. The top-performing measure is VAR$_{19}$ while, interestingly, the remaining ones processing the \colorbox{dispcolor}{disparity map} perform poorly this time because of the much smoother outputs by Census-SGM. Measures processing \colorbox{localcolor}{local properties} or the \colorbox{fullcolor}{entire cost curve} perform better in general, with MM, NLM and some of those based on peak ratio (\eg APKR$_{5}$ and PKR) achieving excellent results.
Confidence estimated from \colorbox{lrcolor}{left-right consistency}, UCC and LRD in particular, turns out better than measures computed from the left disparity map only. Among \colorbox{distcolor}{self-matching} measures, DSM achieves the best accuracy despite far from top-performing ones, while estimating confidence only from \colorbox{imagecolor}{image properties} confirms ineffective once again.
\colorbox{sgmcolor}{SGM-specific measures} tailored for SGM such as PS and SCS show an average performance, placing at the middle of the leaderboard.

\begin{figure}[t]
    \centering
    \begin{tabular}{c}
         \includegraphics[width=0.38\textwidth]{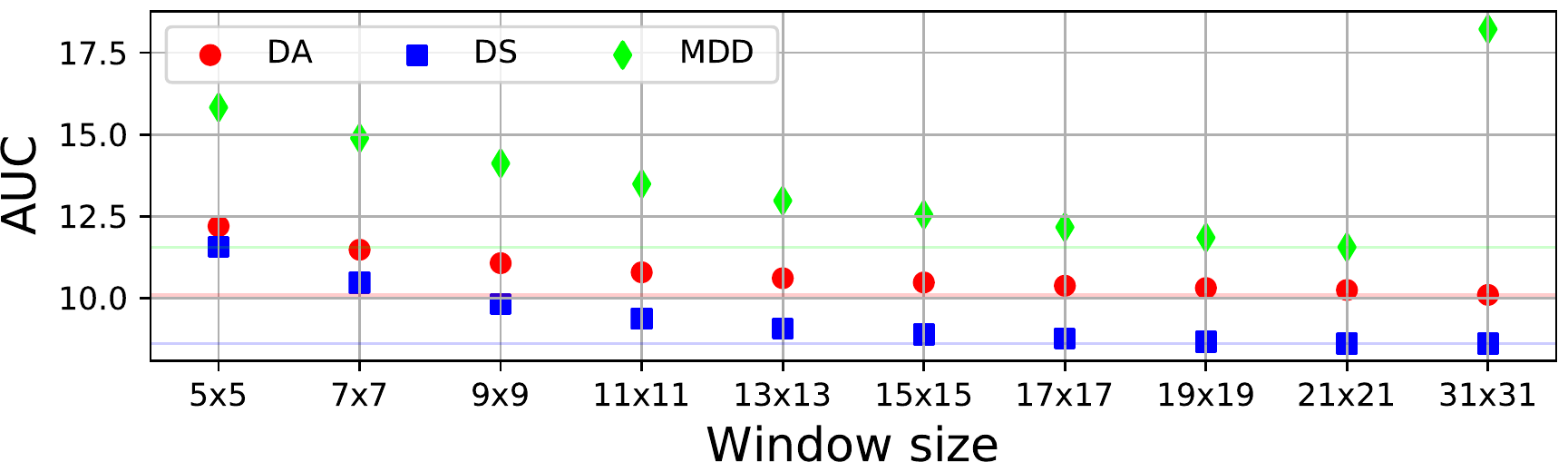} \\
         \includegraphics[width=0.38\textwidth]{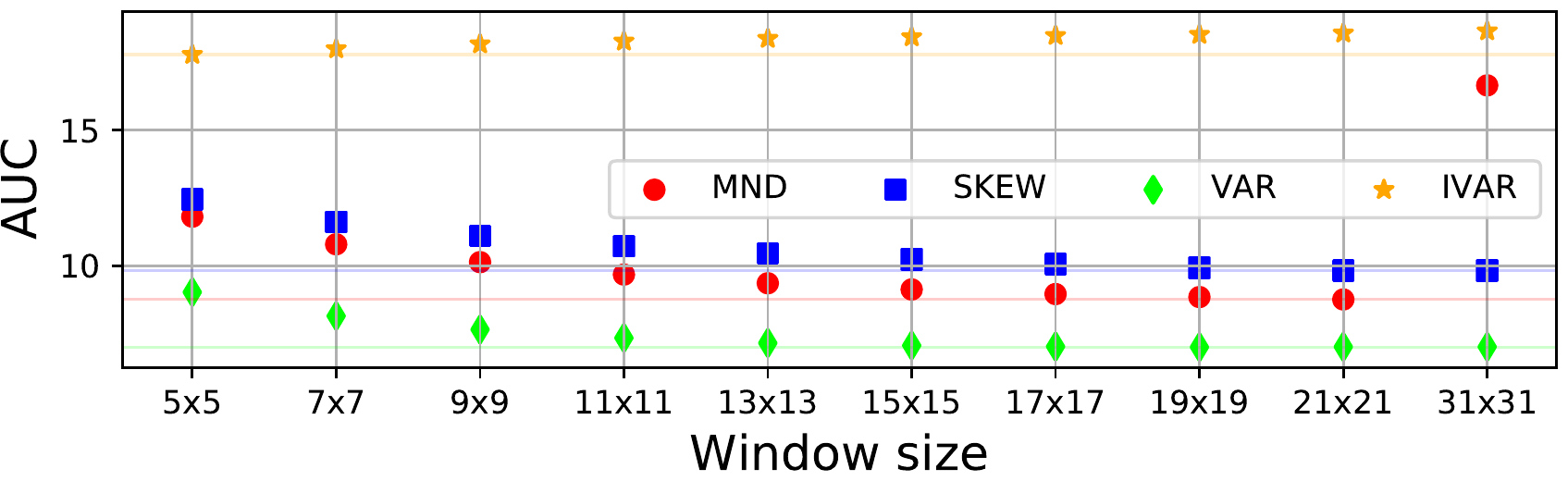} \\
         \includegraphics[width=0.38\textwidth]{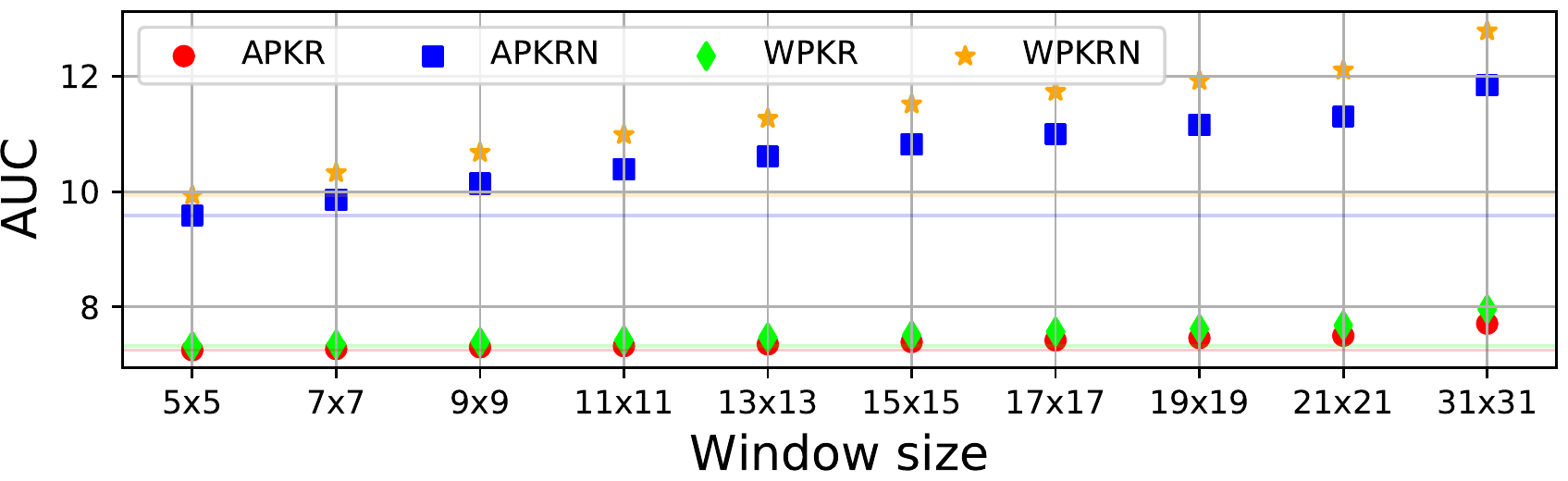} \\
    \end{tabular}
    \caption{\textbf{Impact of N(p) size}, Census-SGM algorithm.}
    \label{fig:census_sgm_radius}
\end{figure}

\textbf{Impact of the windows size.} Figure \ref{fig:census_sgm_radius} plots the AUC achieved by varying the radius of $N(p)$ for measures computed over a local neighborhood. Despite the very different outcome of the stereo algorithm deployed in this experiment, we can observe behaviors similar to the Census-CBCA case. For instance, DA and DS perform better on $31\times31$ windows, MDD and MND get the best results with a size of 21 while those based on \colorbox{localcolor}{local properties} perform better with kernels of size 5. IVAR yields the best performance with the smallest 5$\times$5 kernel.

\textbf{Learned measures, synthetic data training.} Table \ref{tab:census_sgm_learning}, on the left, collects results for learned measures when trained on synthetic images from the Driving train split. 

Concerning results on the synthetic test split, \colorbox{deepvol}{cost-volume CNNs} achieve the top-4 positions with LAF leading, followed by methods acting in the disparity domain MMC, LGC and O2. This outcome is not surprising since for disparity maps with a much lower error rate, as in the case of Census-SGM, the cost volume becomes a much more meaningful cue to detect outliers. Both \colorbox{deepdisp}{disparity CNNs} and \colorbox{forestdisp}{disparity forests} can compete only when using a very large receptive field. Compared to general-purpose methods, most of them outperform \colorbox{sgmcolor}{SGMF} specifically tailored for SGM whose ranking is 16.

Analyzing the capability of generalizing to real data, very rarely learned measures perform better than the top-1 hand-crafted measure VAR on KITTI datasets. This evidence suggests that the domain shift impacts more when dealing with more accurate stereo algorithms producing smoother disparity maps. Moreover, the effect is even more evident on Middlebury and ETH3D. Despite this outcome, interestingly, LAF keeps rank 1, showing to be the most stable solution in these experiments with LEV$_{50}$, CVA and ConfNet following.  MPN and O2 substantially keep their position moving from synthetic to real data showing a much higher generalization capability than most other methods. SGMF achieves average generalization performance yet improving its ranking from 16 to 10.

Looking at patch-based methods, PBCP$_{r}$ again better generalizes than CCNN and PBCP$_{d}$, although the latter is better on the same training domain. 
As for Census-CBCA, measures processing the reference image witness drop in ranking when crossing the domains since the RGB data is directly exposed to image content variation. However, such a drop is moderate for LGC and a notable exception is ConfNet that jumps from rank 12 to 4.

\begin{table}[t]
    \centering
    \scalebox{0.68}{
    \renewcommand{\tabcolsep}{2pt}
    \begin{tabularx}{\textwidth}{cc}
    
    \begin{tabular}{.l;c|c|c|c|c;c.}
    \toprule
    & Driv. & 2012 & 2015 & Midd. & ETH & R.\\
    \midrule
    \rowcolor{localcolor}
    APKR$_{5}$ & 14.34 & 2.64 & 2.81 & 11.06 & 5.48 & 4\\
    APKRN$_{5}$ & 21.95 & 4.13 & 4.01 & 12.04 & 5.95 & 21\\
    CUR & 23.97 & 5.51 & 4.77 & 13.05 & 6.79 & 29\\
    DAM & 29.20 & 8.67 & 8.21 & 22.67 & 13.28 & 43\\
    LC & 22.08 & 5.47 & 4.91 & 12.96 & 6.95 & 28\\
    MM & 14.47 & 2.82 & 2.83 & \underline{10.68} & 5.39 & 2\\
    MMN & 25.95 & 5.36 & 5.05 & 13.40 & 6.94 & 30\\
    MSM & 14.52 & 3.55 & 3.46 & 13.04 & 7.82 & 14\\
    NLM & 14.47 & 2.82 & 2.83 & 10.68 & 5.39 & 3\\
    NLMN & 25.95 & 5.36 & 5.05 & 13.40 & 6.94 & 31\\
    PKR & 13.85 & 2.81 & 2.92 & 11.15 & 5.60 & 5\\
    PKRN & 20.62 & 4.03 & 3.88 & 11.74 & 6.01 & 20\\
    SGE & 14.22 & 3.38 & 3.30 & 13.11 & 7.83 & 12\\
    WPKR$_{5}$ & 14.60 & 2.70 & 2.86 & 11.03 & 5.49 & 7\\
    WPKRN$_{5}$ & 21.98 & 4.64 & 4.50 & 12.52 & 6.21 & 25\\
    \rowcolor{fullcolor}
    ALM & 13.57 & 2.97 & 2.89 & 12.05 & 6.52 & 10\\
    LMN & 30.86 & 7.43 & 5.90 & 20.92 & 11.24 & 40\\
    MLM & 13.49 & 2.74 & 2.70 & 11.29 & 6.13 & 6\\
    NEM & 24.76 & 10.33 & 9.04 & 28.90 & 14.59 & 44\\
    NOI & 30.49 & 14.77 & 12.09 & 29.36 & 15.00 & 48\\
    PER & \underline{13.55} & 2.91 & 2.82 & 11.83 & 6.42 & 9\\
    PWCFA & 14.94 & 3.29 & 3.22 & 12.03 & 6.48 & 11\\
    WMN & 13.79 & 2.89 & 3.04 & 11.50 & 5.95 & 8\\
    WMNN & 18.78 & 3.61 & 3.47 & 11.44 & 5.95 & 16\\
    \rowcolor{sgmcolor}
    PS & 21.50 & 5.37 & 4.79 & 11.98 & 7.24 & 27\\
    \midrule
    \rowcolor{white}
    Opt. & 6.85 & 0.79 & 0.74 & 4.57 & 2.14 & -\\
    D1(\%) & 33.25 & 10.33 & 9.00 & 26.68 & 15.74 & -\\
    \midrule
    \end{tabular}
    &
    \begin{tabular}{.l;c|c|c|c|c;c.}
    \toprule
    & Driv. & 2012 & 2015 & Midd. & ETH & R.\\
    \midrule
    \rowcolor{dispcolor}
    DA$_{31}$ & 23.53 & 3.92 & 4.23 & 14.51 & \underline{4.39} & 26\\
    DMV & 24.59 & 4.77 & 4.67 & 18.77 & 11.10 & 34\\
    DS$_{31}$ & 18.41 & 3.05 & 3.47 & 12.80 & 5.38 & 15\\
    DTD & 14.89 & 3.61 & 3.80 & 17.67 & 8.68 & 22\\
    MDD$_{21}$ & 21.24 & 3.70 & 3.64 & 18.97 & 10.35 & 32\\
    MND$_{21}$ & 16.34 & 2.99 & 3.00 & 14.23 & 7.27 & 18\\
    SKEW$_{21}$ & 15.92 & 4.03 & 4.12 & 15.91 & 9.21 & 24\\
    VAR$_{19}$ & 13.75 & \underline{1.97} & \underline{1.92} & 12.02 & 5.37 & 1\\
    \rowcolor{lrcolor}
    ACC & 24.05 & 6.08 & 5.59 & 18.43 & 11.09 & 35\\
    LRC & 25.30 & 6.16 & 5.57 & 19.65 & 11.38 & 37\\
    LRD & 20.12 & 3.08 & 3.22 & 11.28 & 5.88 & 17\\
    UC & 24.23 & 6.28 & 5.83 & 18.79 & 11.37 & 36\\
    UCC & 14.76 & 3.48 & 3.48 & 13.04 & 7.41 & 13\\
    UCO & 26.94 & 7.44 & 6.41 & 19.95 & 11.88 & 39\\
    ZSAD & 21.59 & 9.49 & 8.14 & 19.86 & 12.59 & 38\\
    \rowcolor{distcolor}
    DTS & 37.20 & 22.53 & 6.39 & 21.17 & 15.03 & 49\\
    DSM & 16.43 & 4.50 & 2.78 & 13.66 & 7.80 & 19\\
    SAMM & 14.31 & 11.89 & 3.71 & 19.17 & 8.97 & 33\\
    \rowcolor{imagecolor}
    DB & 27.68 & 8.45 & 8.75 & 23.84 & 11.87 & 42\\
    DLB & 29.02 & 6.79 & 6.93 & 23.26 & 13.01 & 41\\
    DTE & 33.69 & 9.36 & 8.92 & 26.45 & 13.55 & 47\\
    HGM & 32.49 & 9.27 & 8.19 & 25.86 & 13.99 & 46\\
    IVAR$_{5}$ & 34.08 & 8.64 & 8.51 & 25.41 & 12.47 & 45\\
    & & & & & &  \\
    \rowcolor{sgmcolor}
    SCS & 22.13 & 3.55 & 3.79 & 13.08 & 6.54 & 23\\
    \midrule
    \rowcolor{white}
    Opt. & 6.85 & 0.79 & 0.74 & 4.57 & 2.14 & -\\
    D1(\%) & 33.25 & 10.33 & 9.00 & 26.68 & 15.74 & -\\
    \midrule
    \end{tabular}    
    \end{tabularx}
    }
    \caption{\textbf{Results with Census-SGM algorithm}, hand-crafted measures.}
    \label{tab:census_sgm_hand}
\end{table}

\begin{table}[t]
    \centering
    \scalebox{0.68}{
    \renewcommand{\tabcolsep}{2pt}
    \begin{tabularx}{\textwidth}{cc}
    
    \begin{tabular}{.l;c;c|c;c|c;c;c.}
    \multicolumn{8}{c}{Train set: Driving} \\
    \toprule
    & Driv. & 2012 & 2015 & Midd. & ETH & R. & CR.\\
    \midrule
    \rowcolor{forestvol}
    ENS$_{23}$ & 11.16 & 2.87 & 3.21 & 15.63 & 9.06 & 17 & 17\\
    GCP & 11.42 & 2.35 & 2.52 & 16.52 & 8.95 & 19 & 16\\
    LEV$_{22}$ & 10.68 & \underline{1.84} & \underline{1.99} & 16.48 & 11.33 & 13 & 20\\
    LEV$_{50}$ & 10.12 & 2.04 & 2.08 & 12.90 & 6.96 & 9 & 2\\
    FA & 10.83 & 3.63 & 3.51 & 14.28 & 9.41 & 14 & 18\\    
    \rowcolor{forestdisp}
    ENS$_{7}$ & 13.46 & 4.30 & 4.63 & 16.17 & 9.49 & 23 & 22\\
    O1 & 10.20 & 3.08 & 3.25 & 13.45 & 6.84 & 10 & 14\\
    O2 & 9.92 & 2.66 & 3.01 & 12.77 & 6.29 & 7 & 6\\
    \rowcolor{deepdisp}
    CCNN & 10.84 & 2.73 & 3.03 & 15.64 & 7.90 & 15 & 15\\
    PBCP$_{r}$ & 12.02 & 2.96 & 2.73 & 13.37 & 5.98 & 22 & 9\\
    PBCP$_{d}$ & 11.76 & 3.09 & 3.93 & 12.56 & 6.87 & 20 & 13\\
    EFN & 32.42 & 10.07 & 8.77 & 26.01 & 15.08 & 24 & 24\\
    LFN & 10.37 & 4.31 & 4.71 & 16.27 & 9.03 & 11 & 21\\
    MMC & 9.63 & 2.93 & 2.97 & 13.76 & 6.54 & 5 & 11\\
    ConfNet & 10.39 & 2.42 & 3.11 & 12.89 & 5.92 & 12 & 4\\
    LGC & 9.82 & 2.25 & 2.79 & 13.77 & 6.20 & 6 & 8\\
    \rowcolor{deepvol}
    RCN & 11.99 & 4.08 & 3.68 & 21.00 & 9.25 & 21 & 23\\
    MPN & 9.60 & 1.93 & 2.23 & 12.95 & 7.39 & 4 & 5\\
    UCN & 9.58 & 2.35 & 2.80 & 12.97 & 6.80 & 3 & 7\\
    LAF & \underline{8.41} & 2.24 & 2.54 & \underline{11.48} & 7.21 & 1 & 1\\
    ACN & 9.51 & 2.38 & 2.30 & 12.89 & 8.82 & 2 & 12\\
    CRNN  & 11.41 & 3.71 & 3.35 & 17.05 & 7.19 & 18 & 19\\
    CVA & 10.11 & 2.84 & 2.99 & 13.00 & \underline{5.18} & 8 & 3\\
    \rowcolor{sgmcolor}
    SGMF & 11.00 & 3.16 & 3.29 & 12.59 & 6.38 & 16 & 10\\    
    \midrule
    \rowcolor{white}
    Opt. & 6.85 & 0.79 & 0.74 & 4.57 & 2.14 & - & -\\
    D1(\%) & 33.25 & 10.33 & 9.00 & 26.68 & 15.74 & - & -\\
    \midrule
    \end{tabular}    
    &
    \begin{tabular}{.l;c|c;c|c;c.}
    \multicolumn{6}{c}{Train set: KITTI 2012} \\
    \toprule
    & 2012 & 2015 & Midd. & ETH & R.\\
    \midrule
    \rowcolor{forestvol}
    ENS$_{23}$ & 1.99 & 2.18 & 12.82 & 7.82 & 16\\
    GCP & 2.02 & 2.47 & 13.12 & 6.34 & 15\\
    LEV$_{22}$ & 1.79 & 1.98 & 12.20 & 6.79 & 11\\
    LEV$_{50}$ & 1.72 & 1.89 & 12.50 & 7.47 & 14\\
    FA & 2.24 & 2.35 & 13.02 & 7.55 & 18\\
    \rowcolor{forestdisp}
    ENS$_{7}$ & 2.71 & 2.86 & 15.21 & 8.71 & 21\\
    O1 & 1.69 & 1.72 & 11.40 & 7.20 & 9\\
    O2 & 1.64 & 1.55 & 10.82 & 8.08 & 10\\
    \rowcolor{deepdisp}
    CCNN & 1.90 & 1.92 & 12.01 & 7.39 & 13\\
    PBCP$_{r}$ & 1.87 & 2.03 & 10.73 & 7.33 & 8\\
    PBCP$_{d}$ & 1.92 & 2.04 & 16.27 & 14.03 & 22\\
    EFN & 2.27 & 2.14 & 13.98 & 7.53 & 20\\
    LFN & 2.02 & 2.04 & 12.74 & 8.04 & 17\\
    MMC & 1.75 & 1.69 & 11.64 & 7.89 & 12\\
    ConfNet & 2.33 & 2.27 & 13.84 & 7.08 & 19\\
    LGC & 1.80 & \underline{1.49} & 10.89 & 6.63 & 5\\
    \rowcolor{deepvol}
    RCN & 3.04 & 2.65 & 20.67 & 11.94 & 24\\
    MPN & 1.57 & 1.67 & 8.92 & \underline{5.16} & 1\\
    UCN & 1.57 & 1.62 & \underline{8.88} & 5.28 & 2\\
    LAF & \underline{1.44} & 1.60 & 10.44 & 6.44 & 4\\
    ACN & 1.70 & 1.74 & 9.20 & 5.19 & 3\\
    CRNN & 3.14 & 2.80 & 18.54 & 10.49 & 23\\
    CVA & 2.20 & 2.23 & 10.65 & 6.45 & 7\\
    \rowcolor{sgmcolor}
    SGMF & 2.16 & 2.28 & 11.04 & 5.47 & 6\\    
    \midrule
    \rowcolor{white}
    Opt. & 0.79 & 0.74 & 4.57 & 2.14 & -\\
    D1(\%) & 10.33 & 9.00 & 26.68 & 15.74 & -\\
    \midrule
    \end{tabular}    
    \end{tabularx}
    }
    \caption{\textbf{Results with Census-SGM algorithm}, learned measures.}
    \label{tab:census_sgm_learning}
\end{table}

\begin{figure*}[t]
    \scriptsize
    
    \centering
    \renewcommand{\tabcolsep}{1pt}
        \rotatebox{90}{KITTI} \rotatebox{90}{2015}
        \subfigure{
        \includegraphics[width=0.11\textwidth, frame]{images/qualitative/census-CBCA/2015/000197_10.png}}
        \subfigure{
        \includegraphics[width=0.11\textwidth, frame]{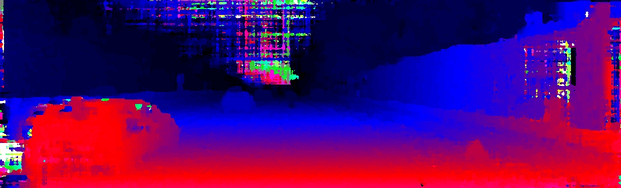}}
        \subfigure{
         \includegraphics[width=0.11\textwidth, frame]{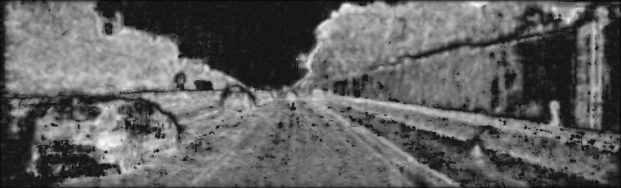}}
         \subfigure{
        \includegraphics[width=0.11\textwidth, frame]{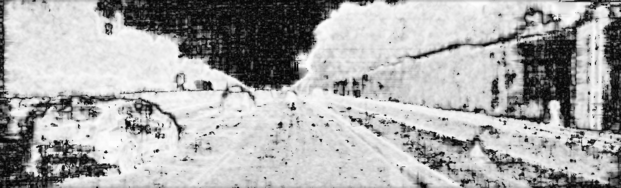}}
        \subfigure{
        \includegraphics[width=0.11\textwidth, frame]{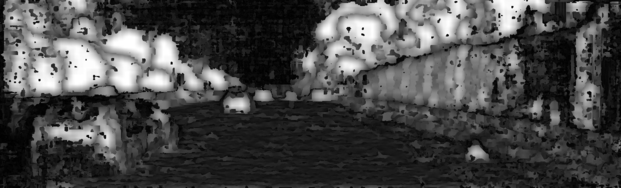}}
        \subfigure{
        \includegraphics[width=0.11\textwidth, frame]{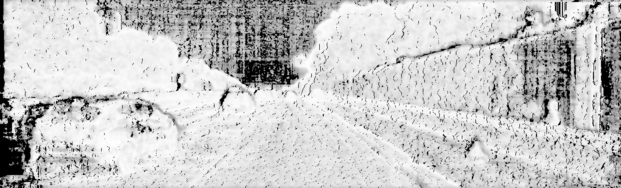}}
        \subfigure{
        \includegraphics[width=0.11\textwidth, frame]{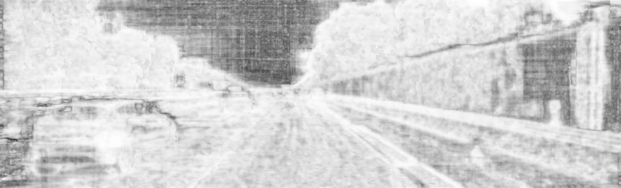}}
        \subfigure{
        \includegraphics[width=0.11\textwidth, frame]{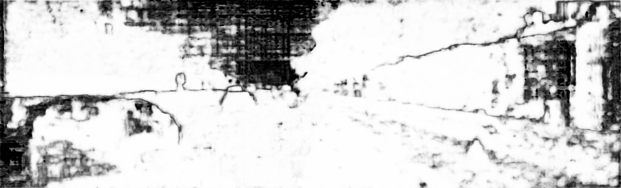}  }
        \\
        \rotatebox{90}{Middlebury}
        \subfigure{
        \includegraphics[width=0.11\textwidth, frame]{images/qualitative/census-CBCA/MiddEval3/Playtable.png}}
        \subfigure{
        \includegraphics[width=0.11\textwidth, frame]{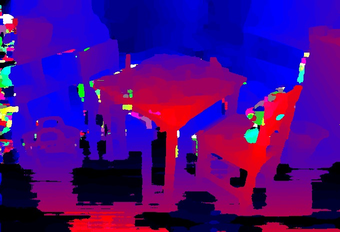}}
        \subfigure{
        \includegraphics[width=0.11\textwidth, frame]{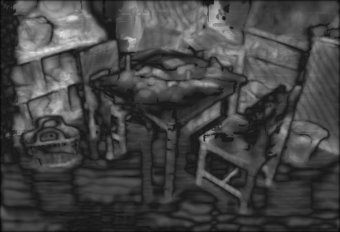}}
        \subfigure{
        \includegraphics[width=0.11\textwidth, frame]{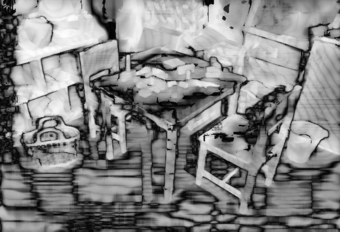}} 
        \subfigure{
        \includegraphics[width=0.11\textwidth, frame]{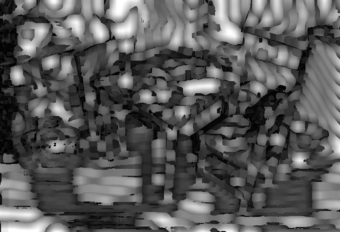}}
        \subfigure{
        \includegraphics[width=0.11\textwidth, frame]{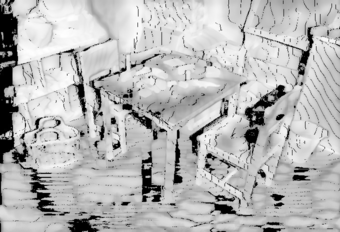}} 
        \subfigure{
        \includegraphics[width=0.11\textwidth, frame]{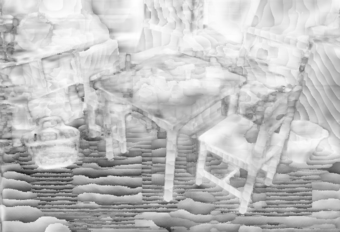}}
        \subfigure{
        \includegraphics[width=0.11\textwidth, frame]{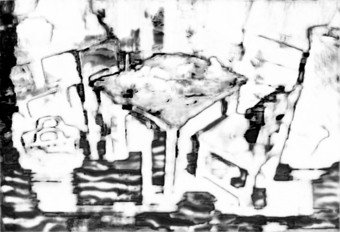}  }
        \\
    
    \caption{\textbf{Qualitative results concerning Census-SGM algorithm.} Results on KITTI 2015 and Middlebury showing a variety of confidence measures. From left to right: reference image, disparity map and confidence maps by APKR$_7$, WMN, DA$_{31}$, UCC, SAMM and LAF.} 
    \label{fig:qualitative_census_sgm}
\end{figure*}

\textbf{Learned measures, real data training.} Table \ref{tab:census_sgm_learning}, on the right, collects results for learned measures when trained on KITTI 2012 20 training images. On KITTI datasets and Middlebury, they frequently outperform VAR thanks to the much more similar domain observed during training. Specifically, this fact always occurs for O1, O2, MMC, LGC, MPN, UN, ACN and LAF. On the other hand, on ETH3D, they are competitive but often worse than top-performing hand-crafted measures. Only the overall top-performing LAF, UN and ACN, are always more effective than VAR with each dataset. This outcome confirms \colorbox{deepvol}{Cost-volume CNNs} as the most effective solution followed by LGC and \colorbox{sgmcolor}{SGMF}, the latter tailored explicitly for SGM pipelines.
Interestingly, \colorbox{forestdisp}{disparity forests} methods O1 and O2 outperform \colorbox{forestvol}{cost-volume forests}, as well as most \colorbox{deepdisp}{disparity CNNs}.
About the latter, using the right disparity map allows PBCP$_r$ to perform better than CCNN in the case of smooth disparity maps, while PBCP$_d$ still generalizes worse. Again, including the reference image and the disparity map only is effective only in the case of a large receptive field, as for MMC.

\textbf{Qualitative results.} Figure \ref{fig:qualitative_census_sgm}, as for previous qualitative results, reports an example of disparity map computed by the Census-SGM algorithm on KITTI 2015 and Middlebury and the outcome of six confidence measures, five hand-crafted and LAF in the rightmost column. Other than the different behavior of hand-crafted measures, we can notice the effects of the different domains on the learned one.
In this case, being both cost volumes and disparity maps much smoother, hand-crafted measures now assign high confidence to most pixels as well as learned measures (\ie{} LAF) do.

\textbf{Summary.} When dealing with more accurate stereo algorithms, based on SGM, the disparity map alone rarely allows for top-performing confidence estimation, except in the case of VAR. Indeed, the much smoother disparity map makes it harder to detect outliers without taking into account the cost volume. This is observed for learned measures as well, among which those processing the cost volume results more effective with fewer exceptions (when using large receptive fields -- LGC, or the right disparity map as well -- PBCP$_r$). Measures tailored to SGM results in average performance.

\begin{table}[t]
    \centering
    \scalebox{0.68}{
    \renewcommand{\tabcolsep}{2pt}
    \begin{tabularx}{\textwidth}{cc}
    
    \begin{tabular}{.l;c|c|c|c|c;c.}
    \toprule
    & Driv. & 2012 & 2015 & Midd. & ETH & R.\\
    \midrule
    \rowcolor{localcolor}
    APKR$_{5}$ & 10.88 & 0.86 & 1.57 & 6.05 & 4.03 & 4\\
    APKRN$_{5}$ & 17.75 & 2.48 & 2.73 & 8.44 & 4.82 & 22\\
    CUR & 17.14 & 2.84 & 2.93 & 9.65 & 6.62 & 29\\
    DAM & 23.16 & 5.26 & 5.19 & 17.97 & 10.73 & 44\\
    LC & 16.53 & 3.05 & 3.09 & 10.14 & 7.52 & 32\\
    MM & 10.60 & 1.11 & 1.94 & 6.95 & 5.26 & 8\\
    MMN & 17.98 & 3.08 & 3.04 & 10.10 & 6.26 & 33\\
    MSM & 13.30 & 0.88 & 1.96 & 6.09 & 3.59 & 7\\
    NLM & 10.60 & 1.11 & 1.94 & 6.95 & 5.26 & 9\\
    NLMN & 17.98 & 3.08 & 3.04 & 10.10 & 6.26 & 34\\
    PKR & 10.51 & 0.88 & 1.65 & 5.93 & 4.19 & 3\\
    PKRN & 15.34 & 2.11 & 2.38 & 8.07 & 5.05 & 16\\
    SGE & 13.12 & 0.81 & 1.86 & 5.96 & 3.41 & 6\\
    WPKR$_{5}$ & 10.91 & 0.90 & 1.64 & 6.19 & 4.11 & 5\\
    WPKRN$_{5}$ & 17.63 & 2.86 & 3.05 & 9.05 & 5.05 & 26\\
    \rowcolor{fullcolor}
    ALM & 12.45 & 1.87 & 2.69 & 12.78 & 7.79 & 25\\
    LMN & 23.28 & 3.15 & 3.75 & 11.34 & 5.73 & 35\\
    MLM & 12.22 & 1.75 & 2.58 & 12.13 & 7.53 & 21\\
    NEM & 17.66 & 4.38 & 4.43 & 18.24 & 12.03 & 41\\
    NOI & 30.79 & 12.18 & 8.88 & 26.70 & 15.41 & 49\\
    PER & 12.43 & 1.86 & 2.67 & 12.69 & 7.75 & 24\\
    PWCFA & 11.85 & 1.45 & 2.18 & 8.33 & 6.33 & 13\\
    WMN & 10.91 & 0.81 & 1.58 & \underline{5.45} & 3.33 & 1\\
    WMNN & 13.80 & 1.33 & 2.03 & 6.53 & 4.07 & 11\\
    \rowcolor{sgmcolor}
    PS & 15.83 & 2.33 & 2.83 & 9.88 & 7.46 & 27\\
    \midrule
    \rowcolor{white}
    Opt. & 4.57 & 0.25 & 0.44 & 2.94 & 1.41 & -\\
    D1(\%) & 26.92 & 6.08 & 6.03 & 21.80 & 12.59 & -\\
    \midrule
    \end{tabular}
    &
    \begin{tabular}{.l;c|c|c|c|c;c.}
    \toprule
    & Driv. & 2012 & 2015 & Midd. & ETH & R.\\
    \midrule
    \rowcolor{dispcolor}
    DA$_{15}$ & 18.31 & 2.65 & 3.18 & 7.63 & \underline{3.20} & 17\\
    DMV & 18.18 & 2.43 & 2.90 & 10.37 & 6.27 & 31\\
    DS$_{15}$ & 15.01 & 1.43 & 2.08 & 6.93 & 3.76 & 12\\
    DTD & 11.39 & 2.26 & 1.89 & 12.46 & 7.76 & 20\\
    MDD$_{21}$ & 15.82 & 2.40 & 2.61 & 12.18 & 5.93 & 28\\
    MND$_{21}$ & 12.49 & 1.62 & 1.84 & 9.86 & 4.92 & 14\\
    SKEW$_{21}$ & 12.31 & 1.97 & 2.36 & 10.84 & 7.71 & 19\\
    VAR$_{17}$ & \underline{10.11} & \underline{0.78} & \underline{1.22} & 6.97 & 3.64 & 2\\
    \rowcolor{lrcolor}
    ACC & 19.09 & 3.21 & 3.58 & 14.01 & 9.20 & 36\\
    LRC & 20.04 & 3.58 & 4.08 & 14.92 & 9.08 & 38\\
    LRD & 14.52 & 1.69 & 2.28 & 7.50 & 5.15 & 15\\
    UC & 19.31 & 3.40 & 3.70 & 14.48 & 9.60 & 37\\
    UCC & 13.10 & 1.06 & 1.99 & 6.37 & 3.70 & 10\\
    UCO & 21.17 & 4.40 & 4.60 & 15.94 & 9.74 & 40\\
    ZSAD & 15.53 & 5.39 & 4.69 & 15.94 & 10.36 & 39\\
    \rowcolor{distcolor}
    DTS & 24.67 & 15.47 & 7.42 & 28.52 & 14.30 & 48\\
    DSM & 13.35 & 6.96 & 2.98 & 7.17 & 4.67 & 18\\
    SAMM & 11.56 & 7.90 & 2.14 & 11.73 & 6.44 & 30\\
    \rowcolor{imagecolor}
    DB & 22.39 & 5.07 & 5.61 & 19.43 & 8.96 & 43\\
    DLB & 22.45 & 3.64 & 4.34 & 18.25 & 9.98 & 42\\
    DTE & 29.67 & 5.73 & 6.13 & 22.40 & 10.59 & 47\\
    HGM & 27.25 & 5.57 & 5.72 & 21.04 & 11.17 & 45\\
    IVAR$_{5}$ & 30.37 & 5.22 & 5.90 & 21.06 & 9.74 & 46\\
    & & & & & &  \\
    \rowcolor{sgmcolor}
    SCS & 16.50 & 2.40 & 3.17 & 9.11 & 5.14 & 23\\
    \midrule
    \rowcolor{white}
    Opt. & 4.57 & 0.25 & 0.44 & 2.94 & 1.41 & -\\
    D1(\%) & 26.92 & 6.08 & 6.03 & 21.80 & 12.59 & -\\
    \midrule
    \end{tabular}    
    \end{tabularx}
    }
    \caption{\textbf{Results with MCCNN-SGM algorithm}, hand-crafted measures.}
    \label{tab:mccnn_sgm_hand}
\end{table}

\subsubsection{MCCNN-SGM}

\textbf{Hand-crafted measures.} Table \ref{tab:mccnn_sgm_hand} reports the performance achieved by hand-crafted measures with the accurate MCCNN-SGM algorithm. The top-performing measure is WMN, followed by VAR$_{17}$ and PKR, all using different strategies. The outcome of this evaluation is very similar to the behavior observed in the Census-SGM experiments, confirming with SGM pipelines the excellent affinity of VAR and measures built from $c_{d_1}$ and $c_{d_2}$.
Similarly to the previous experiments, the smooth disparity maps make measures processing the \colorbox{dispcolor}{disparity map} much less effective, with DS$_{15}$ the second-best in the category and only ranking 12 overall.  On the other hand, confidences based on \colorbox{localcolor}{local properties} covers ranks from 3 to 9 while measures exploiting the \colorbox{fullcolor}{entire cost curve}, excluding WMN, shows up at positions 11 and 13 with WMNN and PWCFA. The best \colorbox{lrcolor}{left-right consistency} features rank 10 and 15, respectively, with UCC and LRD.
As for Census-SGM experiments, DSM confirms the best among \colorbox{distcolor}{self-matching} measures ranking 18 and \colorbox{sgmcolor}{SGM-specific measures} show an average performance, placing over the middle of the leaderboard. As usual, \colorbox{imagecolor}{image properties} confirms ineffective.

\begin{figure}[t]
    \centering
    \begin{tabular}{c}
         \includegraphics[width=0.38\textwidth]{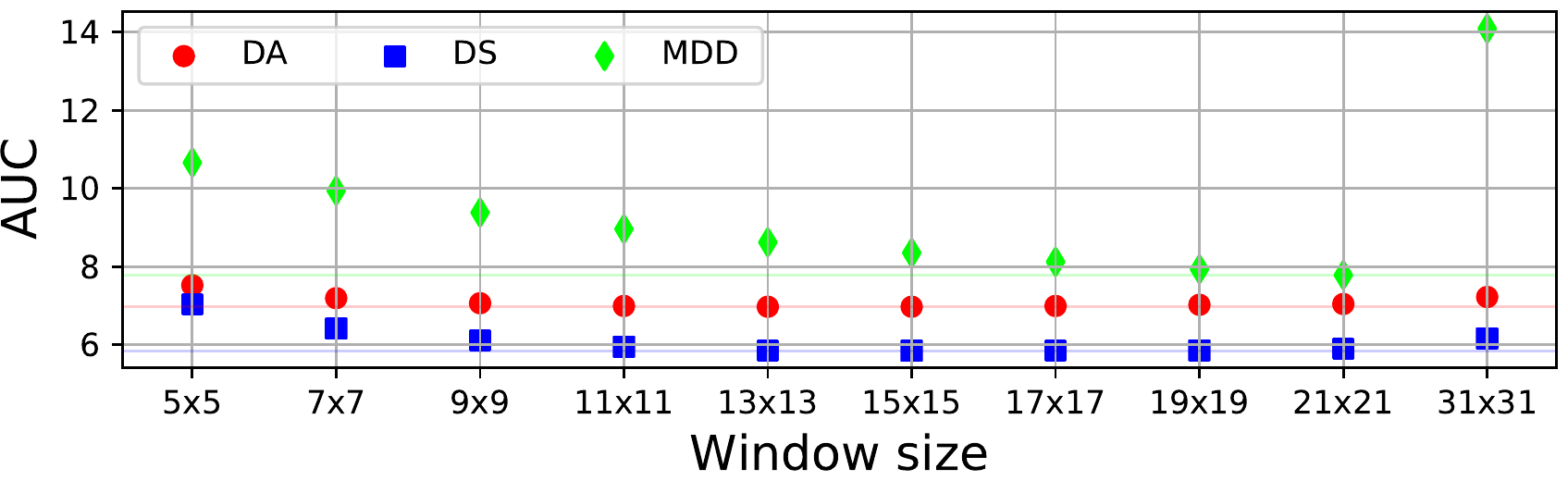} \\
         \includegraphics[width=0.38\textwidth]{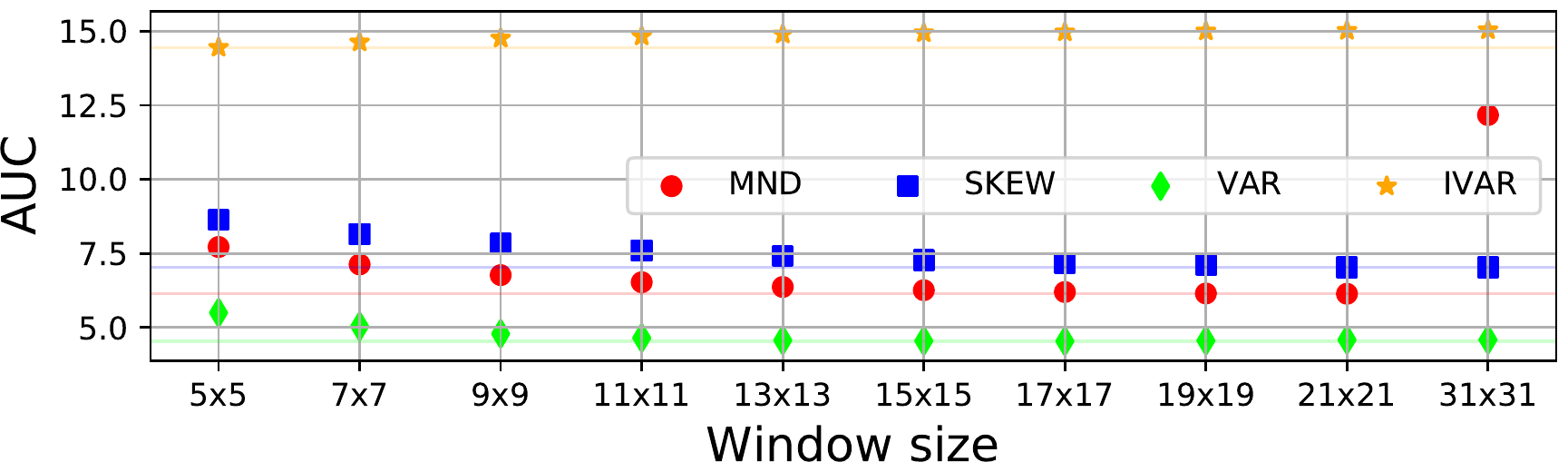} \\
         \includegraphics[width=0.38\textwidth]{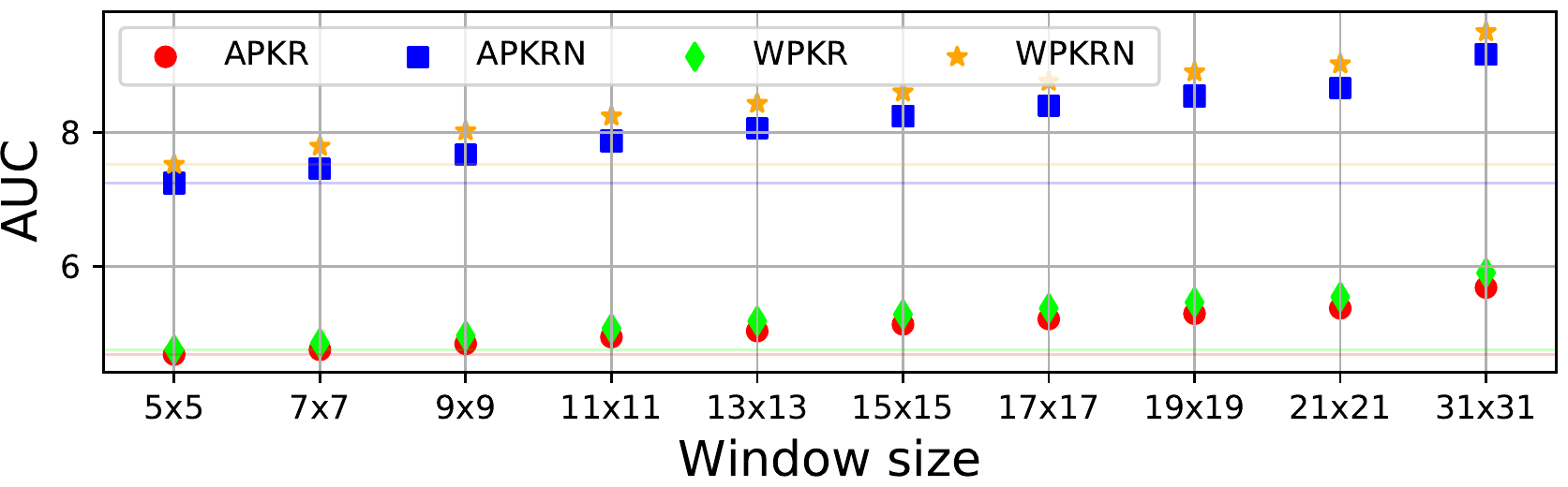} \\
    \end{tabular}
    \caption{\textbf{Impact of N(p) size}, MCCNN-SGM algorithm.}
    \label{fig:mccnn_sgm_radius}
\end{figure}

\textbf{Impact of the windows size.} Figure \ref{fig:mccnn_sgm_radius} plots the AUC achieved by varying the radius of $N(p)$ for measures computed over a local neighborhood. DA now saturates on $13\times13$ windows, 
while the other \colorbox{dispcolor}{disparity map} features mostly perform better with size 21, except for DS and VAR preferring respectively 15 and 17 windows size. All measures computed from \colorbox{localcolor}{local properties} and IVAR achieve their best results on $5\times5$ windows.

\begin{table}[t]
    \centering
    \scalebox{0.68}{
    \renewcommand{\tabcolsep}{2pt}
    \begin{tabularx}{\textwidth}{cc}
    
    \begin{tabular}{.l;c;c|c;c|c;c;c.}
    \multicolumn{8}{c}{Train set: Driving} \\
    \toprule
    & Driv. & 2012 & 2015 & Midd. & ETH & R. & CR.\\
    \midrule
    \rowcolor{forestvol}
    ENS$_{23}$ & 8.40 & 1.40 & 1.97 & 8.45 & 5.97 & 18 & 15\\
    GCP & 9.00 & 1.37 & 1.66 & 12.33 & 6.57 & 19 & 20\\
    LEV$_{22}$ & 8.29 & 1.26 & 1.71 & 7.78 & 4.83 & 17 & 11\\
    LEV$_{50}$ & 7.60 & 1.06 & 1.51 & 6.75 & 4.23 & 11 & 5\\
    FA & 7.07 & 2.54 & 2.38 & 10.60 & 6.83 & 5 & 21\\
    \rowcolor{forestdisp}
    ENS$_{7}$ & 9.41 & 2.15 & 2.59 & 9.64 & 6.11 & 21 & 18\\
    O1 & 7.48 & 1.70 & 2.09 & 8.10 & 5.04 & 10 & 14\\
    O2 & 7.34 & 1.59 & 2.15 & 7.84 & 4.82 & 9 & 13\\
    \rowcolor{deepdisp}
    CCNN & 8.05 & 1.64 & 2.05 & 10.10 & 4.74 & 16 & 16\\
    PBCP$_{r}$ & 9.74 & \underline{0.91} & \underline{1.40} & \underline{5.98} & 3.33 & 23 & 1\\
    PBCP$_{d}$ & 7.76 & 1.41 & 1.98 & 6.74 & \underline{3.25} & 15 & 4\\
    EFN & 9.40 & 3.26 & 3.52 & 11.45 & 6.42 & 20 & 22\\
    LFN & 7.67 & 2.66 & 2.96 & 10.60 & 5.34 & 13 & 19\\
    MMC & 7.24 & 1.60 & 1.99 & 8.34 & 4.26 & 6 & 12\\
    ConfNet & 7.31 & 1.16 & 1.81 & 8.40 & 3.31 & 8 & 8\\
    LGC & 7.03 & 1.15 & 1.76 & 7.56 & 3.76 & 4 & 6\\
    \rowcolor{deepvol}
    RCN & 11.40 & 1.83 & 2.86 & 15.58 & 9.46 & 24 & 24\\
    MPN & 7.30 & 1.43 & 1.60 & 5.99 & 3.85 & 7 & 2\\
    UCN & 7.01 & 1.23 & 1.63 & 6.22 & 3.80 & 3 & 3\\
    LAF & \underline{6.21} & 0.99 & 1.76 & 6.88 & 5.96 & 1 & 10\\
    ACN & 6.81 & 1.59 & 2.00 & 6.46 & 4.28 & 2 & 7\\
    CRNN & 9.54 & 1.30 & 2.21 & 16.61 & 9.38 & 22 & 23\\
    CVA & 7.62 & 1.61 & 2.23 & 9.49 & 5.36 & 12 & 17\\
    \rowcolor{sgmcolor}
    SGMF & 7.71 & 1.72 & 1.53 & 6.64 & 4.85 & 14 & 9\\
    \midrule
    \rowcolor{white}
    Opt. & 4.57 & 0.25 & 0.44 & 2.94 & 1.41 & - & -\\
    D1(\%) & 26.92 & 6.08 & 6.03 & 21.80 & 12.59 & - & -\\
    \midrule
    \end{tabular}    
    &
    \begin{tabular}{.l;c|c;c|c;c.}
    \multicolumn{6}{c}{Train set: KITTI 2012} \\
    \toprule
    & 2012 & 2015 & Midd. & ETH & R.\\
    \midrule
    \rowcolor{forestvol}
    ENS$_{23}$ & 0.82 & 1.64 & 7.94 & 5.39 & 17\\
    GCP & 1.00 & 1.91 & 7.53 & 5.61 & 18\\
    LEV$_{22}$ & 0.81 & 1.37 & 8.43 & 4.13 & 13\\
    LEV$_{50}$ & 0.75 & 1.15 & 6.46 & 3.92 & 3\\
    FA & 1.08 & 1.33 & 7.76 & 5.50 & 16\\
    \rowcolor{forestdisp}
    ENS$_{7}$ & 1.27 & 1.82 & 9.84 & 6.55 & 21\\
    O1 & 0.80 & 1.21 & 6.46 & 5.03 & 10\\
    O2 & 0.72 & \underline{1.07} & 6.24 & 5.36 & 9\\
    \rowcolor{deepdisp}
    CCNN & 0.89 & 1.22 & 7.64 & 5.11 & 14\\
    PBCP$_{r}$ & 0.86 & 1.25 & 6.09 & 5.07 & 7\\
    PBCP$_{d}$ & 1.08 & 1.44 & 12.71 & 12.18 & 22\\
    EFN & 1.27 & 1.41 & 9.52 & 4.71 & 19\\
    LFN & 0.99 & 1.17 & 7.75 & 5.35 & 15\\
    MMC & 0.93 & 1.11 & 7.01 & 4.75 & 12\\
    ConfNet & 0.87 & 1.36 & 6.93 & \underline{3.40} & 5\\
    LGC & 0.85 & 1.10 & 6.87 & 4.83 & 11\\
    \rowcolor{deepvol}
    RCN & 1.02 & 2.52 & 22.54 & 12.38 & 24\\
    MPN & 0.63 & 1.14 & 6.76 & 3.82 & 4\\
    UCN & 0.67 & 1.19 & 6.32 & 3.54 & 2\\
    LAF & \underline{0.61} & 1.21 & \underline{6.01} & 3.72 & 1\\
    ACN & 0.63 & 1.22 & 7.31 & 3.83 & 6\\
    CRNN & 0.98 & 2.24 & 21.80 & 12.20 & 23\\
    CVA & 0.77 & 1.49 & 8.61 & 6.06 & 20\\
    \rowcolor{sgmcolor}
    SGMF & 0.83 & 1.83 & 6.46 & 4.25 & 8\\
    \midrule
    \rowcolor{white}
    Opt. & 0.25 & 0.44 & 2.94 & 1.41 & -\\
    D1(\%) & 6.08 & 6.03 & 21.80 & 12.59 & -\\
    \midrule
    \end{tabular}    
    \end{tabularx}
    }
    \caption{\textbf{Results with MCCNN-SGM algorithm}, learned measures.}
    \label{tab:mccnn_sgm_learning}
\end{table}

\begin{figure*}[t]
    \scriptsize
    
    \centering
    \renewcommand{\tabcolsep}{1pt}
        \rotatebox{90}{KITTI} \rotatebox{90}{2015}
        \subfigure{
        \includegraphics[width=0.11\textwidth, frame]{images/qualitative/census-CBCA/2015/000197_10.png}}
        \subfigure{
        \includegraphics[width=0.11\textwidth, frame]{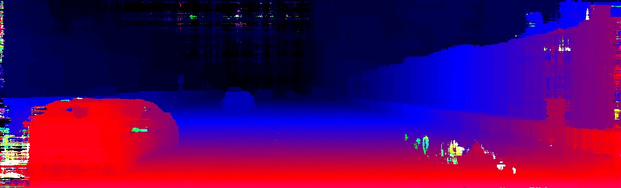}}
        \subfigure{
         \includegraphics[width=0.11\textwidth, frame]{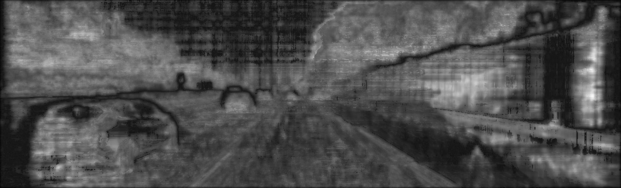}}
         \subfigure{
        \includegraphics[width=0.11\textwidth, frame]{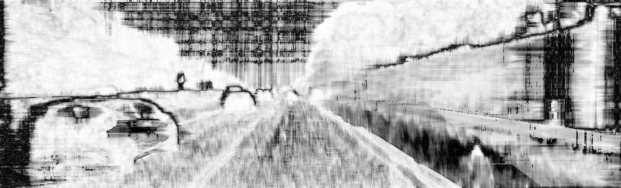}}
        \subfigure{
        \includegraphics[width=0.11\textwidth, frame]{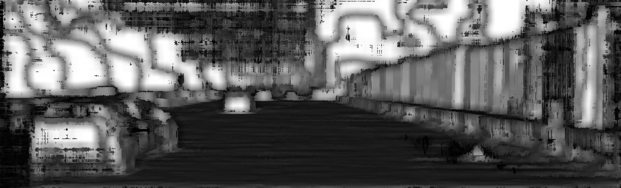}}
        \subfigure{
        \includegraphics[width=0.11\textwidth, frame]{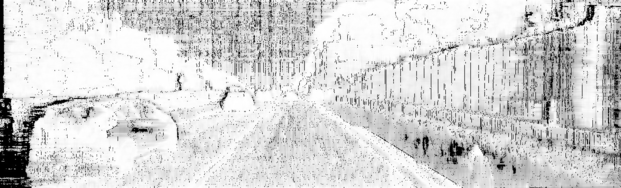}}
        \subfigure{
        \includegraphics[width=0.11\textwidth, frame]{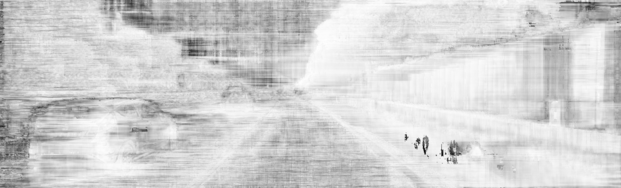}}
        \subfigure{
        \includegraphics[width=0.11\textwidth, frame]{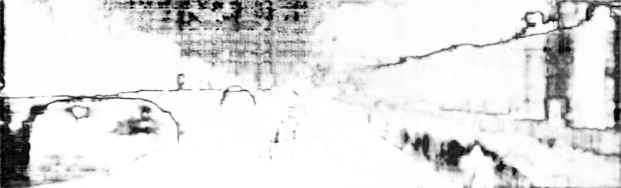}  }
        \\
        \rotatebox{90}{Middlebury}
        \subfigure{
        \includegraphics[width=0.11\textwidth, frame]{images/qualitative/census-CBCA/MiddEval3/Playtable.png}}
        \subfigure{
        \includegraphics[width=0.11\textwidth, frame]{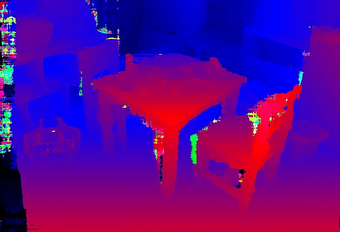}}
        \subfigure{
        \includegraphics[width=0.11\textwidth, frame]{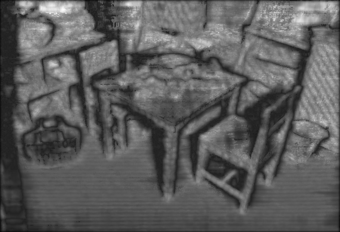}}
        \subfigure{
        \includegraphics[width=0.11\textwidth, frame]{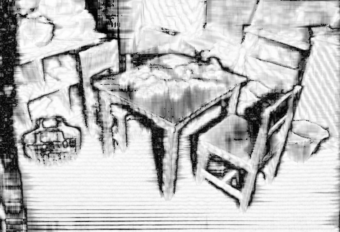}} 
        \subfigure{
        \includegraphics[width=0.11\textwidth, frame]{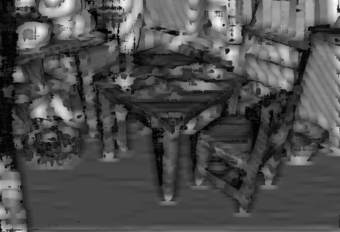}}
        \subfigure{
        \includegraphics[width=0.11\textwidth, frame]{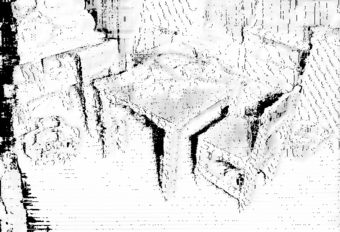}} 
        \subfigure{
        \includegraphics[width=0.11\textwidth, frame]{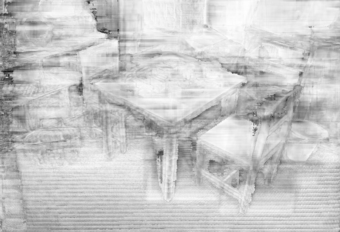}}
        \subfigure{
        \includegraphics[width=0.11\textwidth, frame]{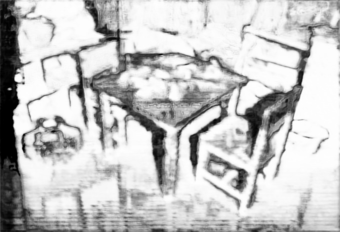}  }
        \\

    \caption{\textbf{Qualitative results concerning MCCNN-SGM algorithm.} Results on KITTI 2015 and Middlebury showing a variety of confidence measures. From top left to bottom right: reference image, disparity map and confidence maps by APKR$_7$, WMN, DA$_{31}$, UCC, SAMM and LAF.} 
    \label{fig:qualitative_mccnn_sgm}
\end{figure*}

\textbf{Learned measures, synthetic data training.} Table \ref{tab:mccnn_sgm_learning}, on the left, collects results for learned measures when trained on synthetic images from the Driving train split. 

As for Census-SGM, on the synthetic test split, LAF performs the best, followed by ACN and UN  all of the \colorbox{deepvol}{cost-volume CNNs}. Close follow-up methods are LGC and FA, respectively a \colorbox{deepdisp}{disparity CNN} and a \colorbox{forestdisp}{disparity forest}. 
In general, we can notice two opposite trends with CNNs and forests. When processing the cost volume, the former class is typically more effective, while forest-based methods processing the disparity map (\eg, O1 and O2) yield better results than those working in the cost volume space (\eg, LEV$_{50}$, LEV and GCP).
Focusing on patch-based models, the left-right consistency enforced by PBCP$_{d}$ allows it to outperform CCNN, while PBCP$_{r}$  results less effective than both. Finally, \colorbox{sgmcolor}{SGMF} ranks in the lower half of the leaderboard, yielding, for instance, better accuracy than the majority of \colorbox{forestvol}{cost-volume forests}.

Concerning generalization to real data, as for Census-SGM, the impact of domain shift is more significant. This time, none of the learned measures outperform the top-performing hand-crafted one WMN on KITTI 2012. On the other hand, this occurs in only three cases on KITTI 2015 (SGMF, LEV$_{50}$ and PBCP$_r$) and ETH3D (both PBCBs and ConfNet). In contrast, WMN is always more effective than any learned method on the Middlebury dataset. Conversely to Census-SGM, LAF loses rank 1, dropping to 10 in favor of PBCP$_{r}$. This outcome and the fourth place achieved by PBCP$_{d}$ highlights that the information from the right disparity map is highly impactful for MCCNN-SGM, a stereo algorithm producing very smooth disparity maps. In particular, enforcing left-right consistency allows for more substantial generalization even on Middlebury and ETH3D. Again, PBCP$_r$ better generalizes than PBCP$_d$. 
Most \colorbox{forestvol}{cost-volume forests} gain positions, conversely to \colorbox{forestdisp}{disparity forests} and \colorbox{deepdisp}{disparity CNNs}. In general, most \colorbox{deepvol}{cost-volume CNNs} keep their rankings with LAF, ACN and CVA notable negative exceptions and MPN a positive one. Finally, \colorbox{sgmcolor}{SGMF} improves its position from 14 to 9 confiming its effectiveness with SGM-based stereo methods.

\textbf{Learned measures, real data training.} Table \ref{tab:mccnn_sgm_learning}, on the right, collects results for learned measures when trained on KITTI 2012 20 training images. 
We can notice that many of them now outperform the top-1 hand-crafted measure on the KITTI 2012 dataset, and more frequently on KITTI 2015, thanks to the much more similar domain observed during training. Specifically, the following measures, belonging to four different categories, are more effective than WMN on both KITTI datasets: LEV$_{50}$, O1, O2, MPN, UN, LAF, ACN and CVA. In contrast, none of the learned measures achieves better accuracy than WMN on Middlebury and ETH3D.  
In summary, \colorbox{deepvol}{cost-volume CNNs} confirm to be the most effective solution, with LAF and UN covering the top-2 positions, followed by LEV$_{50}$, MPN and ConfNet.

O1 and O2, \colorbox{forestdisp}{disparity forests}, results the best in their category and better than all \colorbox{forestvol}{cost-volume forests} except LEV$_{50}$. Overall, in this experiment, SGMF achieves an accuracy slightly better than O1 and O2. 
Concerning patch-based networks, PBCP$_r$ once again stands as the best choice in these experiments. As already witnessed with the previous Census-SGM, \colorbox{deepdisp}{disparity CNNs} are rarely effective although ConfNet ranks 5.

\textbf{Qualitative results.} Finally, as for previous qualitative results, Figure \ref{fig:qualitative_mccnn_sgm} shows an example of disparity maps from KITTI 2015 and Middlebury computed with the MCCNN-SGM algorithm and the output of six confidence maps, five hand-crafted and one learned (rightmost column).
As observed in the case of Census-SGM, most measures assign high confidence to most pixels, correctly finding out that the amount of outliers in the disparity maps is very low.

\textbf{Summary.} As observed for Census-SGM, the cost volume becomes a precious source of information to estimate confidence. Measures processing the disparity map alone rarely ranks on top of the leaderboard. A similar trend is observed, again, for learned measures as well that can properly learn to estimate confidence from the disparity map when processing a large receptive field or the right disparity map as well, while measures tailored to SGM confirm their average effectiveness among all methods.

\subsection{GANet}

To conclude our evaluation, we report experiments carried out with GANet to highlight how the final volumes produced by 3D neural networks for stereo can be converted into costs, allowing for the deployment of traditional and learned measures. Such an evaluation, using volumes from a deep neural network, is performed here for the first time.

\textbf{Hand-crafted measures.} Table \ref{tab:ganet_hand} shows the performance achieved by hand-crafted measures.
At first, we can notice how measures processing the \colorbox{dispcolor}{disparity map} performs much worse in this case. All of them dropping their rank below 20. We ascribe this fact to the extremely smooth disparity maps delivered by GANet, making it extremely hard to find outliers by only looking at disparity distributions. 
Most of top-20 positions mix measures processing \colorbox{localcolor}{local properties} or the \colorbox{fullcolor}{entire cost curve}. In particular, we point out the excellent performance achieved by MSM, reaching rank 1. We ascribe this fact to the soft-argmax operator used during training, forcing the output volume to have a strong maximum (converted to minima in our experiments). The results achieved by MSM suggest that the network itself produces weaker maxima when it is less certain about the predicted disparity. Other classic measures perform very well, such as ALM and MLM, rarely ranking in the top-10 positions in the previous experiments.
Concerning measures with \textit{naive} variants, for the first time, some of them perform better than the original counterpart, such as PKRN, NLMN, and MMN. Probably, as another effect of the soft-argmax operator used during training.
UCC is the first measure leveraging the \colorbox{lrcolor}{left-right consistency} and reaches rank 8,  while within \colorbox{distcolor}{self-matching} measures SAMM achieves the best results and ranks 20.
Finally, measures based on \colorbox{imagecolor}{image properties} confirm ineffective as in any previous experiment.

\begin{table}[t]
    \centering
    \scalebox{0.68}{
    \renewcommand{\tabcolsep}{2pt}
    \begin{tabularx}{\textwidth}{cc}
    
    \begin{tabular}{.l;c|c|c|c|c;c.}
    \toprule
    & Driv. & 2012 & 2015 & Midd. & ETH & R.\\
    \midrule
    \rowcolor{localcolor}
    APKR$_{5}$ & 11.45 & \underline{2.48} & \underline{3.91} & \underline{12.82} & \underline{3.47} & 11\\
    APKRN$_{5}$ & 10.75 & 3.61 & 5.41 & 15.21 & 4.71 & 17\\
    CUR & 5.96 & 2.74 & 4.38 & 14.08 & 4.01 & 5\\
    DAM & 16.10 & 8.33 & 9.69 & 27.82 & 10.50 & 40\\
    LC & \underline{5.42} & 2.88 & 4.59 & 16.67 & 6.43 & 12\\
    MM & 17.98 & 8.96 & 10.54 & 26.08 & 9.81 & 43\\
    MMN & 8.35 & 3.87 & 5.64 & 17.09 & 5.33 & 18\\
    MSM & 6.71 & 2.69 & 4.33 & 13.47 & 3.65 & 1\\
    NLM & 17.97 & 8.95 & 10.53 & 26.07 & 9.81 & 42\\
    NLMN & 8.35 & 3.87 & 5.64 & 17.09 & 5.33 & 19\\
    PKR & 21.75 & 11.67 & 12.87 & 33.87 & 17.70 & 47\\
    PKRN & 8.09 & 3.58 & 5.40 & 15.73 & 4.88 & 14\\
    SGE & 7.11 & 2.60 & 4.26 & 13.37 & 4.16 & 7\\
    WPKR$_{5}$ & 10.55 & 2.61 & 4.28 & 13.03 & 3.59 & 10\\
    WPKRN$_{5}$ & 9.72 & 3.83 & 5.76 & 15.49 & 4.77 & 16\\
    \rowcolor{fullcolor}
    ALM & 6.71 & 2.69 & 4.33 & 13.47 & 3.65 & 2\\
    LMN & 16.61 & 8.12 & 9.52 & 23.78 & 7.85 & 36\\
    MLM & 6.71 & 2.69 & 4.33 & 13.47 & 3.65 & 3\\
    NEM & 6.56 & 2.67 & 4.33 & 13.73 & 4.08 & 6\\
    NOI & 10.46 & 5.66 & 6.98 & 23.10 & 8.69 & 25\\
    PER & 6.71 & 2.69 & 4.33 & 13.47 & 3.65 & 4\\
    PWCFA & 7.14 & 2.72 & 4.20 & 14.48 & 3.59 & 9\\
    WMN & 16.04 & 8.31 & 9.89 & 26.81 & 10.72 & 39\\
    WMNN & 8.05 & 3.53 & 5.36 & 15.43 & 4.79 & 13\\
    \midrule
    \rowcolor{white}
    Opt. & 1.69 & 0.57 & 0.85 & 5.28 & 0.99 & -\\
    D1(\%) & 16.66 & 8.62 & 10.02 & 28.61 & 10.80 & -\\
    \midrule
    \end{tabular}
    &
    \begin{tabular}{.l;c|c|c|c|c;c.}
    \toprule
    & Driv. & 2012 & 2015 & Midd. & ETH & R.\\
    \midrule
    \rowcolor{dispcolor}
    DA$_{11}$ & 12.85 & 7.85 & 9.21 & 23.41 & 6.24 & 29\\
    DMV & 16.84 & 7.16 & 8.55 & 24.39 & 8.84 & 35\\
    DS$_{15}$ & 12.95 & 6.44 & 7.68 & 23.32 & 6.63 & 26\\
    DTD & 13.21 & 7.61 & 9.17 & 27.13 & 7.88 & 34\\
    MDD$_{21}$ & 11.44 & 6.42 & 7.31 & 25.02 & 9.20 & 28\\
    MND$_{21}$ & 11.04 & 4.91 & 5.50 & 23.74 & 8.90 & 24\\
    SKEW$_{21}$ & 9.90 & 3.95 & 4.51 & 22.57 & 8.23 & 23\\
    VAR$_{21}$ & 7.08 & 3.42 & 4.26 & 22.52 & 6.74 & 21\\
    \rowcolor{lrcolor}
    ACC & 13.89 & 6.59 & 8.18 & 24.20 & 9.45 & 30\\
    LRC & 13.53 & 6.98 & 8.09 & 25.39 & 9.86 & 32\\
    LRD & 8.06 & 3.53 & 5.22 & 16.67 & 5.17 & 15\\
    UC & 13.81 & 6.80 & 8.45 & 24.39 & 9.61 & 31\\
    UCC & 7.02 & 2.82 & 4.45 & 13.93 & 3.67 & 8\\
    UCO & 15.33 & 7.77 & 8.84 & 26.95 & 10.01 & 38\\
    ZSAD & 7.28 & 7.23 & 8.62 & 24.53 & 9.57 & 27\\
    \rowcolor{distcolor}
    DTS & 18.87 & 12.36 & 12.14 & 34.33 & 14.69 & 46\\
    DSM & 9.26 & 4.60 & 7.11 & 16.58 & 9.15 & 22\\
    SAMM & 8.01 & 3.63 & 5.11 & 19.12 & 6.29 & 20\\
    \rowcolor{imagecolor}
    DB & 12.62 & 7.63 & 10.26 & 26.10 & 7.93 & 33\\
    DLB & 16.34 & 7.05 & 9.09 & 26.50 & 9.35 & 37\\
    DTE & 21.34 & 8.19 & 9.95 & 28.04 & 10.01 & 44\\
    HGM & 18.33 & 8.10 & 9.49 & 27.02 & 9.81 & 41\\
    IVAR$_{5}$ & 23.52 & 8.08 & 9.99 & 26.71 & 10.01 & 45\\
    & & & & & &  \\
    \midrule
    \rowcolor{white}
    Opt. & 1.69 & 0.57 & 0.85 & 5.28 & 0.99 & -\\
    D1(\%) & 16.66 & 8.62 & 10.02 & 28.61 & 10.80 & -\\
    \midrule
    \end{tabular}    
    \end{tabularx}
    }
    \caption{\textbf{Results with GANet model}, hand-crafted measures.}
    \label{tab:ganet_hand}
\end{table}

\begin{figure}[t]
    \centering
    \begin{tabular}{c}
         \includegraphics[width=0.38\textwidth]{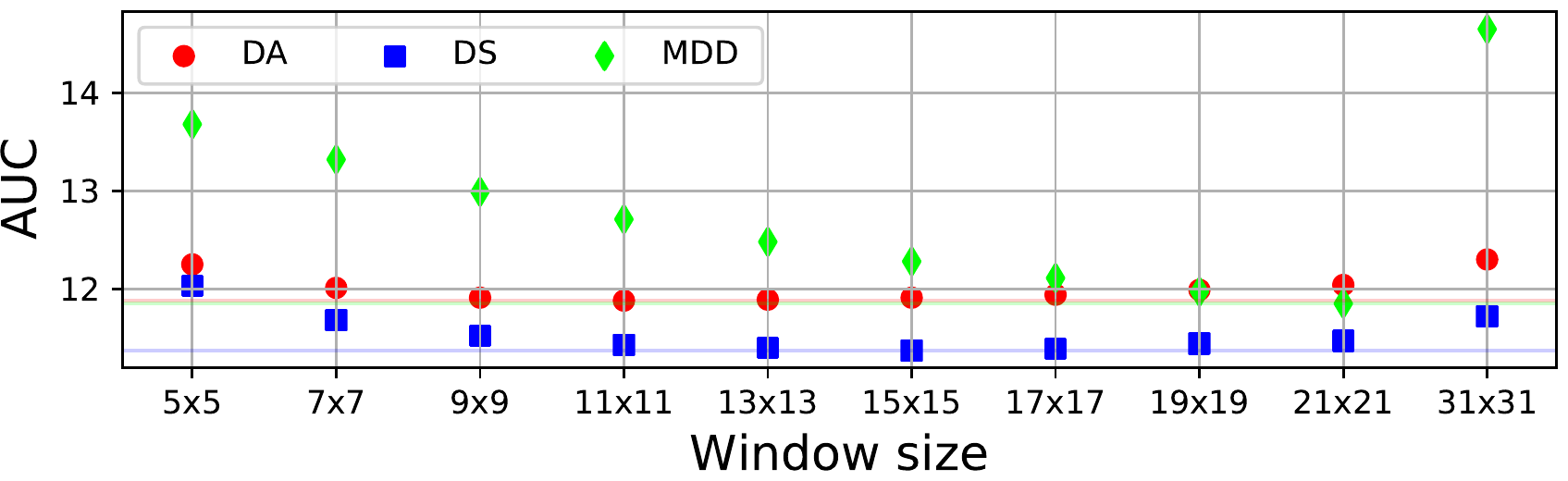} \\
         \includegraphics[width=0.38\textwidth]{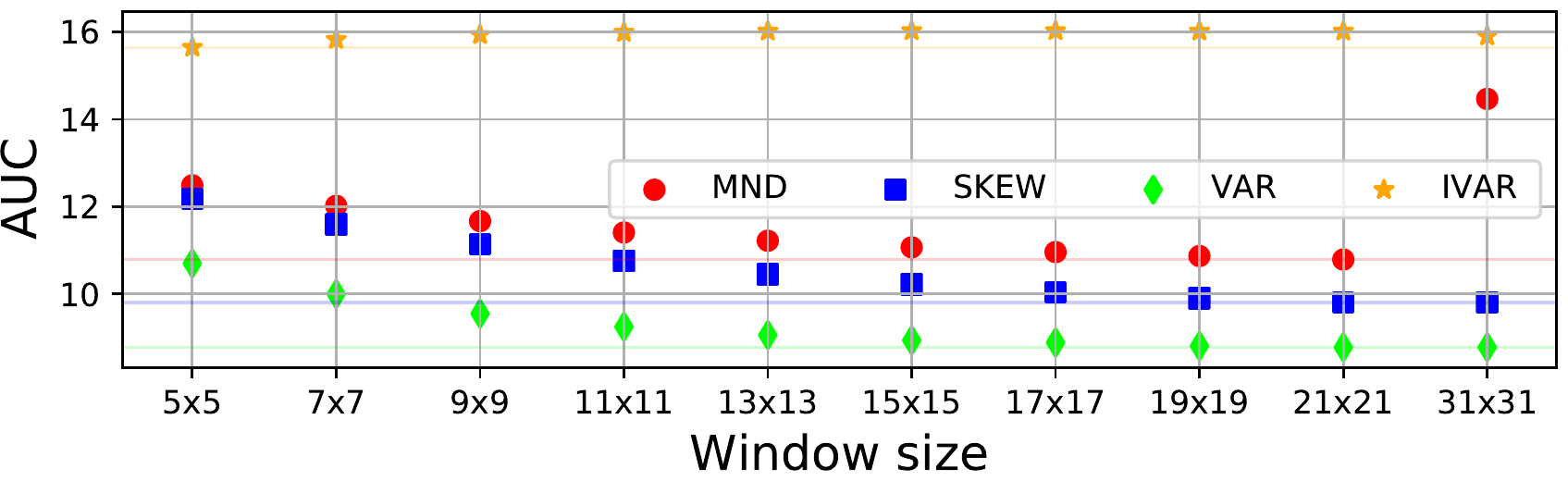} \\
         \includegraphics[width=0.38\textwidth]{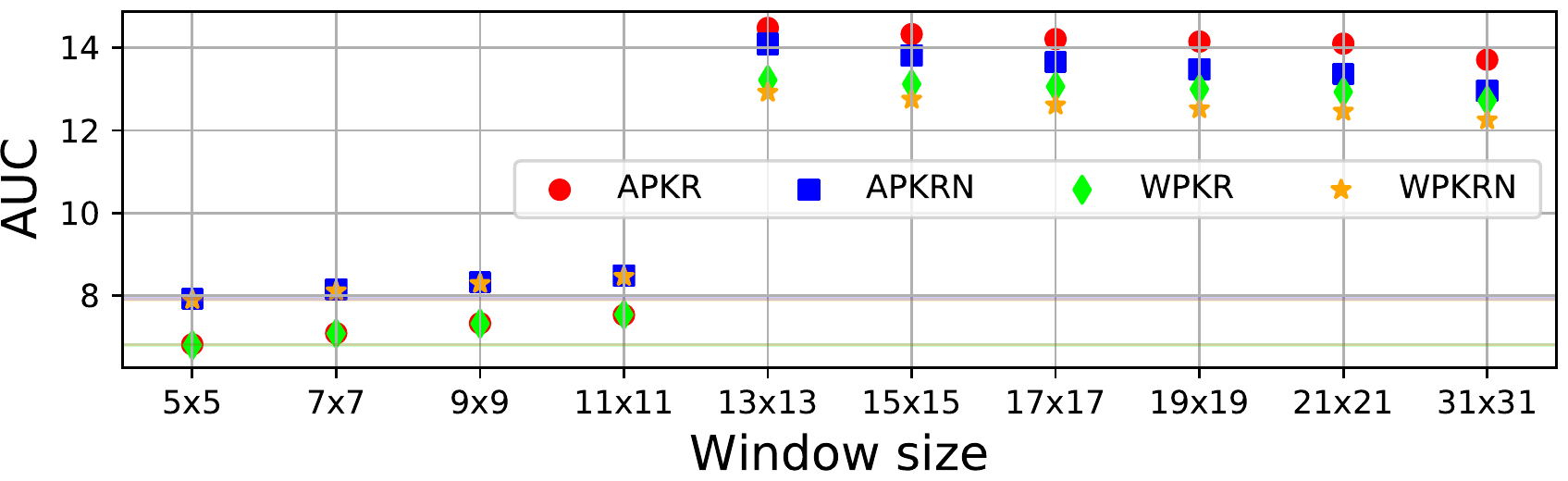} \\
    \end{tabular}
    \caption{\textbf{Impact of N(p) size}, GANet model.}
    \label{fig:ganet_radius}
\end{figure}

\begin{table}[t]
    \centering
    \scalebox{0.68}{
    \renewcommand{\tabcolsep}{2pt}
    \begin{tabularx}{\textwidth}{cc}
    
    \begin{tabular}{.l;c;c|c;c|c;c;c.}
    \multicolumn{8}{c}{Train set: Driving} \\
    \toprule
    & Driv. & 2012 & 2015 & Midd. & ETH & R. & CR. \\
    \midrule
    \rowcolor{forestvol}
    ENS$_{23}$ & 3.48 & 4.15 & 4.85 & 27.53 & 8.58 & 6 & 12\\
    GCP & 5.35 & 3.60 & 5.07 & 20.68 & 7.35 & 19 & 6\\
    LEV$_{22}$ & 3.67 & 4.63 & 4.66 & 30.98 & 7.62 & 10 & 14\\
    LEV$_{50}$ & 3.49 & 3.18 & 4.96 & 20.73 & 6.19 & 7 & 4\\
    FA & 3.41 & 4.54 & 5.63 & 27.81 & 10.11 & 4 & 16\\
    \rowcolor{forestdisp}
    ENS$_{7}$ & 5.75 & 7.09 & 8.43 & 25.49 & 9.19 & 20 & 19\\
    O1 & 3.62 & 7.65 & 9.66 & 25.63 & 8.68 & 9 & 21\\
    O2 & 3.55 & 7.53 & 9.80 & 25.38 & 8.05 & 8 & 20\\
    \rowcolor{deepdisp}
    CCNN & 4.33 & 4.78 & 6.72 & 24.71 & 9.75 & 14 & 13\\
    PBCP$_{r}$ & 4.81 & 3.51 & 4.28 & 23.48 & 7.95 & 17 & 8\\
    PBCP$_{d}$ & 4.48 & 5.69 & 6.96 & 22.55 & 8.56 & 15 & 10\\
    EFN & 8.91 & 7.20 & 9.38 & 27.05 & 10.48 & 23 & 22\\
    LFN & 5.88 & 5.92 & 8.08 & 26.35 & 9.16 & 22 & 18\\
    MMC & 5.19 & 5.04 & 6.78 & 24.07 & 8.22 & 18 & 11\\
    ConfNet & 5.85 & 6.33 & 8.38 & 29.12 & 11.12 & 21 & 23\\
    LGC & 3.72 & 5.73 & 7.43 & 25.59 & 9.28 & 11 & 15\\
    \rowcolor{deepvol}
    RCN & 4.61 & 3.63 & 5.38 & \underline{16.56} & \underline{5.18} & 16 & 1\\
    MPN & \underline{3.21} & \underline{2.71} & \underline{4.06} & 20.19 & 6.43 & 1 & 3\\
    UCN & 3.76 & 3.27 & 4.98 & 21.02 & 6.49 & 12 & 5\\
    LAF & 3.46 & 3.81 & 4.94 & 22.64 & 7.32 & 5 & 7\\
    ACN & 3.25 & 3.72 & 5.58 & 23.64 & 7.42 & 2 & 9\\
    CRNN & 3.32 & 3.20 & 5.28 & 17.93 & 6.29 & 3 & 2\\
    CVA & 3.81 & 5.54 & 7.92 & 26.11 & 9.92 & 13 & 17\\
    \midrule
    \rowcolor{white}
    Opt. & 1.69 & 0.57 & 0.85 & 5.28 & 0.99 & - & -\\
    D1(\%) & 16.66 & 8.62 & 10.02 & 28.61 & 10.80 & - & -\\
    \midrule
    \end{tabular}    
    &
    \begin{tabular}{.l;c|c;c|c;c.}
    \multicolumn{6}{c}{Train set: KITTI 2012} \\
    \toprule
    & 2012 & 2015 & Midd. & ETH & R.\\
    \midrule
    \rowcolor{forestvol}
    ENS$_{23}$ & 3.18 & 4.52 & 18.28 & 5.58 & 7\\
    GCP & 4.19 & 5.39 & 21.00 & 5.24 & 9\\
    LEV$_{22}$ & 2.26 & 3.32 & 17.58 & \underline{3.70} & 1\\
    LEV$_{50}$ & 1.99 & 2.87 & 17.87 & 4.17 & 2\\
    FA & 4.59 & 5.68 & 23.09 & 7.92 & 20\\
    \rowcolor{forestdisp}
    ENS$_{7}$ & 5.05 & 6.00 & 22.46 & 6.37 & 18\\
    O1 & 3.20 & 4.02 & 22.54 & 7.36 & 11\\
    O2 & 2.76 & 3.67 & 22.03 & 7.10 & 8\\
    \rowcolor{deepdisp}
    CCNN & 3.29 & 3.93 & 23.34 & 7.02 & 14\\
    PBCP$_{r}$ & 4.02 & 4.93 & 19.83 & 8.59 & 13\\
    PBCP$_{d}$ & 3.46 & 4.14 & 22.50 & 12.56 & 22\\
    EFN & 4.84 & 5.36 & 23.83 & 4.96 & 17\\
    LFN & 3.77 & 4.25 & 23.00 & 6.13 & 12\\
    MMC & 3.84 & 4.49 & 23.89 & 5.45 & 16\\
    ConfNet & 5.46 & 5.21 & 22.31 & 4.69 & 15\\
    LGC & 3.53 & 4.61 & 22.42 & 6.31 & 10\\
    \rowcolor{deepvol}
    RCN & 2.81 & 3.82 & \underline{16.12} & 6.04 & 4\\
    MPN & 3.89 & 4.40 & 26.64 & 12.67 & 23\\
    UCN & 2.62 & 3.18 & 23.31 & 12.23 & 21\\    
    LAF & \underline{1.70} & \underline{2.58} & 18.19 & 7.40 & 5\\
    ACN & 2.17 & 2.96 & 17.99 & 6.89 & 6\\
    CRNN & 2.41 & 3.20 & 16.33 & 6.18 & 3\\
    CVA & 4.00 & 4.90 & 23.12 & 8.75 & 19\\    
    \midrule
    \rowcolor{white}
    Opt. & 0.57 & 0.85 & 5.28 & 0.99 & -\\
    D1(\%) & 8.62 & 10.02 & 28.61 & 10.80 & -\\
    \midrule
    \end{tabular}    
    \end{tabularx}
    }
    \caption{\textbf{Results with GANet}, learned measures.}
    \label{tab:ganet_learning}
\end{table}

\begin{figure*}[t]
    \scriptsize
    
    \centering
    \renewcommand{\tabcolsep}{1pt}
        \rotatebox{90}{KITTI} \rotatebox{90}{2015}
        \subfigure{
        \includegraphics[width=0.11\textwidth, frame]{images/qualitative/census-CBCA/2015/000197_10.png}}
        \subfigure{
        \includegraphics[width=0.11\textwidth, frame]{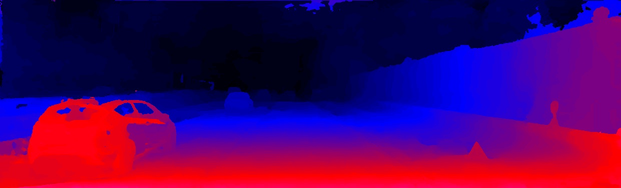}}
        \subfigure{
         \includegraphics[width=0.11\textwidth, frame]{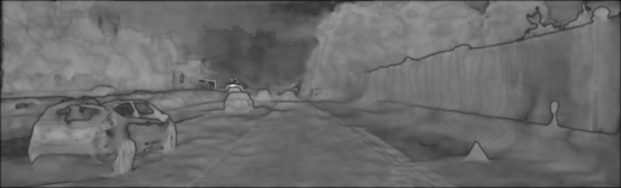}}
         \subfigure{
        \includegraphics[width=0.11\textwidth, frame]{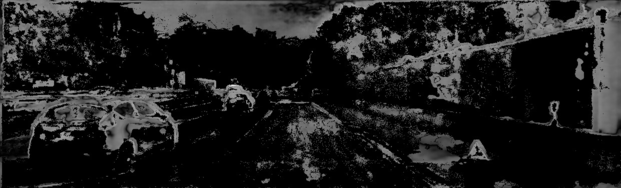}}
        \subfigure{
        \includegraphics[width=0.11\textwidth, frame]{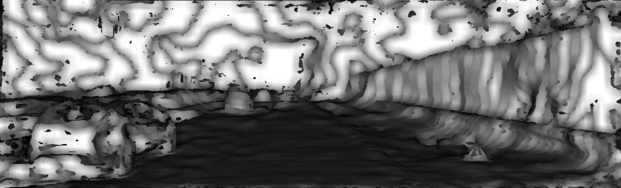}}
        \subfigure{
        \includegraphics[width=0.11\textwidth, frame]{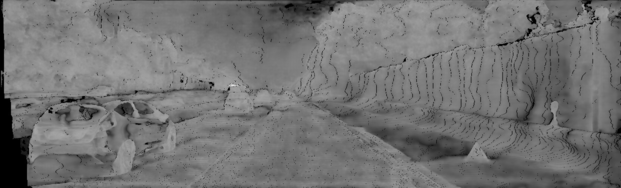}}
        \subfigure{
        \includegraphics[width=0.11\textwidth, frame]{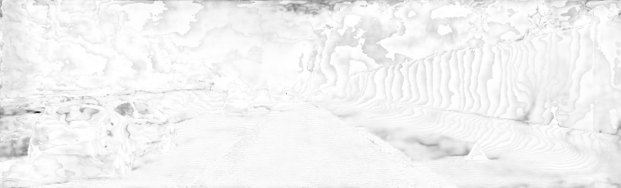}}
        \subfigure{
        \includegraphics[width=0.11\textwidth, frame]{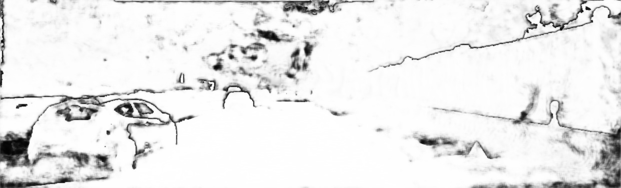}  }
        \\
        \rotatebox{90}{Middlebury}
        \subfigure{
        \includegraphics[width=0.11\textwidth, frame]{images/qualitative/census-CBCA/MiddEval3/Playtable.png}}
        \subfigure{
        \includegraphics[width=0.11\textwidth, frame]{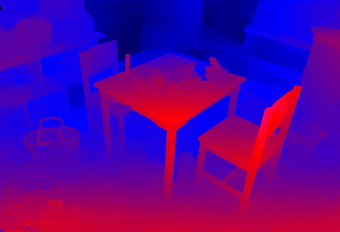}}
        \subfigure{
        \includegraphics[width=0.11\textwidth, frame]{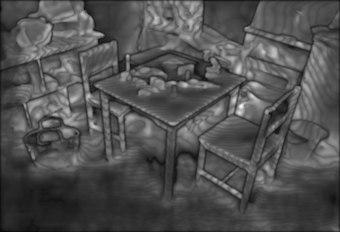}}
        \subfigure{
        \includegraphics[width=0.11\textwidth, frame]{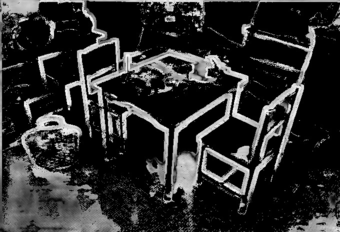}} 
        \subfigure{
        \includegraphics[width=0.11\textwidth, frame]{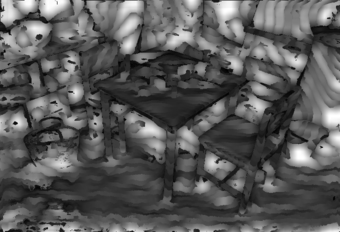}}
        \subfigure{
        \includegraphics[width=0.11\textwidth, frame]{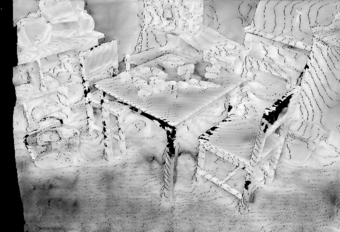}} 
        \subfigure{
        \includegraphics[width=0.11\textwidth, frame]{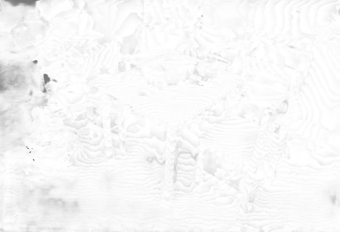}}
        \subfigure{
        \includegraphics[width=0.11\textwidth, frame]{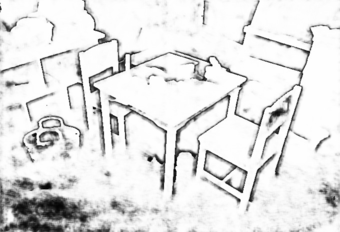}  }
        \\

    \caption{\textbf{Qualitative results concerning GANet.} Results on KITTI 2015 and Middlebury showing a variety of confidence measures. From top left to right: reference image, disparity map and confidence maps by APKR$_7$, WMN, DA$_{31}$, UCC, SAMM and LAF.} 
    \label{fig:qualitative_ganet}
\end{figure*}

\textbf{Impact of the windows size.} Figure \ref{fig:ganet_radius} plots the AUC achieved by varying the radius of $N(p)$ for measures computed over a local neighborhood. Except for DA and DS, saturating respectively on $11\times11$ and $15\times15$ windows, all features computed from the \colorbox{dispcolor}{disparity map} show their best performance with a window size of 21. \colorbox{localcolor}{Local properties} and IVAR achieve their best accuracy on $5\times5$ kernels, rapidly degrading with larger windows.

\textbf{Learned measures, synthetic data training.} Table \ref{tab:ganet_learning}, on the left, collects results for learned measures when trained on synthetic images from the Driving train split. 
Not surprisingly, \colorbox{deepvol}{cost-volume CNNs} cover the top 3 positions with MPN, ACN and CRNN, followed by \colorbox{forestvol}{cost-volume forests}. O1 and O2 are the only \colorbox{forestdisp}{disparity forests} appearing in the first ten positions and \colorbox{deepdisp}{disparity CNNs} perform much worse with the best one, LGC, ranking 11. This outcome occurs because of the very smooth disparity maps produced by GANet, over which finding outliers without analyzing the cost volume is particularly challenging.

Regarding generalization to real data, \colorbox{deepvol}{cost-volume CNNs} cover the top-3 positions with RC, CRNN and MPN. They are followed by LEV$_{50}$ and GCP, respectively, with ranks 4 and 6. The first \colorbox{deepdisp}{disparity CNN} is PBCP$_d$ with rank 8. Finally, \colorbox{forestdisp}{disparity forests} such as O1 and O2, very effective on synthetic data, shows poor generalization and are at the bottom of the leaderboard.

\textbf{Learned measures, real data training.} Table \ref{tab:ganet_learning}, on the right, reports results for learned measures trained on KITTI 2012 20 training images. Surprisingly, the top-2 methods are LEV and LEV$_{50}$, \ie{} \colorbox{forestvol}{cost-volume forests}, followed by \colorbox{deepvol}{cost-volume CNNs} CRNN, RC, LAF and ACN.
The first method processing disparity only is a \colorbox{forestdisp}{disparity forest}, \ie{} O2 ranking 8, while \colorbox{forestdisp}{disparity CNNs} show up only from position 10 with LGC.
Within patch-based methods, LFN, PBCP$_r$ and CCNN are the three most effective, starting with rank 12, while PBCP$_d$ is at the bottom of the leaderboard.

\textbf{Qualitative results.} As for previous experiments, Figure \ref{fig:qualitative_ganet} reports disparity and confidence maps from KITTI 2015 and Middlebury.
When dealing with the volumes produced by GANet, we can notice how some hand-crafted measures are not particularly meaningful, as in the case of WMN, while others remain effective. Not surprisingly, learned measures (\ie{} LAF) better distinguish the few outliers from the large amount of correct matches.

\textbf{Summary.} When dealing with a modern, deep network such as GANet, measures processing the disparity map alone, either learned or not, lose most of their effectiveness. Given the extremely regular structure of the estimated disparity maps, the cost volume becomes a crucial cue to properly estimate the confidence.

\section{Overall summary and discussion}
\label{sec:discussion}

Given the exhaustive experiments carried out in this paper, we summarize next the key findings.

Concerning hand-crafted measures:
\begin{itemize}
    \item For traditional algorithms, \colorbox{dispcolor}{disparity features} are meaningful cues to estimate a confidence measure and some of them (e.g., DA, VAR) often achieves surprising results. 
    
    \item The local content is also a strong cue in both cost volume/disparity map allowing APKR to rank within the top 4 hand-crafted methods with any CBCA/SGM variant.
    
    \item Although very popular, measures exploiting the consistency between \colorbox{lrcolor}{left-right} images achieve average performance. Among them, LRD and, more frequently, the uniqueness constraint consistently represent the best approaches.
    
    \item Not surprisingly, \colorbox{imagecolor}{image priors} alone can not provide reliable information about confidence since different stereo algorithms may be less or more robust to image content, such as in the case of textureless regions. In particular, their AUC is often higher than D1, worse than random selection.
    
    \item When dealing with GANet, \colorbox{dispcolor}{disparity features} alone are no longer enough and consistently achieve poor results. Measures processing the \colorbox{fullcolor}{entire cost curve} or \colorbox{localcolor}{local properties} seem the most effective. Surprisingly, PKR and WMN perform poorly and are outperformed by their naive counterparts, probably because of the soft-argmax operator used for training that forces matching distributions to be unimodal. This effect is softened by local content, as seen for APKR and WPKR.
    
\end{itemize}

Concerning learned measures:
\begin{itemize}
    \item These methods generalize well across synthetic and real environments compared to other tasks, such as disparity estimation, without requiring aggressive data augmentation or thousands of training samples. This behavior is due to the much more regular domain (i.e., disparity and matching costs) observed by forests and networks.

    \item Despite the inliers/outliers distribution is sometimes strongly unbalanced (i.e., for SGM algorithms and GANet), forests and CNNs learn to infer significant confidence scores that outperform traditional ones, although with minor margins compared to what occurs for CBCA algorithms.
    
    \item Drops occurring when moving between KITTI and Middlebury/ETH3D datasets are not marginal, because of the very different structure (\ie, geometry) of the observed environments and results to have higher impact with respect to image content. This fact makes learned methods often close to hand-crafted measures on Middlebury/ETH3D and, in some cases, even outperformed.

    \item Among learned methods, \colorbox{deepvol}{cost-volume CNNs} confirm to be the overall winning family, with \colorbox{deepdisp}{disparity CNNs} being competitive in particular when dealing with noisy stereo algorithms.

\end{itemize}

\section{Conclusion}
\label{sec:conclusions}

In this paper, we have presented an exhaustive review and evaluation of the state-of-the-art strategy to estimate stereo matching confidence. We have reviewed more than ten years of developments in this field, ranging from hand-engineered confidence measures to modern machine learning and deep learning solutions. Moreover, we have carried out an extensive evaluation for a thorough understanding of the topic, involving five stereo algorithms/networks and five datasets.
We believe this review can represent a useful reference for researchers working in depth from stereo and practitioners willing to deploy stereo algorithms in the wild.
Despite the significant improvement yielded by learning-based strategies, improving their generalization across real domains is crucial as a future research direction.

\textbf{Acknowledgments} We gratefully acknowledge the support of NVIDIA Corporation with the donation of the Titan Xp GPU used for this research. The work of S. Kim was supported by the MSIT (Ministry of Science and ICT), Korea, under the ICT Creative Consilience program (IITP-2021-0-01819) supervised by the IITP. 

\bibliographystyle{IEEEtran}
\bibliography{stereo}

\vspace{-1.2cm}
\begin{IEEEbiography}[{\includegraphics[width=1in,height=1.25in,clip,keepaspectratio]{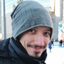}}]{Matteo Poggi}
received his PhD degree in Computer Science and Engineering from University
of Bologna 2018. Currently, he is a Post-doc researcher at Department of Computer Science and Engineering, University of Bologna. His research interests include deep learning for depth estimation and embedded computer vision. He is the author of $\sim$40 papers on these topics.
\end{IEEEbiography}
\vspace{-1.2cm}

\begin{IEEEbiography}[{\includegraphics[width=1in,height=1.25in,clip,keepaspectratio]{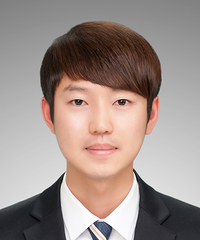}}]{Seungryong Kim}
received the B.S. and Ph.D. degrees from the School of Electrical and Electronic Engineering from Yonsei University, Seoul, Korea, in 2012 and 2018, respectively. From 2018 to 2019, he was Post-Doctoral Researcher in Yonsei University, Seoul, Korea. From 2019 to 2020, he has been Post-Doctoral Researcher in School of Computer and Communication Sciences at \'{E}cole Polytechnique F\'{e}d \'{e}rale de Lausanne (EPFL), Lausanne, Switzerland. Since 2020, he has been an assistant professor with the Department of Computer Science and Engineering, Korea University, Seoul. His current research interests include 2D/3D computer vision, computational photography, and machine learning.
\end{IEEEbiography}
\vspace{-1.2cm}

\begin{IEEEbiography}[{\includegraphics[width=1in,height=1.25in,clip,keepaspectratio]{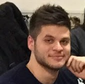}}]{Fabio Tosi}
 received the Master degree in Computer Science and Engineering at Alma
Mater Studiorum, University of Bologna in 2017. He is currently in the PhD program in Computer Science and Engineering of University of Bologna, where he conducts research in deep learning and depth sensing related topics.
\end{IEEEbiography}
\vspace{-1.2cm}

\begin{IEEEbiography}[{\includegraphics[width=1in,height=1.25in,clip,keepaspectratio]{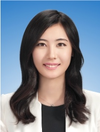}}]{Sunok Kim}
    (M'18) received the B.S. and ph.D. degrees from the School of Electrical and Electronic Engineering from Yonsei University, Seoul, Korea, in 2014 and 2019. 
	Since 2019, she has been Post-Doctoral Researcher in School of Electrical and Electronic Engineering at Yonsei University.
	Her current research interests include 3D image processing and computer vision, 
	in particular, stereo matching, depth super-resolution, and confidence estimation.
\end{IEEEbiography}
\vspace{-1.2cm}

\begin{IEEEbiography}[{\includegraphics[width=1in,height=1.25in,clip,keepaspectratio]{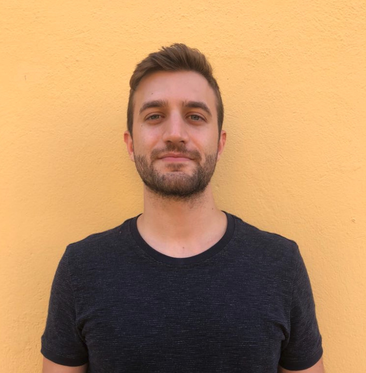}}]{Filippo Aleotti}
 received the Master degree in Computer Science and Engineering at Alma
Mater Studiorum, University of Bologna in 2018. He is currently in the PhD program in Structural and Environmental Health Monitoring and Management (SEHM2) of University of Bologna, where he conducts research in deep
learning for depth sensing.
\end{IEEEbiography}
\vspace{-1.2cm}

\begin{IEEEbiography}[{\includegraphics[width=1in,height=1.25in,clip,keepaspectratio]{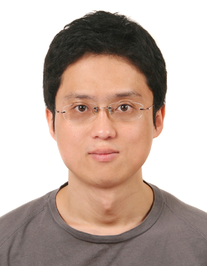}}]{Dongbo Min}
received the BS, MS, and PhD degrees from the School of Electrical and Electronic Engineering, Yonsei University, Seoul, South Korea, in 2003, 2005, and 2009, respectively. From 2009 to 2010, he was a post-doctoral researcher with Mitsubishi Electric Research Laboratories, Cambridge, Massachusetts. From 2010 to 2015, he was with the Advanced Digital Sciences Center, Singapore. From 2015 to 2018, he was an assistant professor in the Department of Computer Science and Engineering, Chungnam National University, Daejeon, South Korea. Since 2018, he has been in the Department of Computer Science and Engineering, Ewha Womans University, Seoul. His current research interests include computer vision, deep learning, video processing, and continuous/discrete optimization. He is a senior member of the IEEE.
\end{IEEEbiography}
\vspace{-1.2cm}

\begin{IEEEbiography}[{\includegraphics[width=1in,height=1.25in,clip,keepaspectratio]{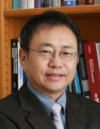}}]{Kwanghoon Sohn}
received the B.E. degree in electronic engineering from
Yonsei University, Seoul, Korea, in 1983, the
M.S.E.E. degree in electrical engineering from
the University of Minnesota, Minneapolis, MN,
USA, in 1985, and the Ph.D. degree in electrical
and computer engineering from North Carolina
State University, Raleigh, NC, USA, in 1992. He
was a Senior Member of the Research engineer with
the Satellite Communication Division, Electronics
and Telecommunications Research Institute,
Daejeon, Korea, from 1992 to 1993, and a Post-Doctoral Fellow with the
MRI Center, Medical School of Georgetown University, Washington, DC,
USA, in 1994. He was a Visiting Professor with Nanyang Technological
University, Singapore, from 2002 to 2003. He is currently an Underwood Distinguished Professor
with the School of Electrical and Electronic Engineering, Yonsei University.
His research interests include 3D image processing and computer vision.
\end{IEEEbiography}
\vspace{-1.2cm}
\begin{IEEEbiography}[{\includegraphics[width=1in,height=1.25in,clip,keepaspectratio]{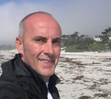}}]{Stefano Mattoccia}
received a Ph.D. degree in Computer Science Engineering from the University of Bologna in 2002. Currently he is an associate professor at the Department of Computer Science and Engineering of
the University of Bologna. His research interest
is mainly focused on computer vision, depth perception, embedded vision and deep learning.
\end{IEEEbiography}

\end{document}